\documentclass[10pt,twocolumn,letterpaper]{article}

\usepackage[pagenumbers]{cvpr} %

\usepackage{graphicx}
\usepackage{amsmath}
\usepackage{amssymb}
\usepackage{booktabs}
\usepackage{duckuments}
\usepackage{algorithm}
\usepackage{algpseudocode}
\usepackage{amsmath}
\usepackage{amsfonts}
\usepackage{amssymb}
\usepackage{mathtools}
\usepackage{float}
\usepackage[10pt]{moresize}

\usepackage{color,xcolor}
\usepackage{epsfig}
\usepackage{graphicx}
\usepackage{times}

\usepackage{comment}

\usepackage{pifont}

\usepackage{microtype}
\usepackage[font=smaller]{caption}
\usepackage{arydshln}
\usepackage{tabularx}
\usepackage{adjustbox}
\usepackage{array}
\usepackage{booktabs}
\usepackage{colortbl}
\usepackage{float,wrapfig}
\usepackage{makecell}
\usepackage{hhline}
\usepackage{multirow}
\usepackage{subcaption} %
\usepackage[percent]{overpic}
\usepackage{graphbox}
\usepackage{diagbox}
\usepackage{stmaryrd}

\usepackage{amsmath,amsfonts,amssymb}
\usepackage{bm}
\usepackage{nicefrac}
\usepackage{microtype}
\usepackage{dsfont}
\usepackage{changepage}
\usepackage{extramarks}
\usepackage{fancyhdr}
\usepackage{setspace}
\usepackage{soul}
\usepackage{xspace}
\usepackage{hhline}

\usepackage[pagebackref,breaklinks,colorlinks]{hyperref}
\usepackage{url}

\usepackage{enumitem}
\usepackage[title]{appendix}

\usepackage{amsmath,amsfonts,bm,relsize,dsfont}
\usepackage[numbers]{natbib}
\usepackage{anyfontsize}

\algrenewcommand{\algorithmicrequire}{\textbf{Input:}}
\algrenewcommand{\algorithmicensure}{\textbf{Output:}}

\definecolor{MyDarkBlue}{rgb}{0,0.08,1}
\definecolor{MyDarkGreen}{rgb}{0.02,0.6,0.02}
\definecolor{MyDarkRed}{rgb}{0.8,0.02,0.02}
\definecolor{MyDarkOrange}{rgb}{0.40,0.2,0.02}
\definecolor{MyPurple}{RGB}{111,0,255}
\definecolor{MyRed}{rgb}{1.0,0.0,0.0}
\definecolor{MyGold}{rgb}{0.75,0.6,0.12}
\definecolor{MyDarkgray}{rgb}{0.66, 0.66, 0.66}
\definecolor{MyDarkCyan}{rgb}{0.05, 0.55, 0.45}
\definecolor{MyBlack}{rgb}{0., 0., 0.}
\definecolor{MyMagenta}{rgb}{1., 0., 1.}
\definecolor{BerkeleyYellow}{RGB}{255,204,41}
\definecolor{BerkeleyLightBlue}{RGB}{94,146,221}
\definecolor{BkDarkBlue}{rgb}{.05,.07,.353}

\definecolor{MyDarkGray2}{rgb}{0.6, 0.6, 0.6}

\NewDocumentCommand{\anote}{}{\makebox[0pt][l]{$^*$}}

\newcommand{\reffig}[1]{Figure~\ref{fig:#1}}
\newcommand{\refsec}[1]{Section~\ref{sec:#1}}

\newcommand{\reftbl}[1]{Table~\ref{tbl:#1}}
\newcommand{\refalg}[1]{Algorithm~\ref{alg:#1}}

\renewcommand{\refeq}[1]{Eqn.~\ref{eq:#1}}

\newcommand{\lblfig}[1]{\label{fig:#1}}
\newcommand{\lblsec}[1]{\label{sec:#1}}
\newcommand{\lbleq}[1]{\label{eq:#1}}
\newcommand{\lbltbl}[1]{\label{tbl:#1}}
\newcommand{\lblalg}[1]{\label{alg:#1}}

\newcommand{\ignorethis}[1]{}

\newcommand{\myparagraph}[1]{\vspace{1pt} \noindent \textbf{#1} \ }

\def\1{\bm{1}}

\newcolumntype{L}[1]{>{\raggedright\let\newline\\\arraybackslash\hspace{0pt}}m{#1}}
\newcolumntype{C}[1]{>{\centering\let\newline\\\arraybackslash\hspace{0pt}}m{#1}}
\newcolumntype{R}[1]{>{\raggedleft\let\newline\\\arraybackslash\hspace{0pt}}m{#1}}

\newcommand{\ignore}[1]{}

\makeatletter
\DeclareRobustCommand\onedot{\futurelet\@let@token\@onedot}
\def\@onedot{\ifx\@let@token.\else.\null\fi\xspace}
\def\eg{e.g\onedot,\xspace} 

\def\ie{i.e\onedot,\xspace}

\makeatother

\usepackage[capitalize]{cleveref}
\crefname{section}{Sec.}{Secs.}
\Crefname{section}{Section}{Sections}
\Crefname{table}{Table}{Tables}
\crefname{table}{Tab.}{Tabs.}

\begin{document}

\title{Dataset Distillation by Matching Training Trajectories}

\author{George Cazenavette\textsuperscript{1}\;\;\;\;\;Tongzhou Wang\textsuperscript{2}\;\;\;\;\;Antonio Torralba\textsuperscript{2}\;\;\;\;\;Alexei A. Efros\textsuperscript{3}\;\;\;\;\;Jun-Yan Zhu\textsuperscript{1}\\
\\
\textsuperscript{1}Carnegie Mellon University\;\;\;\;\;\textsuperscript{2}Massachusetts Institute of Technology\;\;\;\;\;\textsuperscript{3}UC Berkeley
}

\newcommand\coverwidth{0.093}

\twocolumn[{
\maketitle
\ssmall
\begin{center}
\begingroup
\setlength{\tabcolsep}{1pt}
\begin{tabular}{ccccccccccc}

\rotatebox[origin=c]{90}{\scriptsize{CIFAR-100}} &
    \includegraphics[align=c,width=\coverwidth\linewidth]{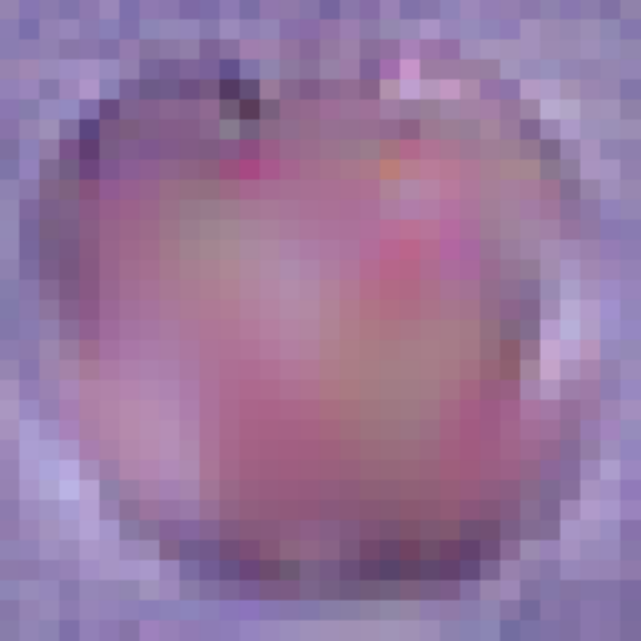} &
    \includegraphics[align=c,width=\coverwidth\linewidth]{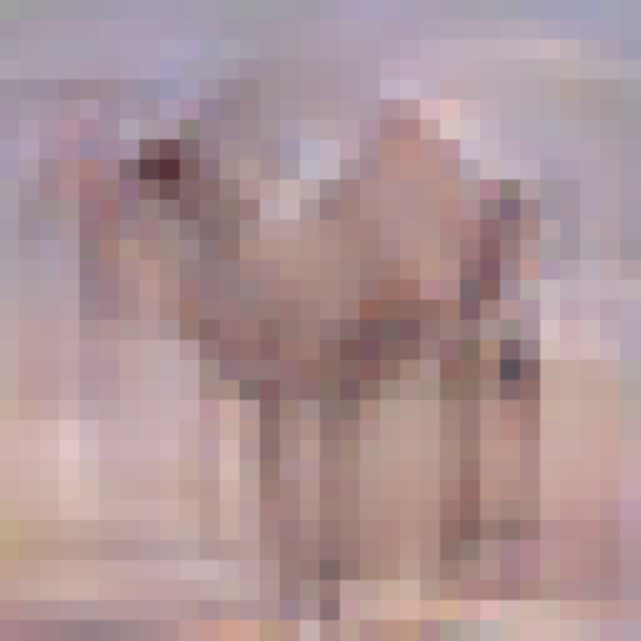} &
    \includegraphics[align=c,width=\coverwidth\linewidth]{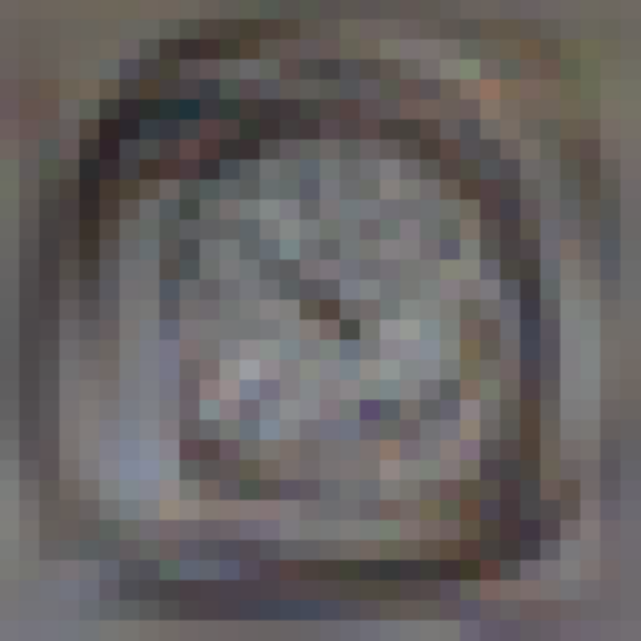} &
    \includegraphics[align=c,width=\coverwidth\linewidth]{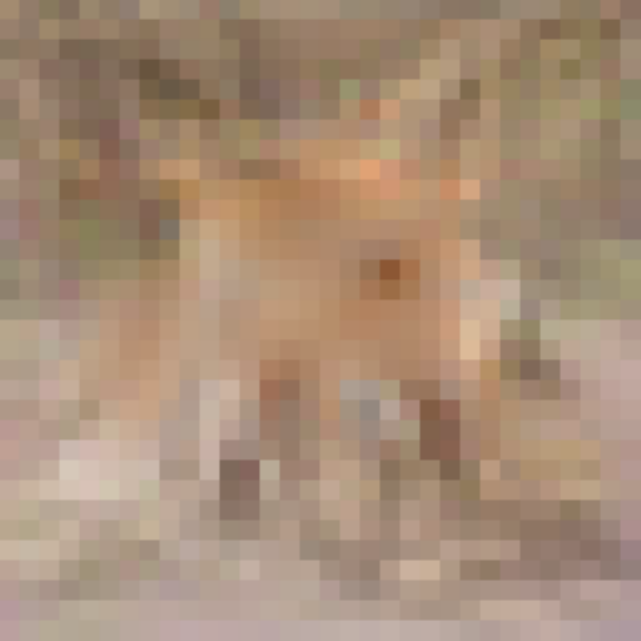} &
    \includegraphics[align=c,width=\coverwidth\linewidth]{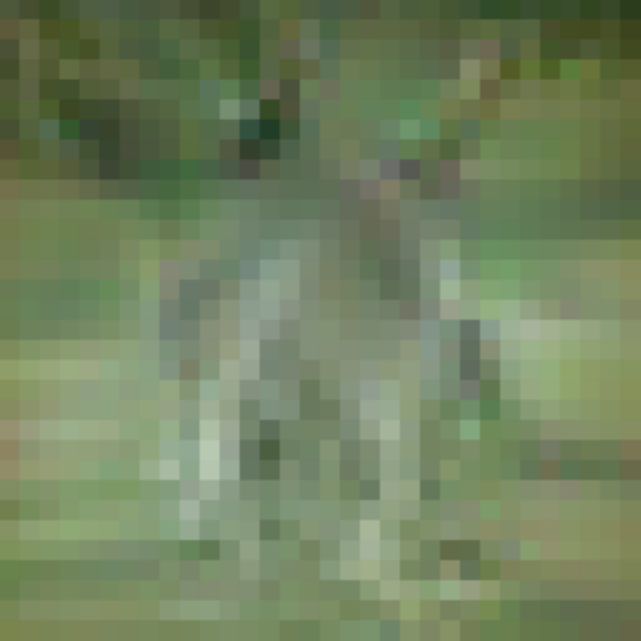} &
    \includegraphics[align=c,width=\coverwidth\linewidth]{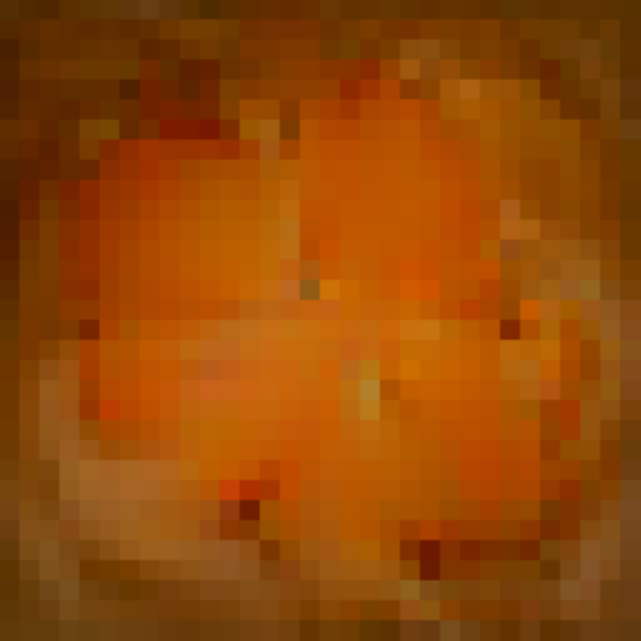} &
    \includegraphics[align=c,width=\coverwidth\linewidth]{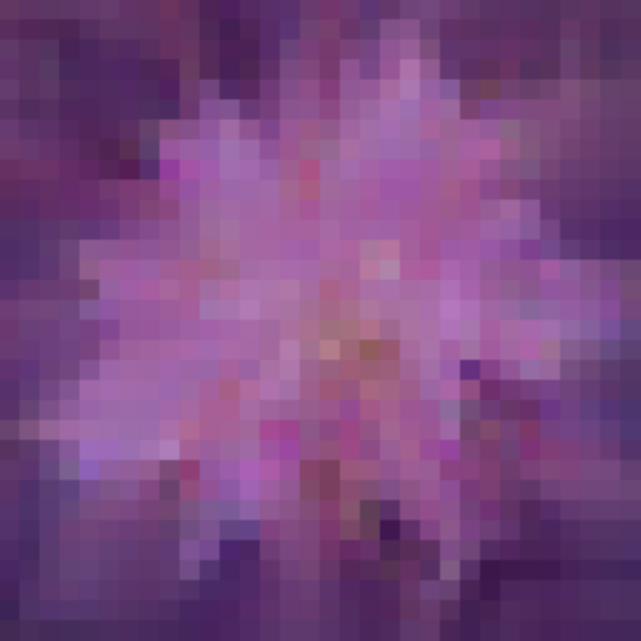} &
    \includegraphics[align=c,width=\coverwidth\linewidth]{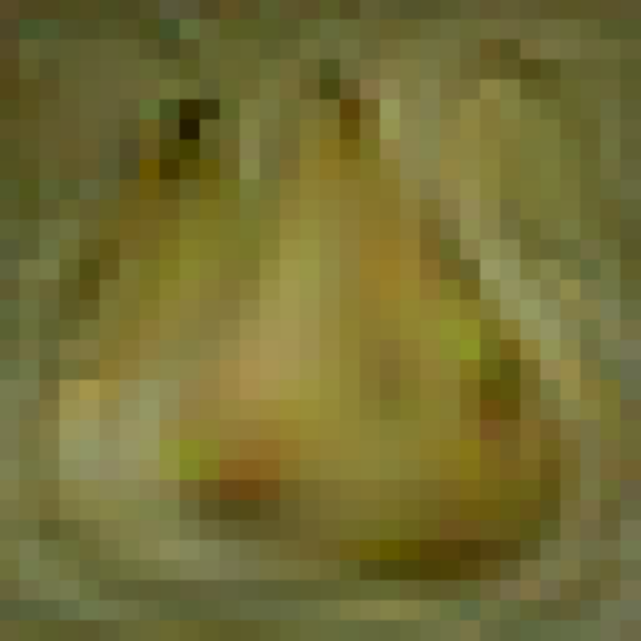} &
    \includegraphics[align=c,width=\coverwidth\linewidth]{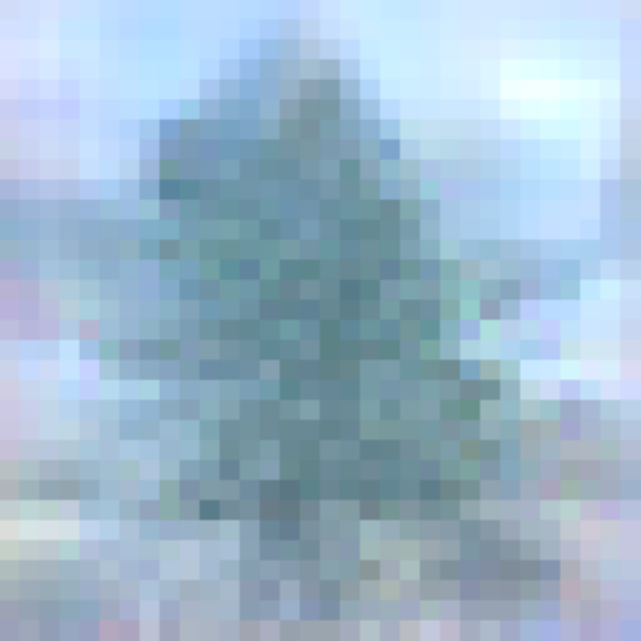} &
    \includegraphics[align=c,width=\coverwidth\linewidth]{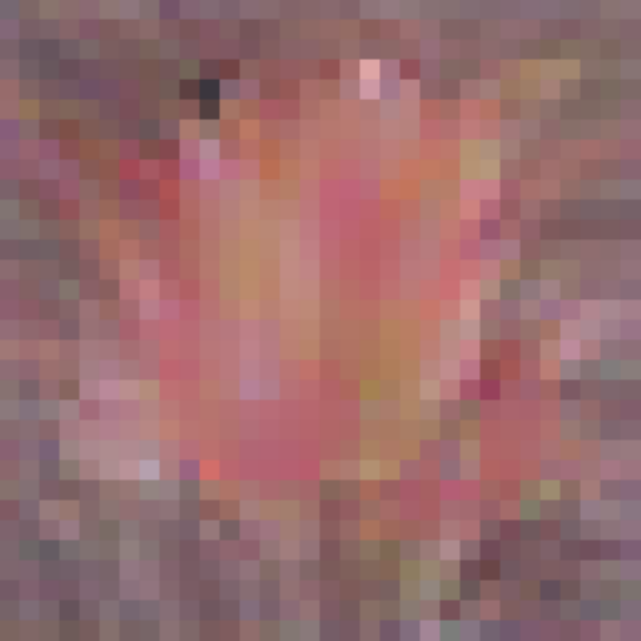} \\[7.8ex]
    & Apple & Camel & Clock & Fox & Kangaroo & Orange & Orchid & Pear & Pine Tree & Tulip\\[-0.15cm]\\
    \rotatebox[origin=c]{90}{\scriptsize{Tiny ImageNet}} &
    \includegraphics[align=c,width=\coverwidth\linewidth]{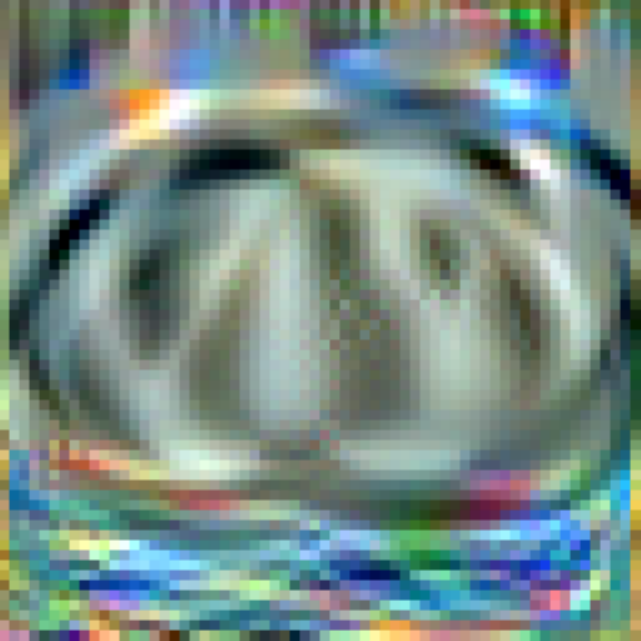} &
    \includegraphics[align=c,width=\coverwidth\linewidth]{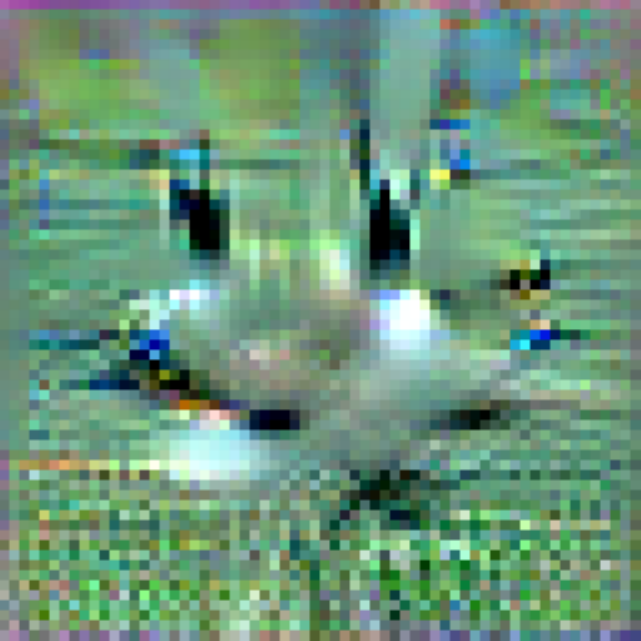} &
    \includegraphics[align=c,width=\coverwidth\linewidth]{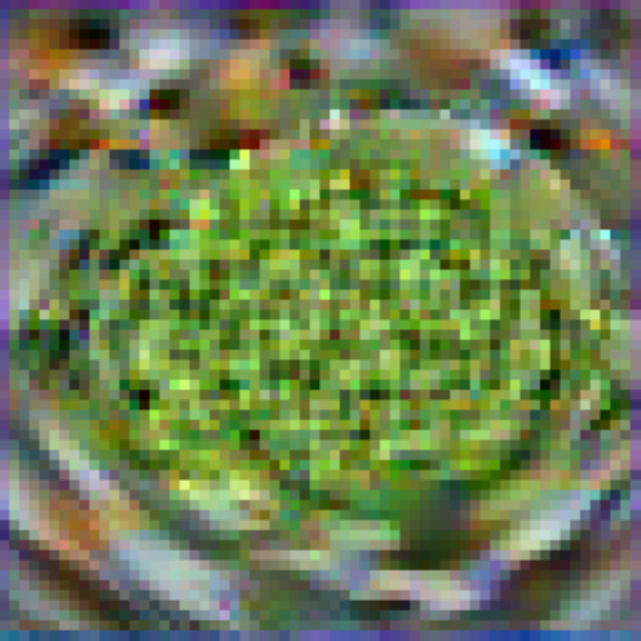} &
    \includegraphics[align=c,width=\coverwidth\linewidth]{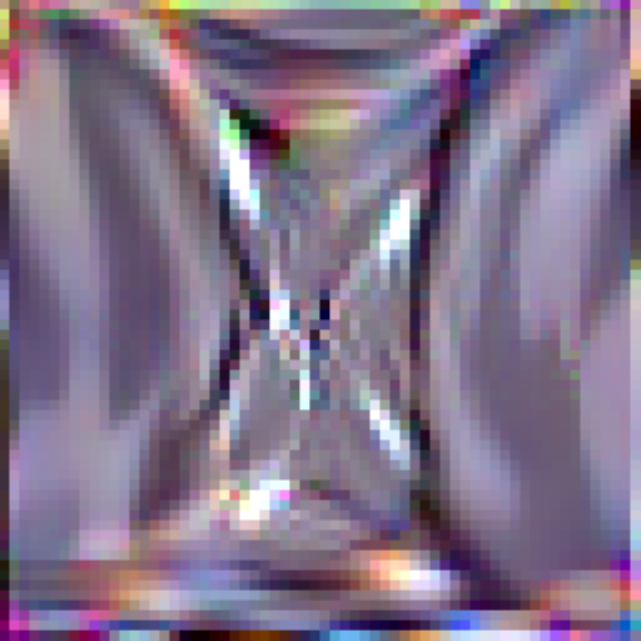} &
    \includegraphics[align=c,width=\coverwidth\linewidth]{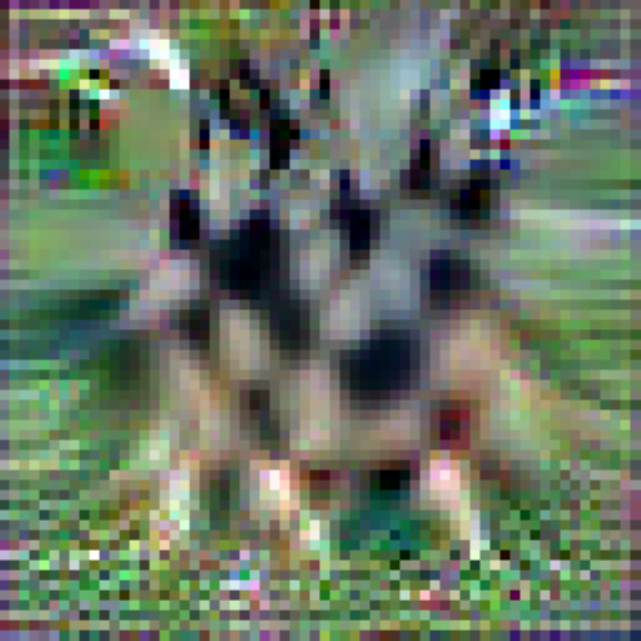} &
    \includegraphics[align=c,width=\coverwidth\linewidth]{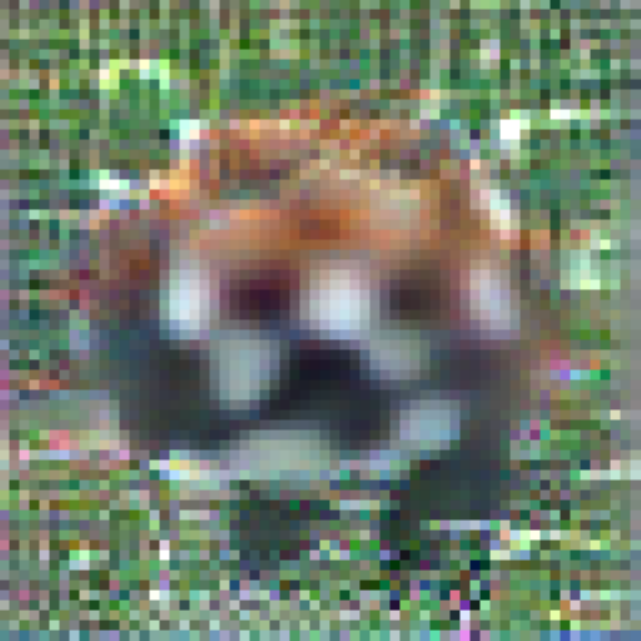} &
    \includegraphics[align=c,width=\coverwidth\linewidth]{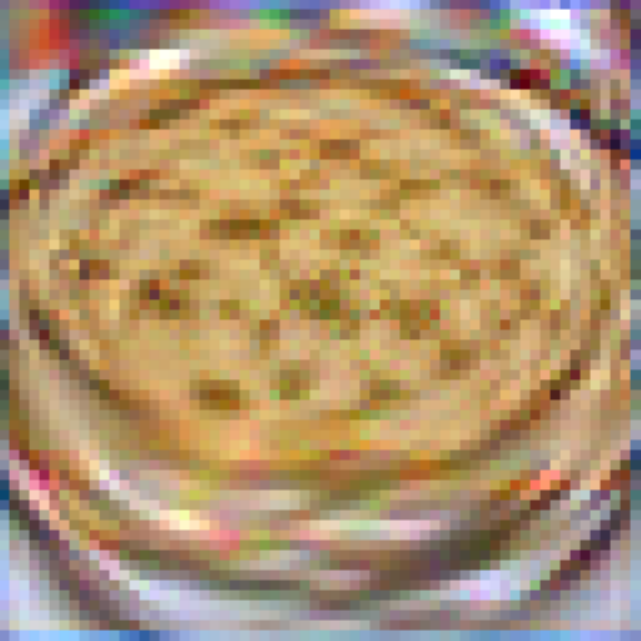} &
    \includegraphics[align=c,width=\coverwidth\linewidth]{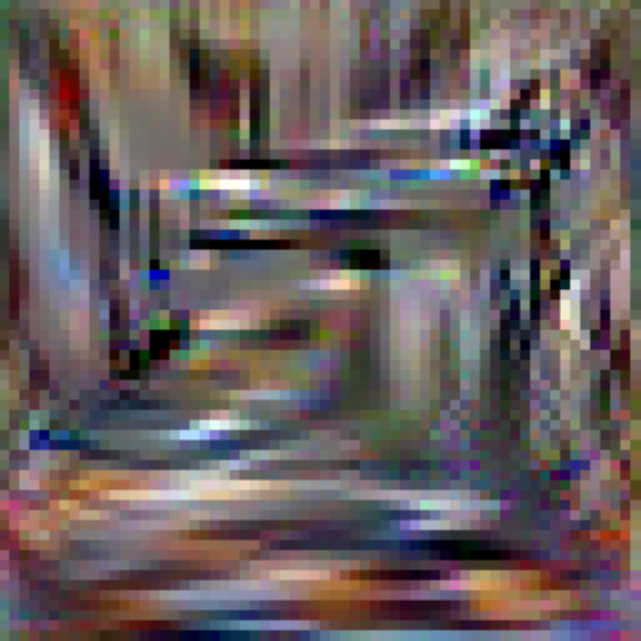} &
    \includegraphics[align=c,width=\coverwidth\linewidth]{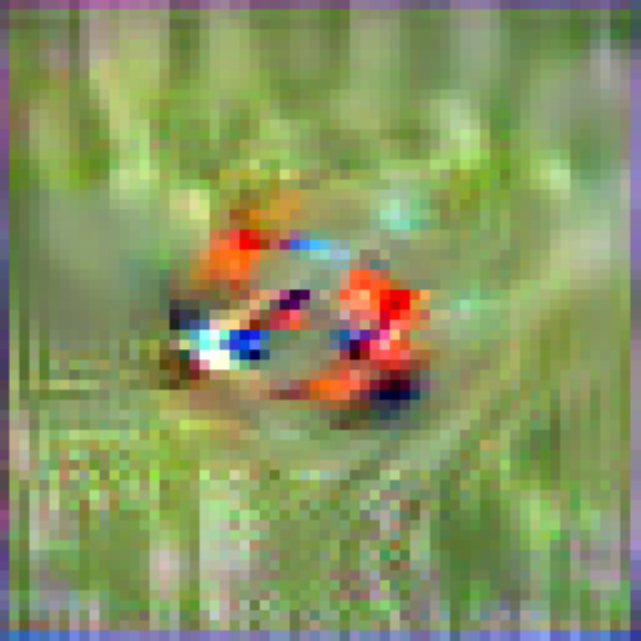} &
    \includegraphics[align=c,width=\coverwidth\linewidth]{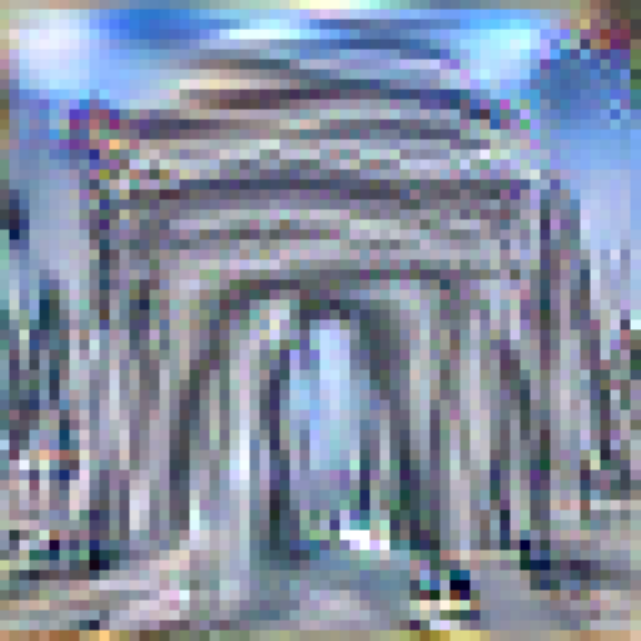}\\[7.8ex]
    &Guinea Pig & Goose & Guacamole & Hourglass & German Shepard & Red Panda & Potpie & Sewing Machine & Ladybug & Triumphal Arch\\[-0.15cm]\\
    \rotatebox[origin=c]{90}{\scriptsize{ImageNet}} &
    \includegraphics[align=c,width=\coverwidth\linewidth]{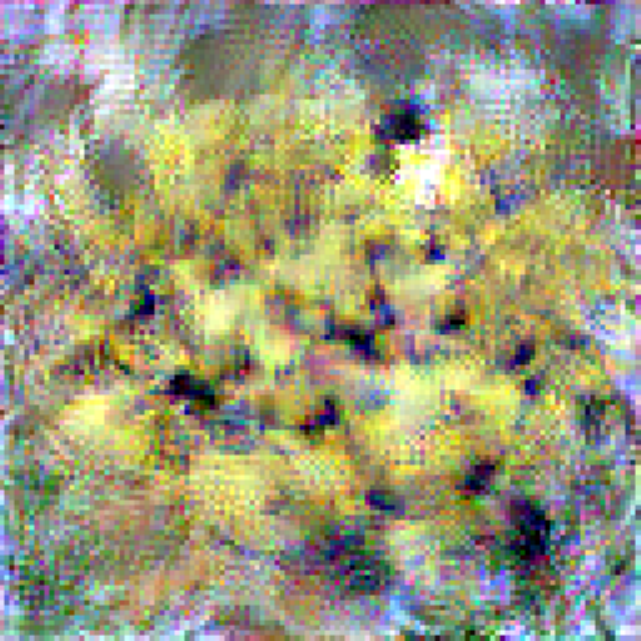} &
    \includegraphics[align=c,width=\coverwidth\linewidth]{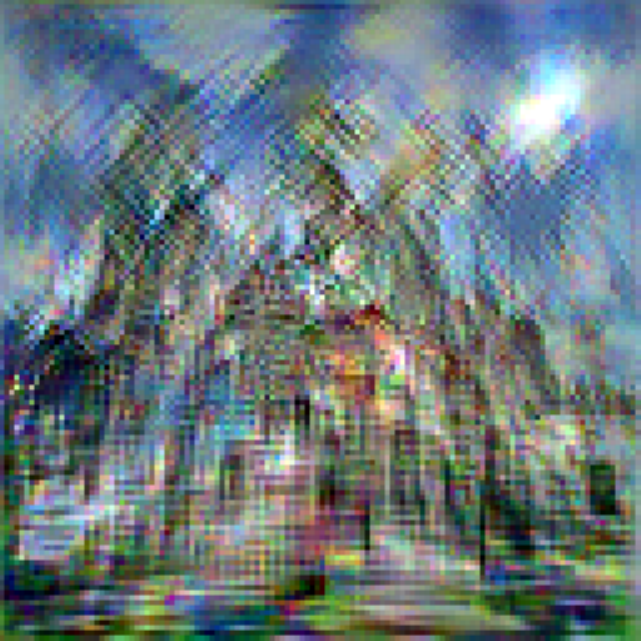} &
    \includegraphics[align=c,width=\coverwidth\linewidth]{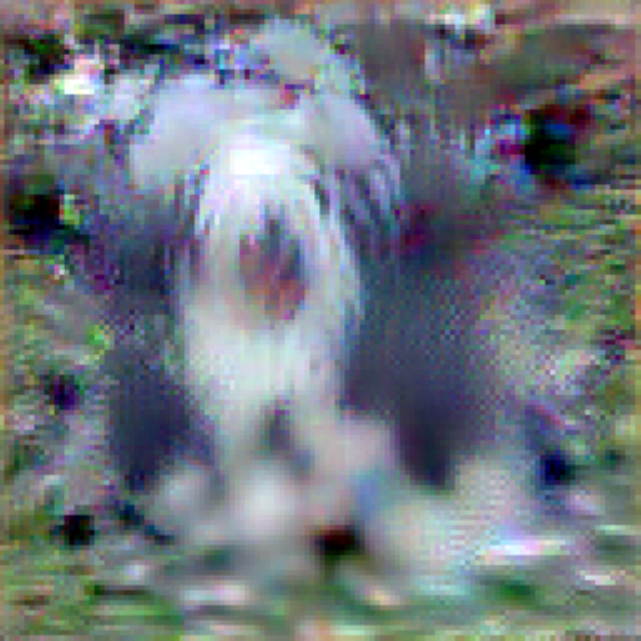} &
    \includegraphics[align=c,width=\coverwidth\linewidth]{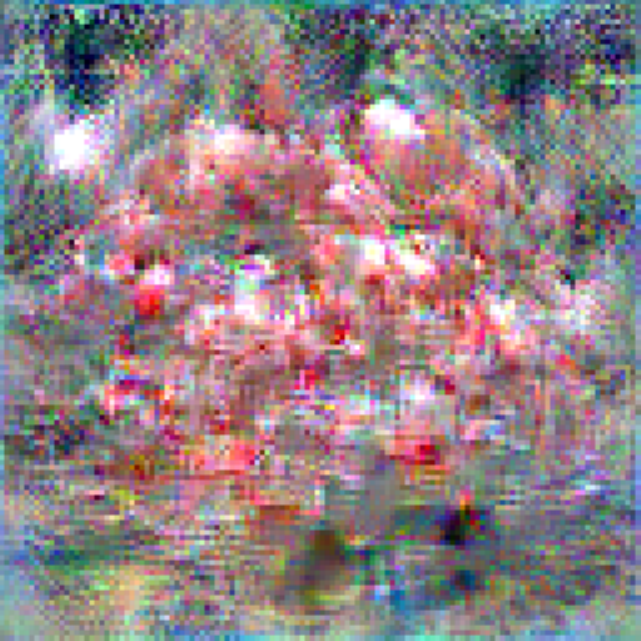} &
    \includegraphics[align=c,width=\coverwidth\linewidth]{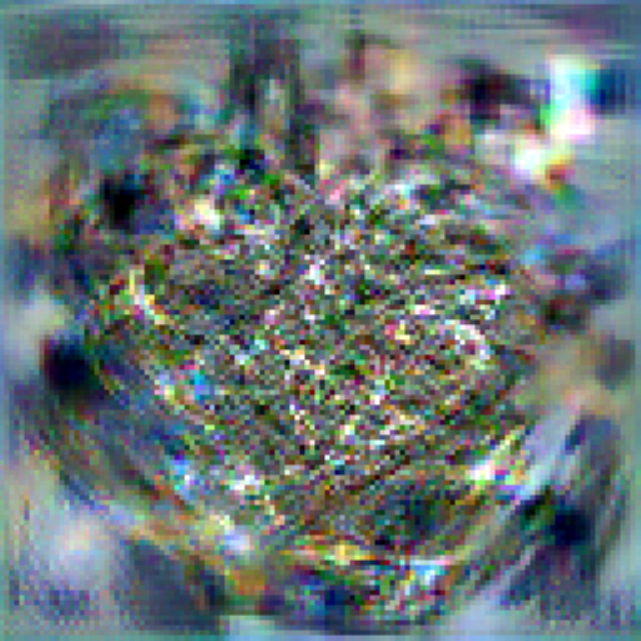} &
    \includegraphics[align=c,width=\coverwidth\linewidth]{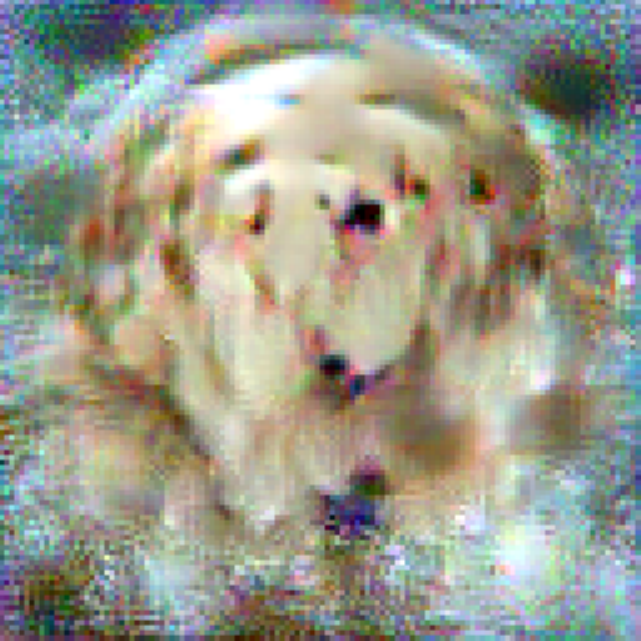} &
    \includegraphics[align=c,width=\coverwidth\linewidth]{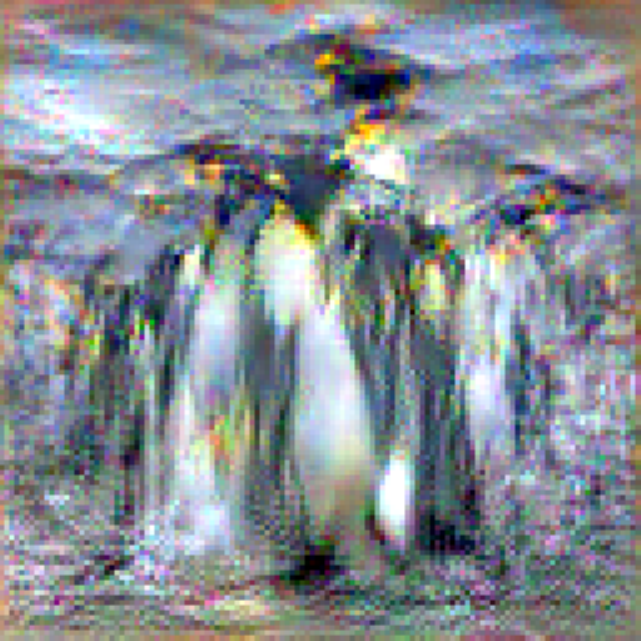} &
    \includegraphics[align=c,width=\coverwidth\linewidth]{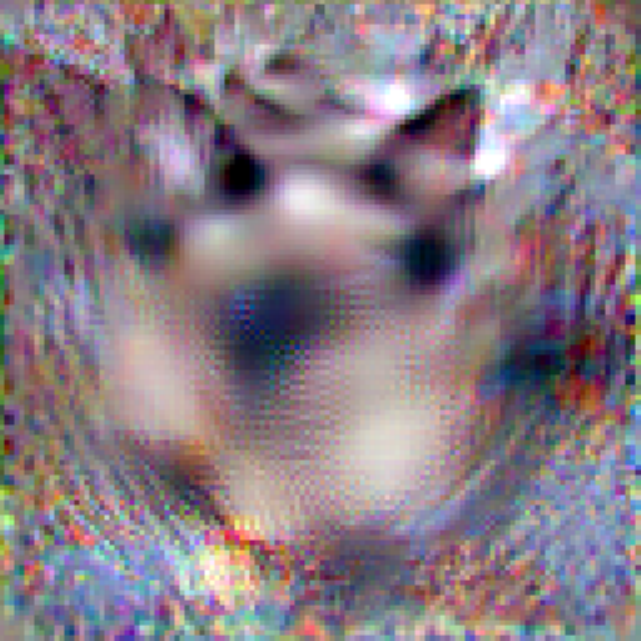} &
    \includegraphics[align=c,width=\coverwidth\linewidth]{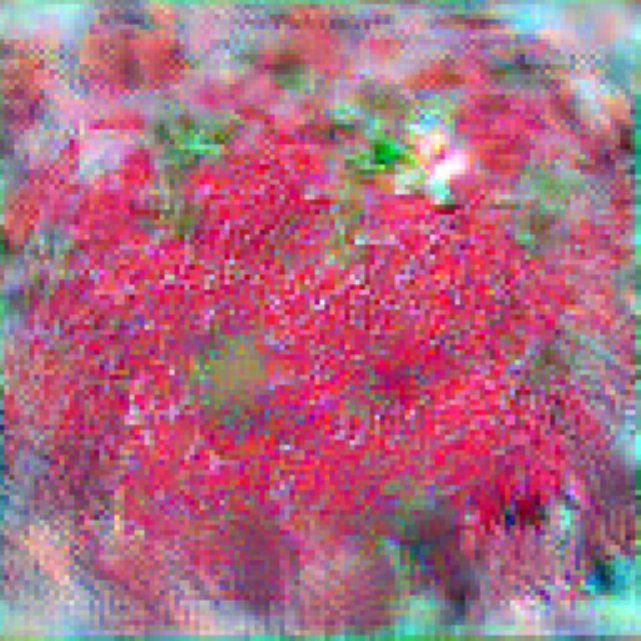} &
    \includegraphics[align=c,width=\coverwidth\linewidth]{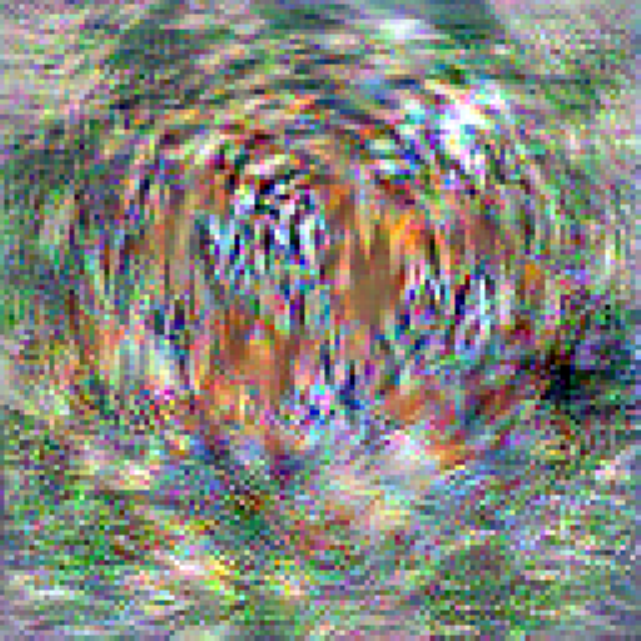} \\[7.8ex]
    & Banana & Church & Sheepdog & Flamingo & French Horn & Golden Retriever & King Penguin & Siamese Cat & Strawberry & Tiger 
\end{tabular}
\endgroup\vspace{-0.25cm}
    \captionof{figure}{Example distilled images from 32x32 CIFAR-100 (top), 64x64 Tiny ImageNet (middle), and 128x128 ImageNet subsets (bottom). 
    Training a standard CNN using only such distilled images (as few as one per category) yields a trained model capable of test accuracy significantly better than previous methods of dataset distillation. Please see more results at \href{https://georgecazenavette.github.io/mtt-distillation/}{\color{blue}https://georgecazenavette.github.io/mtt-distillation/}.
    }
    \lblfig{teaser}
\end{center}
}]

\begin{abstract}
   Dataset distillation is the task of synthesizing a small dataset such that a model trained on the synthetic set will match the test accuracy of the model trained on the full dataset.
   In this paper, we propose a new formulation that optimizes our distilled data to guide networks to a similar state as those trained on real data across many training steps. 
   Given a network, we train it for several iterations on our distilled data and optimize the distilled data with respect to the distance between the synthetically trained parameters and the parameters trained on real data. To efficiently obtain the initial and target network parameters for large-scale datasets, we pre-compute and store training trajectories of expert networks trained on the real dataset. 
   Our method handily outperforms existing methods and also allows us to distill higher-resolution visual data.
\end{abstract}

\section{Introduction}
\lblsec{intro}

\begin{figure}
    \centering
    \vspace{-0.5cm}%
    \includegraphics[scale=0.58, trim=230 106 150 20, clip]{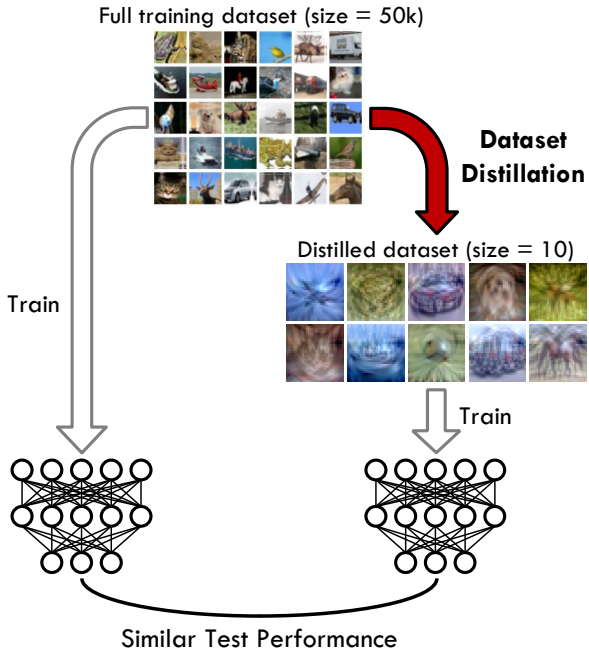}\vspace{-0.85cm}
    \caption{Dataset distillation aims to generate a small synthetic dataset for which a model trained on it can achieve a similar test performance as a model trained on the whole real train set.}%
    \lblfig{dataset_distillation}\vspace{-5pt}
\end{figure}

In the seminal 2015 paper, Hinton et al.~\cite{hinton2015distilling} proposed {\em model distillation}, which aims to distill the knowledge of a complex model into a simpler one.  
{\em Dataset distillation}, proposed by Wang et al.~\cite{dd}, is a related but orthogonal task: rather
than distilling the model, the idea is to distill the dataset. 
As shown in Figure~\ref{fig:dataset_distillation}, the goal is to distill the knowledge from a large training dataset into a very small set of synthetic training images (as low as one image per class) such that training a model on the distilled data would give a similar test performance as training one on the original dataset. Dataset distillation has become a lively research topic in machine learning~\cite{bohdal2020flexible,sucholutsky2021soft,dc,dsa,nguyen2020dataset,nguyen2021dataset,dm} with various applications, such as continual learning, neural architecture search, and privacy-preserving ML. Still, the problem has so far been of mainly theoretical interest, since most prior methods focus on toy datasets, like MNIST and \mbox{CIFAR}, while struggling on real, higher-resolution images. In this work, we present a new approach to dataset distillation that not only outperforms previous work in performance, but is also applicable to large-scale datasets, as shown in \reffig{teaser}.

Unlike classical data compression, dataset distillation aims for a small synthetic dataset that still retains adequate task-related information so that models trained on it can generalize to unseen test data, as shown in \reffig{dataset_distillation}. Thus, the distilling algorithm must strike a delicate balance by heavily compressing information without completely obliterating the discriminative features.  
To do this, dataset distillation methods attempt to discover exactly which aspects of the real data are critical for learning said discrimination. Several methods consider end-to-end training~\cite{dd,nguyen2020dataset,nguyen2021dataset} but often require huge compute and memory and suffer from inexact relaxation~\cite{nguyen2020dataset,nguyen2021dataset} or training instability of unrolling many iterations~\cite{dd,maclaurin2015gradient}. To reduce the optimization difficulty, other methods~\cite{dc,dsa} focus on short-range behavior, enforcing a single training step on distilled data to match that on real data. However, error may accumulate in evaluation, where the distilled data is applied over many steps. We confirm this hypothesis experimentally in \refsec{shortlongrange}.

To address the above challenges, we sought to directly imitate the long-range training dynamics of networks trained on real datasets. In particular, we match segments of parameter trajectories trained on synthetic data with segments of pre-recorded trajectories from models trained on real data and thus avoid being short-sighted (\ie focusing on single steps) or difficult to optimize (\ie modeling the full trajectories). 
Treating the real dataset as the gold standard for guiding the network's training dynamics, we can consider the induced sequence of network parameters to be an \emph{expert trajectory}. If our distilled dataset were to induce a network's training dynamics to follow these expert trajectories, then the synthetically trained network would land at a place close to the model trained on real data (in the parameter space) and achieve similar test performance. %

\begin{figure*}
    \centering
    \vspace{-7pt}
    \includegraphics[scale=0.6, trim=0 42 0 190, clip]{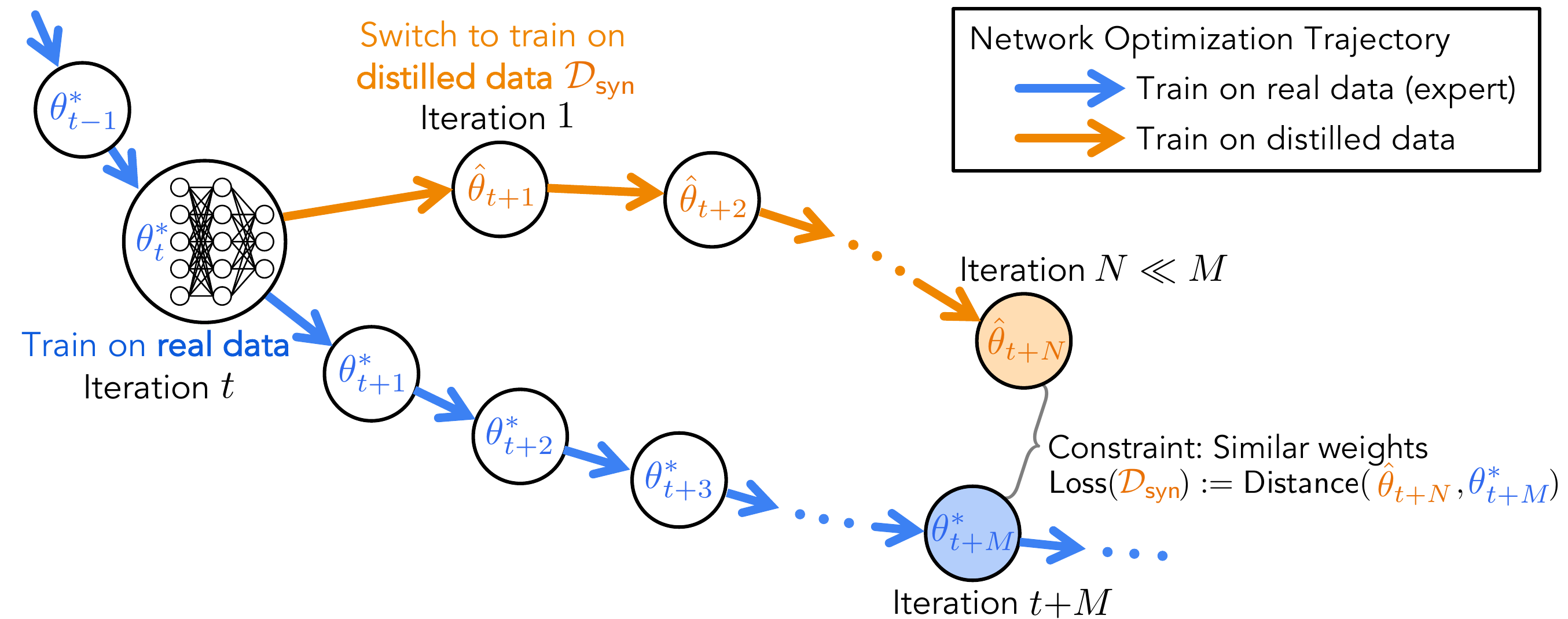}\vspace{-0.18cm}
    \caption{We perform long-range parameter matching between training on distilled synthetic data and training on real data. Starting from the same initial parameters, we train distilled data $\mathcal{D}_\mathsf{syn}$ such that $N$ training steps on them match the same result (in parameter space) from much more $M$ steps on real data.}
    \lblfig{method_traj_match}
         \vspace{-6pt}
\end{figure*}

In our method, our loss function \textit{directly} encourages the distilled dataset to guide the network optimization along a similar trajectory (\Cref{fig:method_traj_match}). We first train a set of models from scratch on the real dataset and record their expert training trajectories. We then initialize a new model with a random time step from a randomly chosen expert trajectory and train for several iterations on the \textit{synthetic} dataset. Finally, we penalize the distilled data based on how far this synthetically trained network deviated from the expert trajectory and back-propagate through the training iterations. Essentially, we transfer the knowledge from many expert training trajectories to the distilled images. 

Extensive experiments show that our method handily outperforms existing dataset distillation methods as well as coreset selection methods on standard datasets, including CIFAR-10, CIFAR-100, and Tiny ImageNet. For example, we achieve 46.3\% with a \emph{single} image per class and 71.5\% with 50 images per class on CIFAR-10, compared to the previous state of the art (28.8\% / 63.0\% from \cite{dsa, dm} and 36.1\% / 46.5\% from \cite{nguyen2021dataset}).
Furthermore, our method also generalizes well to larger data, allowing us to see high $128\times128$-resolution images distilled from ImageNet \cite{deng2009imagenet} for the first time. Finally, we analyze our method through additional ablation studies and visualizations. %
\href{https://github.com/GeorgeCazenavette/mtt-distillation}{Code} and models are also available on our \href{https://georgecazenavette.github.io/mtt-distillation/}{webpage}.
\lblfig{intro}

\section{Related Work}

\myparagraph{Dataset Distillation.}
Dataset distillation was first introduced by Wang et al.~\cite{dd}, who proposed expressing the model weights as a function of distilled images and optimized them using gradient-based hyperparameter optimization~\cite{maclaurin2015gradient}, which is also widely used in meta-learning research \citep{finn2017model,nichol2018first}. %
Subsequently, several works significantly improved the results by learning soft labels \cite{bohdal2020flexible,sucholutsky2021soft}, amplifying learning signal via gradient matching \cite{dc}, adopting augmentations~\cite{dsa}, and optimizing with respect to the infinite-width kernel limit \cite{nguyen2020dataset,nguyen2021dataset}. Dataset distillation has enabled various applications including continual learning~\cite{dd,dc,dsa}, efficient neural architecture search~\cite{dc,dsa}, federated learning~\cite{goetz2020federated,zhou2020distilled,sucholutsky2020secdd}, and privacy-preserving ML~\cite{sucholutsky2020secdd,li2020soft} for images, text, and medical imaging data. As mentioned in the introduction, our method does not rely on single-step behavior matching~\cite{dc,dsa},  costly unrolling of full optimization trajectories~\cite{dd,sucholutsky2021soft}, or large-scale Neural Tangent Kernel computation~\cite{nguyen2020dataset,nguyen2021dataset}. Instead, our method achieves long-range trajectory matching by transferring the knowledge from pre-trained experts. %

Concurrent with our work, the method of Zhao and Bilen~\cite{dm} completely disregards optimization steps, instead focusing on distribution matching between synthetic and real data. While this method is applicable to higher-resolution datasets (\eg Tiny ImageNet) due to reduced memory requirements, it attains inferior performance in most cases (\eg when compared to previous works~\cite{dc,dsa}). In contrast, our method simultaneously reduces memory costs while outperforming existing works~\cite{dc,dsa} and the concurrent method~\cite{dm} on both standard benchmarks and higher-resolution datasets. 

A related line of research learns a generative model to synthesize training data \cite{such2020generative,masarczyk2020reducing}. However, such methods do not generate a small-size dataset, and are thus not directly comparable with dataset distillation methods.

\myparagraph{Imitation Learning.} 
Imitation learning attempts to learn a good policy by observing a collection of expert demonstrations \cite{osa2018algorithmic,peng2018deepmimic,peng2021amp}. Behavior cloning trains the learning policy to act the same as expert demonstrations. Some more sophisticated formulations involve on-policy learning with labeling from the expert \cite{ross2011reduction}, while other approaches avoid any label at all, \eg via distribution matching
\cite{ho2016generative}. Such methods (behavior cloning in particular) have been shown to work well in offline settings \cite{fu2020d4rl,gulcehre2020rl}. 
Our method can be viewed as imitating a collection of expert network training trajectories, which are obtained via training on real datasets. Therefore, it can be considered as doing imitation learning over optimization trajectories.

\myparagraph{Coreset and Instance Selection.}
Similar to dataset distillation, coreset~\cite{tsang2005core,har2007smaller,bachem2017practical,sener2018active,chen2010super} and instance selection~\cite{olvera2010review} aim to select a subset of the entire training dataset, where training on this small subset achieves good performance. Most of such methods do not apply to modern deep learning, but new formulations based on bi-level optimization have shown promising results on applications like continual learning \cite{borsos2020coresets}. Related to coreset, other lines of research aim to understand which training samples are    ``valuable'' for modern machine learning, including measuring single-example accuracy \citep{lapedriza2013all} and counting misclassification rates \citep{toneva2018empirical}. In fact, dataset distillation is a generalization of such ideas, as the distilled data do not need to be realistic or come from the training set.

\section{Method}
\lblsec{method}
\emph{Dataset Distillation} refers to the curation of a small, synthetic training set $\mathcal{D}_\mathsf{syn}$ such that a model trained on this synthetic data will have similar performance on the real test set as a model trained on the large, real training set $\mathcal{D}_\mathsf{real}$. In this section, we describe our method that directly mimics the long-range behavior of real-data training, matching multiple training steps on distilled data to \textit{many} more steps on the real data.

In \refsec{expert}, we discuss how we obtain expert trajectories of networks trained on real datasets. In \refsec{matching}, we describe a new dataset distillation method that explicitly encourages the distilled dataset to induce similar long-range network parameter trajectories as the real dataset, resulting in a synthetically-trained network that performs similarly to a network trained on real data. Finally, \refsec{memory} describes our techniques to reduce memory consumption.

\subsection{Expert Trajectories}
\lblsec{expert}
The core of our method involves using \emph{expert trajectories} $\tau^*$ to guide the distillation of our synthetic dataset. By expert trajectories, we mean the time sequence of parameters $\{\theta^*_t\}_0^T$ obtained during the training of a neural network on the \emph{full, real dataset}. To generate these expert trajectories, we simply train a large number of networks on the real dataset and save their snapshot parameters at every epoch. We call these sequences of parameters ``expert trajectories'' because they represent the theoretical upper bound for the dataset distillation task: the performance of a network trained on the full, real dataset. Similarly, we define student parameters $\hat{\theta}_t$ as the network parameters trained on synthetic images at the training step $t$. Our goal is to distill a dataset that will induce a similar trajectory (given the same starting point) as that induced by the real training set such that we end up with a similar model.

Since these expert trajectories are computed using only real data, we can pre-compute them before distillation. All of our experiments for a given dataset were performed using the same pre-computed set of expert trajectories, allowing for rapid distillation and experimentation.

\begin{algorithm}[t]
    \small
    \caption{Dataset Distillation via Trajectory Matching}
    \label{alg:alg}
    \begin{algorithmic}[1]
        \Require {$\{\tau_i^*\}$: set of expert parameter trajectories trained on $\mathcal{D}_\mathsf{real}$.}
        \Require {$M$: \# of updates between starting and target expert params.}
        \Require {$N$: \# of updates to student network per distillation step.}
        \Require {$\mathcal{A}$: Differentiable augmentation function.}
        \Require {$T^+ < T$: Maximum start epoch.}
        \State {Initialize distilled data $\mathcal{D}_\mathsf{syn} \sim \mathcal{D}_\mathsf{real}$}
        \State {Initialize trainable learning rate $\alpha \coloneqq \alpha_0$ for apply $\mathcal{D}_\mathsf{syn}$}
        \For {\textbf{each} distillation step...}
         \label{step:outer}
            \State {$\triangleright$ Sample expert trajectory: $\tau^* \sim \{\tau^*_i\}$ with $ \tau^* = \{\theta^*_{t}\}_0^T$}
            \State {$\triangleright$ Choose random start epoch, $t \leq T^+$}
            \label{step:time}
            \State {$\triangleright$ Initialize student network with expert params:}
            \State {$\quad\quad \hat{\theta}_t \coloneqq \theta^*_{t}$}
            \For {$n = 0 \to N-1$}
                \State $\triangleright$ Sample a mini-batch of distilled images:
                \State $\quad\quad b_{t+n} \sim \mathcal{D}_\mathsf{syn}$
                \label{step:inner}
                \State $\triangleright$ Update student network w.r.t. classification loss:
                \State $\quad\quad \hat{\theta}_{t+n+1} = \hat{\theta}_{t+n} - \alpha\nabla \ell(\mathcal{A}(b_{t+n}); \hat{\theta}_{t+n})$
            \EndFor
            \State {$\triangleright$ Compute loss between ending student and expert params:}
            \State $\quad\quad\mathcal{L} = \|\hat{\theta}_{t+N} - \theta^*_{t+M}\|_2^2 \;/\; \|\theta^*_{t} - \theta^*_{t+M}\|_2^2$
            \State {$\triangleright$ Update $\mathcal{D}_\mathsf{syn}$ and $\alpha$ with respect to $\mathcal{L}$}
        \EndFor
        \Ensure {distilled data $\mathcal{D}_\mathsf{syn}$ and learning rate $\alpha$}
    \end{algorithmic}
        \lblalg{main}
\end{algorithm}

\subsection{Long-Range Parameter Matching}
\lblsec{matching}
Our distillation process learns from the generated sequences of parameters making up our expert trajectories $\{\theta^*_t\}_0^T$. Unlike previous work, our method directly encourages the long-range training dynamics induced by our synthetic dataset to match those of networks trained on the real data.

At each distillation step, we first sample parameters from one of our expert trajectories at a random timestep $\theta^*_t$ and use these to initialize our student parameters $\hat{\theta}_t \coloneqq \theta^*_t$. Placing an upper bound $T^+$ on $t$ lets us ignore the less informative later parts of the expert trajectories where the parameters do not change much.

With our student network initialized, we then perform $N$ gradient descent updates on the student parameters with respect to the classification loss of the \emph{synthetic} data:
\begin{equation}
    \hat{\theta}_{t+n+1} = \hat{\theta}_{t+n} - \alpha \nabla \ell(\mathcal{A}(\mathcal{D}_\mathsf{syn}); \hat{\theta}_{t+n}),
\end{equation}
where $\mathcal{A}$ is the differentiable augmentation technique~\cite{stylegan2ada,zhao2020image,tran2020towards,diffaug} used in previous work~\cite{dsa}, and $\alpha$ is the (trainable) learning rate used to update the student network. Any data augmentation used during distillation must be differentiable so that we can back-propagate through the augmentation layer to our synthetic data. Our method does not use differentiable \textit{Siamese} augmentation since there is no real data used during the distillation process; we are only applying the augmentations to synthetic data at this time. However, we do use the same types of differentiable augmentations on real data during the generation of the expert trajectories.

From this point, we return to our expert trajectory and retrieve the expert parameters from $M$ training updates after those used to initialize the student network $\theta^*_{t+M}$. Finally, we update our distilled images according to the weight matching loss: i.e., the normalized squared $L_2$ error between the updated student parameters $\hat{\theta}_{t+N}$ and the known future expert parameters $\theta^*_{t+M}$:
\begin{equation}
    \mathcal{L} = \frac{\|\hat{\theta}_{t+N} - \theta^*_{t+M}\|^2_2}{\|\theta^*_{t} - \theta^*_{t+M}\|_2^2}, %
    \lbleq{weight_matching}
\end{equation}%
where we normalize the $L_2$ error by the expert distance traveled so that we still get a strong signal from later training epochs where the expert does not move as much. This normalization also helps self-calibrate the magnitude difference across neurons and layers. We have also experimented with other choices of loss functions such as a cosine distance, but find our simple $L_2$ loss works better empirically. We also tried to match the network's output logits between expert trajectory and student network but did not see a clear improvement. We speculate that backpropagating from the network output to the weights introduces additional optimization difficulty.  %

We then minimize this objective to update the pixels of our distilled dataset, along with our trainable learning rate $\alpha$, by back-propagating through all $N$ updates to the student network. The optimization of trainable learning rate $\alpha$ serves as automatic adjusting for the number of student and expert updates (hyperparameters $M$ and $N$).  We use SGD with momentum to optimize $\mathcal{D}_\mathsf{syn}$ and $\alpha$ with respect to the above objective. \refalg{main} illustrates our main algorithm. 
\begin{table*}[!h]
\renewcommand\arraystretch{0.9}
\centering
\scriptsize
\setlength{\tabcolsep}{2pt}
\resizebox{0.98\linewidth}{!}{
\begin{tabular}{ccc|rrr|rrrrrrcr|c}
\toprule
\multirow{3}{*}{}           & \multirow{2}{*}{Img/Cls} & \multirow{2}{*}{Ratio \%} & \multicolumn{3}{c|}{Coreset Selection}   & \multicolumn{8}{c|}{Training Set Synthesis} & \multirow{2}{*}{Full Dataset} \\ %
                            & &                & \multicolumn{1}{c}{Random}        & \multicolumn{1}{c}{Herding}       & \multicolumn{1}{c|}{Forgetting}     & \multicolumn{1}{c}{\texttt{DD}$^\dagger$\cite{dd}}  & \multicolumn{1}{c}{\texttt{LD}$^\dagger$\cite{bohdal2020flexible}} & \multicolumn{1}{c}{\texttt{DC} \cite{dc}}             & \multicolumn{1}{c}{\texttt{DSA} \cite{dsa}} 					& \multicolumn{1}{c}{\texttt{DM} \cite{dm}}         & \multicolumn{1}{c}{\texttt{CAFE} \cite{wang2022cafe}}         &\multicolumn{1}{c}{\texttt{CAFE+DSA} \cite{wang2022cafe}}         & \multicolumn{1}{c|}{Ours} &    \\ \midrule

\multirow{3}{*}{CIFAR-10}        & 1   & 0.02   & 14.4 $\pm$ 2.0  & 21.5 $\pm$ 1.2  & 13.5 $\pm$ 1.2   & \multicolumn{1}{c}{-}             & 25.7 $\pm$ 0.7  & 28.3 $\pm$ 0.5 & 28.8 $\pm$ 0.7 & 26.0 $\pm$ 0.8 & 30.3 $\pm$ 1.1 & 31.6 $\pm$ 0.8 & \bf{46.3 $\pm$ 0.8\anote\;} & \multirow{3}{*}{84.8 $\pm$ 0.1} \\
                                & 10  & 0.2    & 26.0 $\pm$ 1.2  & 31.6 $\pm$ 0.7  & 23.3 $\pm$ 1.0   & 36.8 $\pm$ 1.2  & 38.3 $\pm$ 0.4  & 44.9 $\pm$ 0.5  & 52.1 $\pm$ 0.5  	& 48.9 $\pm$ 0.6 & 46.3 $\pm$ 0.6 & 50.9 $\pm$ 0.5 & \bf{65.3 $\pm$ 0.7\anote\;}          \\  
                                & 50  & 1      & 43.4 $\pm$ 1.0  & 40.4 $\pm$ 0.6  & 23.3 $\pm$ 1.1   & \multicolumn{1}{c}{-}             & 42.5 $\pm$ 0.4  & 53.9 $\pm$ 0.5  & 60.6 $\pm$ 0.5          & 63.0 $\pm$ 0.4 & 55.5 $\pm$ 0.6 & 62.3 $\pm$ 0.4 & \bf{71.6 $\pm$ 0.2\;}&  \\ \midrule
                                
\multirow{3}{*}{CIFAR-100}     & 1   & 0.2    &  4.2 $\pm$ 0.3  &  8.4 $\pm$ 0.3  &  4.5 $\pm$ 0.2   & \multicolumn{1}{c}{-}             & 11.5 $\pm$ 0.4  & 12.8 $\pm$ 0.3  & 13.9 $\pm$ 0.3    		& 11.4 $\pm$ 0.3 &12.9 $\pm$ 0.3 & 14.0 $\pm$ 0.3 &  \bf{24.3 $\pm$ 0.3\anote\;} & \multirow{3}{*}{56.2 $\pm$ 0.3}\\ 
                              & 10  & 2      & 14.6 $\pm$ 0.5  & 17.3 $\pm$ 0.3  & 15.1 $\pm$ 0.3   & \multicolumn{1}{c}{-}             & \multicolumn{1}{c}{-}             & 25.2 $\pm$ 0.3  & 32.3 $\pm$ 0.3    		& 29.7 $\pm$ 0.3  & 27.8 $\pm$ 0.3 & 31.5 $\pm$ 0.2 &   \bf{40.1 $\pm$ 0.4\;}           \\  
                              & 50  & 10     & 30.0 $\pm$ 0.4  & 33.7 $\pm$ 0.5  & 30.5 $\pm$ 0.3   & \multicolumn{1}{c}{-}             & \multicolumn{1}{c}{-}             & \multicolumn{1}{c}{-}             &   42.8 $\pm$ 0.4                    & 43.6 $\pm$ 0.4  & 37.9 $\pm$ 0.3 & 42.9 $\pm$ 0.2 &      \bf{47.7 $\pm$  0.2\anote\;}        \\  \midrule

\multirow{3}{*}{Tiny ImageNet} & 1   & 0.2    &  1.4 $\pm$ 0.1  &  2.8 $\pm$ 0.2  &  1.6 $\pm$ 0.1   & \multicolumn{1}{c}{-}             & \multicolumn{1}{c}{-}               & \multicolumn{1}{c}{-}             & \multicolumn{1}{c}{-}    		            &  3.9 $\pm$ 0.2  & \multicolumn{1}{c}{-} & \multicolumn{1}{c}{-} &\bf{8.8 $\pm$ 0.3\;}  & \multirow{3}{*}{37.6 $\pm$ 0.4}\\ 
                              & 10  & 2      &  5.0 $\pm$ 0.2  &  6.3 $\pm$ 0.2  &  5.1 $\pm$ 0.2   & \multicolumn{1}{c}{-}             & \multicolumn{1}{c}{-}               & \multicolumn{1}{c}{-}             & \multicolumn{1}{c}{-}  		            & 12.9 $\pm$ 0.4  &          \multicolumn{1}{c}{-} & \multicolumn{1}{c}{-} &\bf{23.2 $\pm$ 0.2\;}      \\  
                              & 50  & 10     & 15.0 $\pm$ 0.4  & 16.7 $\pm$ 0.3  & 15.0 $\pm$ 0.3   & \multicolumn{1}{c}{-}             & \multicolumn{1}{c}{-}               & \multicolumn{1}{c}{-}             &   \multicolumn{1}{c}{-}                   &24.1 $\pm$ 0.3  &         \multicolumn{1}{c}{-} & \multicolumn{1}{c}{-} &\bf{28.0 $\pm$ 0.3\;}       \\  \bottomrule

\end{tabular}
}
\vspace{-7pt}
\caption{Comparing distillation and coreset selection methods. As in previous work, we distill the given number of images per class using the training set, train a neural network on the synthetic set, and evaluate on the test set. To get $\Bar{x}\pm s$, we train 5 networks from scratch on the distilled dataset. Note that the earlier works \texttt{DD}$^\dagger$ and \texttt{LD}$^\dagger$ use different architectures, \ie LeNet \citep{lecun1998gradient} for MNIST and AlexNet \citep{alexnet} for CIFAR-10. All others use a 128-width ConvNet. CIFAR values marked by ($*$) signify best results were obtained with ZCA whitening.}
\lbltbl{sota}
     \vspace{-6pt}
\end{table*}

\subsection{Memory Constraints}
\lblsec{memory}
Given that we are back-propagating through many gradient descent updates, memory consumption quickly becomes an issue when our distilled dataset is sufficiently large, as we have to jointly optimize all the images of all the classes at each optimization step. To reduce memory consumption and ease the learning problem, previous methods distill one class at a time~\cite{dc, dsa, dm}, but this may not be an ideal strategy for our method since the expert trajectories are trained on all classes simultaneously.

We could potentially circumvent this memory constraint by sampling a new mini-batch at every distillation step (the outer loop in Line \ref{step:outer} of \refalg{main}). 
Unfortunately,  this comes with its own issues, as redundant information could be distilled into multiple images across the synthetic dataset, degrading to catastrophic mode collapse in the worst case.

Instead, we can sample a new mini-batch $b$ for every update of the \emph{student network} (i.e., the inner loop in  Line \ref{step:inner} of \refalg{main}) such that all distilled images will have been seen by the time the final weight matching loss (\refeq{weight_matching}) is calculated. The mini-batch $b$ still contains images from different classes but has much fewer images per class. In this case, our student network update then becomes%
\vspace{-5pt}\begin{equation}
\begin{split}
      & b_{t+n} \sim \mathcal{D}_\mathsf{syn}\\
    & \hat{\theta}_{t+n+1} =\hat{\theta}_{t+n} - \alpha \nabla \ell(\mathcal{A}(b_{t+n}); \hat{\theta}_{t+n}).
\end{split}\vspace{-10pt}
\end{equation}%
This method of batching allows us to distill a much larger synthetic dataset while ensuring some amount of heterogeneity among the distilled images of the same class.

\section{Experiments}
\lblsec{expr}
We evaluate our method on various datasets, including \begin{itemize}[topsep=1pt, partopsep=9pt, itemsep=-1pt, parsep=0.5ex]
    \item $32\times32$ CIFAR-10 and CIFAR-100 (\refsec{low-res}), two commonly used datasets in dataset distillation literature,
    \item $64\times 64$ Tiny ImageNet (\refsec{tiny}), a recent benchmark by the concurrent work~\cite{dm}, and 
    \item our new $128 \times 128$ ImageNet subsets (\refsec{imagenet}).
\end{itemize}
We provide additional visualizations and ablation studies in the supplementary material.

\myparagraph{Evaluation and Baselines.} %
We evaluate various methods according to the standard protocol: training a randomly initialized neural network from scratch on \emph{distilled} data and evaluating on the validation set. %

To generate the distilled images for our method, we employ the distillation process detailed in the previous section and Algorithm \ref{alg:alg}, using the same suite of differentiable augmentations as done in previous work~\cite{dsa,dm}. The hyperparameters used for each setting (real epochs per iteration, synthetic updates per iteration, image learning rate, etc.) can be found in the supplemental material.

We compare to several recent methods including Dataset Distillation~\cite{dd} (\texttt{DD}), Flexible Dataset Distillation ~\cite{bohdal2020flexible} (\texttt{LD}), Dataset Condensation~\cite{dc} (\texttt{DC}), and Differentiable Siamese Augmentation~\cite{dsa} (\texttt{DSA}), along with a method based on the infnite-width kernel limit~\citep{nguyen2020dataset,nguyen2021dataset} (\texttt{KIP}) and concurrent works Distribution Matching~\cite{dm} (\texttt{DM}) and Aligning Features~\cite{wang2022cafe} (\texttt{CAFE}). We also compare our methods with instance selection algorithms including random selection (\texttt{random}), herding methods~\cite{chen2010super} (\texttt{herding}), and example forgetting~\cite{toneva2018empirical} (\texttt{forgetting}). 

\myparagraph{Network Architectures.}
Staying with precedent~\cite{dc, dsa, dm, nguyen2021dataset}, we mainly employ a simple ConvNet architecture designed by Gidaris and Komodakis~\cite{gidaris2018dynamic} for our distillation tasks. The architecture consists of several convolutional blocks, each containing a
$3\times 3$ convolution layer with 128 filters, Instance normalization~\cite{ulyanov2016instance}, RELU, and $2\times2$ average pooling with stride 2. After the convoluation blocks, a single linear layer produces the logits. The exact number of such blocks is decided by the dataset resolution and is specified below for each dataset.
Staying with this simple architecture allows us to directly analyze the effectiveness of our core method and remain comparable with previous works.

\subsection{Low-Resolution Data (32$\times$32)}
\lblsec{low-res}

For low-resolution tasks, we begin with the 32$\times$32 CIFAR-10 and CIFAR-100 datasets~\cite{CIFAR10}. For these datasets, we employ ZCA whitening as done in previous work~\cite{nguyen2020dataset, nguyen2021dataset}, using the Kornia \cite{kornia} implementation with default parameters.
Staying with precedent, we use a depth-3 ConvNet taken directly from the open-source code~\cite{dc, dsa}.

As seen in \reftbl{sota}, our method significantly outperforms all baselines in every setting. In fact, on the one image per class setting, we  improve the next best method (\texttt{DSA} \cite{dsa}) to almost twice test accuracy, on both datasets. For CIFAR-10, these distilled images can be seen in \reffig{CIFAR-10}.  CIFAR-100 images are visualized in the supplementary material

In  \reftbl{transfer_nn}, we also compare with a recent method \texttt{KIP}~\cite{nguyen2020dataset,nguyen2021dataset}, where the distilled data is learned with respect to the Neural Tangent Kernel. Because \texttt{KIP} training is agnostic to actual network width, we test their result on both a ConvNet of the same width as us and other methods (128) and a ConvNet of larger width (1024) (which is shown in \texttt{KIP} paper \citep{nguyen2021dataset}).  Based on the the infinite-width network limit, \texttt{KIP} may exhibit a gap with practical finite-width networks. Our method does not suffer from this limitation and generally achieves better performance.  In all settings, our method, trained on the 128-width network, outperforms \texttt{KIP} results evaluated on both widths, except for just one setting where \texttt{KIP} is applied on the much wider 1024-width network. 

As noted in the previous methods \citep{dd}, we also see significant diminishing returns when allowing more images in our synthetic datasets. For instance, on CIFAR-10, we see an increase from 46.3\% to 65.3\% classification accuracy when increasing from 1 to 10 images per class, but only an increase from 65.3\% to 71.5\% when increasing the number of distilled images per class from 10 to 50.

If we look at the one image per class visualizations in \reffig{CIFAR-10} (top), we see very abstract, yet still recognizable, representations of each class. When we limit the task to just one synthetic image per class, the optimization is forced to squeeze as much of the class's distinguishing information as possible into just one sample. When we allow more images in which to disperse the class's information, the optimization has the freedom to spread the class's discriminative features among the multiple samples, resulting in a diverse set of structured images we see in \reffig{CIFAR-10} (bottom) (e.g., different types of cars and horses with different poses).

\begin{table}
\centering\vspace{-8pt}
\resizebox{0.95\linewidth}{!}{
\begin{tabular}{@{}ccc|c|cc}
\toprule
 & Img/Cls & Ratio \% & \makecell{KIP to NN \\ (1024-width)}&\makecell{KIP to NN\\ (128-width)}& \makecell{Ours \\ (128-width)} \\ \midrule 
\multirow{3}{*}{CIFAR-10}  & 1 & 0.02 & 49.9 & 38.3 & \textbf{46.3} \\
                          & 10 & 0.1 & 62.7 &  57.6&\bf{65.3 }\\
                          & 50 & 1 & 68.6 & 65.8 &\bf{71.5 }\\ \midrule
\multirow{2}{*}{CIFAR-100} & 1  & 0.2 & 15.7 & 18.2&\bf{24.3 }\\
                          & 10 & 2 & 28.3 & 32.8&\bf{39.4}\\
\bottomrule
\end{tabular}
}
\vspace{-7pt}
\caption{Kernel Inducing Point (\texttt{KIP})~\cite{nguyen2021dataset} performs distillation using the infinite-width network limit. We consistently outperform \texttt{KIP} when evaluating on the same finite-width network, and almost always outperform \texttt{KIP} applied on wider networks. }
\lbltbl{transfer_nn}
\vspace{-8pt}
\end{table}

\begin{table}
    \centering
    \smaller
    \resizebox{0.9\linewidth}{!}{
    \begin{tabular}{cc|cccc}
        \toprule& &\multicolumn{4}{c}{Evaluation Model}\\
        & & ConvNet & ResNet & VGG & AlexNet  \\
        \midrule \multirow{3}{*}{\rotatebox[origin=c]{90}{Method}} &Ours & \textbf{64.3 $\pm$ 0.7} & \textbf{46.4 $\pm$ 0.6} & \textbf{50.3 $\pm$ 0.8} & \textbf{34.2 $\pm$ 2.6}\\
        &DSA & 52.1 $\pm$ 0.4 & 42.8 $\pm$ 1.0 & 43.2 $\pm$ 0.5 &  \textbf{35.9 $\pm$ 1.3}\\
        &KIP & 47.6 $\pm$ 0.9 & 36.8 $\pm$ 1.0&  42.1 $\pm$ 0.4& 24.4 $\pm$ 3.9 \\\bottomrule
    \end{tabular}
    }
\vspace{-6pt}
    \caption{Despite being trained for a specific architecture, our synthetic images do not seem to suffer from much over-fitting to that model. This is evaluated on CIFAR-10 with 10 images per class.}
    \lbltbl{cross}
\vspace{-7pt}
\end{table}

\myparagraph{Cross-Architecture Generalization.}
We also evaluate how well our synthetic data performs on architectures different from the one used to distill it on the CIFAR-10, 1 image per class task. In \reftbl{cross}, we show our baseline \linebreak ConvNet performance and evaluate on ResNet~\cite{resnet}, VGG~\cite{vgg}, and AlexNet~\cite{alexnet}. 
 
For \texttt{KIP}, instead of the Kornia \cite{kornia} ZCA implementation, we use the authors' custom ZCA implementation for evaluation of their method. Our method is solidly the top performer on all the transfer models except for AlexNet where we lie within one standard deviation of \texttt{DSA}. This could be attributed to our higher baseline performance, but it still shows that our method is robust to changes in architecture.

\begin{figure}
    \centering
    \vspace{-6pt}
    \includegraphics[width=0.88\linewidth]{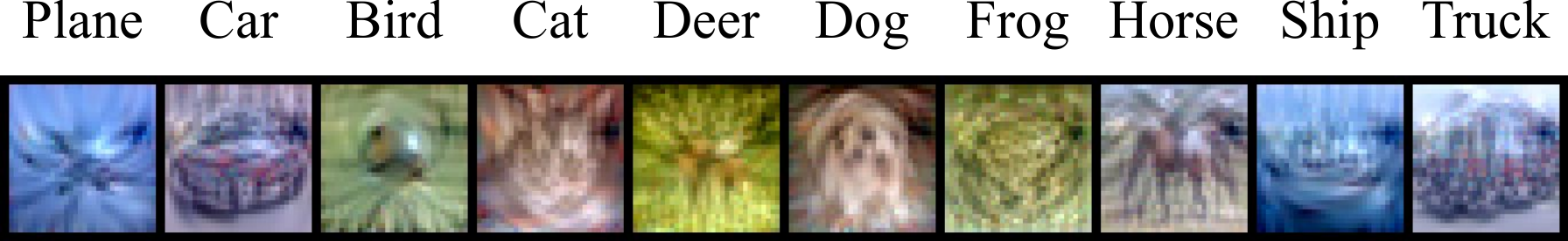}\\[-0.6ex]
    {\smaller 1 image per class} \vspace{0.5cm}\\
        \vspace{-0.34cm}

    \includegraphics[width=0.88\linewidth, trim=0 0 0 20, clip]{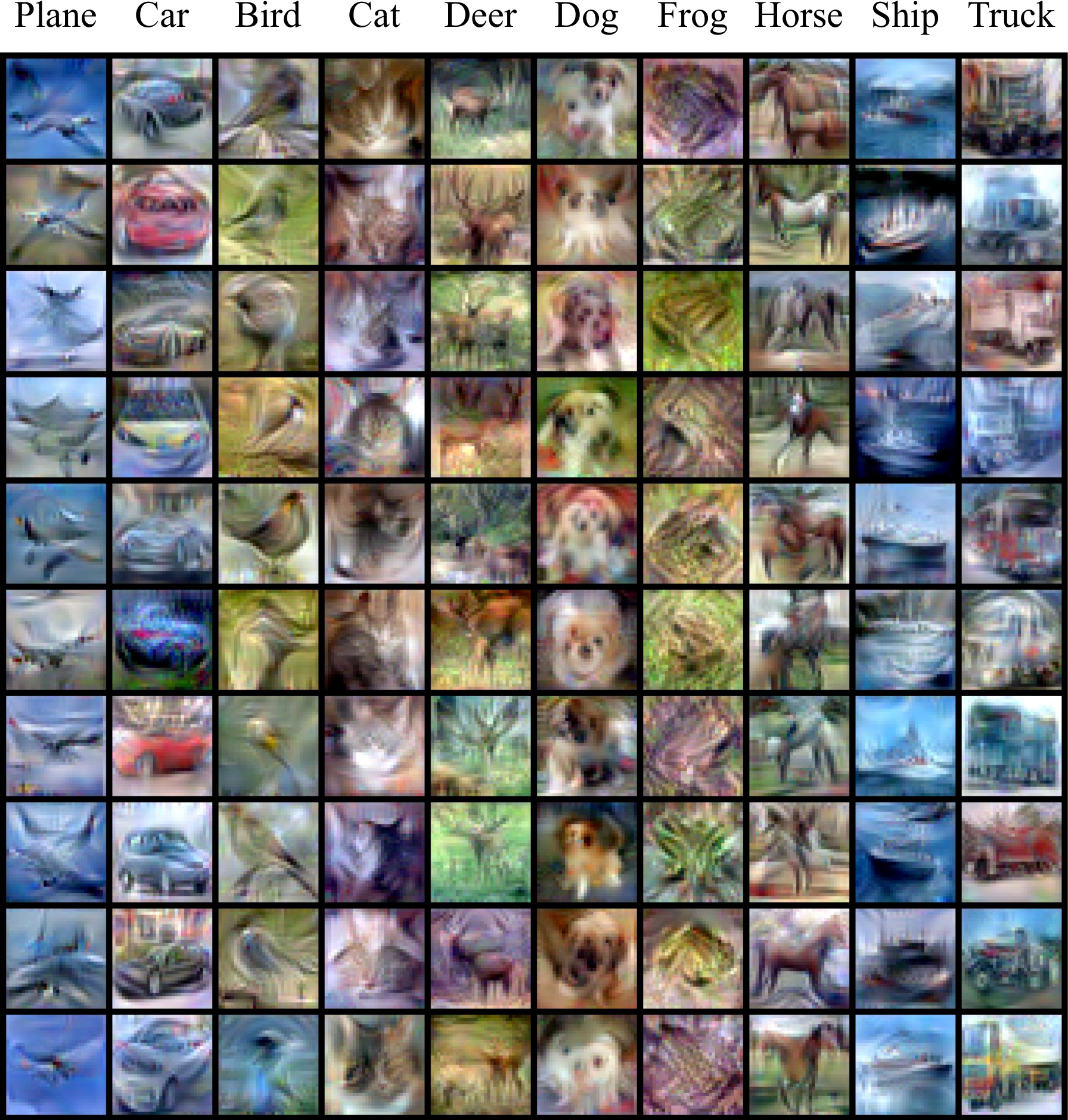}\\[-0.6ex]
    {\smaller 10 images per class}
         \vspace{-7pt}
    \caption{CIFAR-10: The 1 image per class images are more abstract but also more information-dense while the 10 images per class images are more expressive and contain more structure.}
    \lblfig{CIFAR-10}
         \vspace{-7pt}
\end{figure}

\subsection{Short-Range vs.~Long-Range Matching}
\lblsec{shortlongrange}

Unlike some prior works (\texttt{DC} and \texttt{DSA}), our method performs long-range parameter matching, where $N$ training steps on distilled data match a much larger $M$ steps on real data. Methods that optimize over entire training processes (\eg \texttt{DD} and \texttt{KIP}) can be viewed as even longer range matching. However, their performances fall short of our method (\eg in \reftbl{sota}), likely due to related instabilities or inexact approximations. Here, we experimentally confirm our hypothesis that long-range matching achieved by larger $M$ and $N$ in our method is superior to the short-range counterparts (such as small $M$ and $N$ and \texttt{DSA}).

In \reffig{mn-alpha} (left), we evaluate our method on different settings of $M$ and $N$. Really short-range matching (with $N=1$ and small $M$) generally exhibits worse performance than long-range matching, with the best performance attained when both $N$ and $M$ are relatively large. Furthermore, as we increase $N$, the power of  $N$ combined steps (on distilled data) becomes stronger and can approximate longer-range behavior, leading to the optimal $M$ values shifting to greater values correspondingly.

\begin{figure}
\vspace{-4pt}
    \begingroup
    \setlength{\tabcolsep}{1pt}
    \centering
    \begin{tabular}{@{\hskip 0pt}cc@{\hskip 0pt}cc@{\hskip 0pt}}
         \rotatebox[origin=c]{90}{\fontsize{6}{5}\selectfont{Validation Acc. \%}} &  \includegraphics[align=c,width=0.455\linewidth,trim=20 0 0 0,clip]{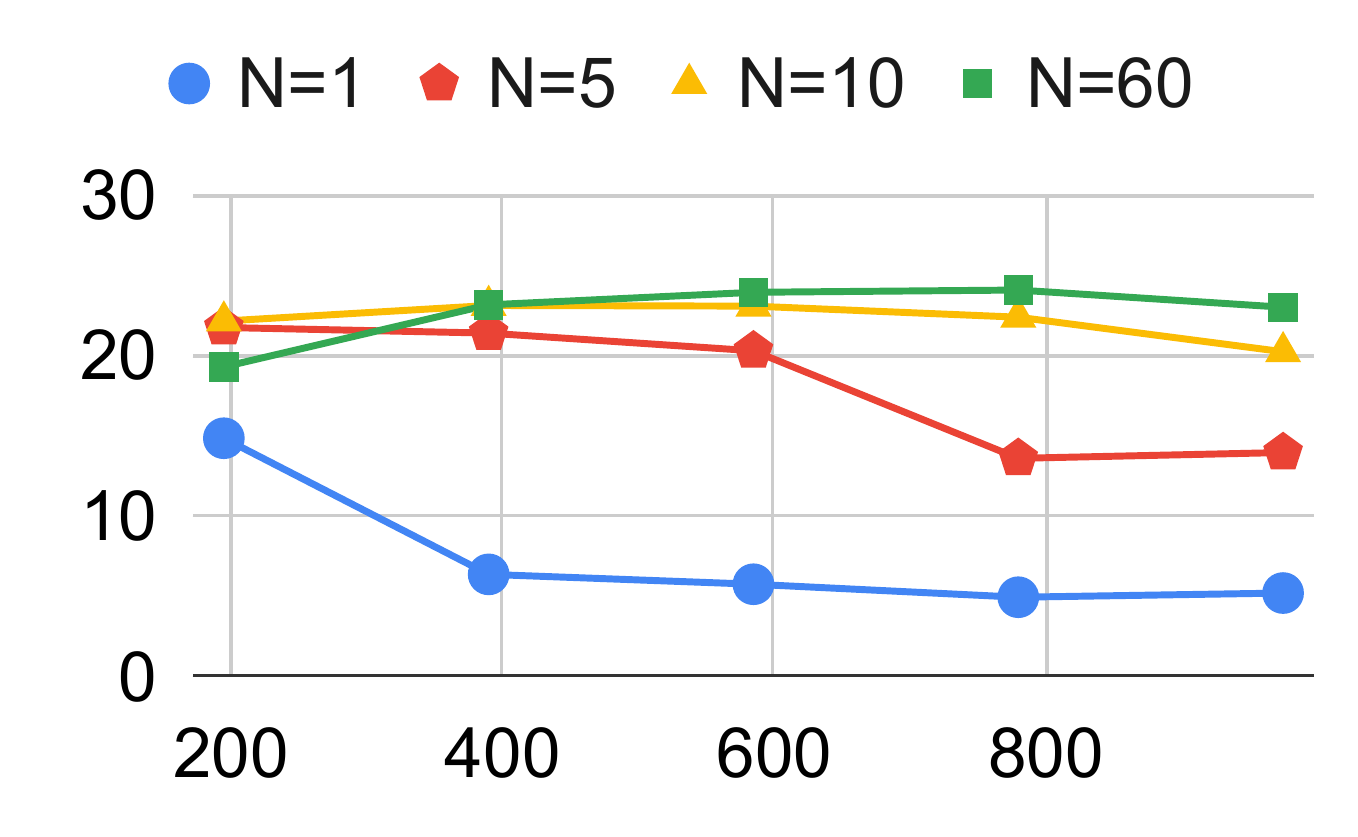}& \hfill\;\;\rotatebox[origin=c]{90}{\fontsize{6}{5}\selectfont{Param. Distance to Target}} &  \includegraphics[align=c,width=0.458\linewidth,trim=13 0 9 0,clip]{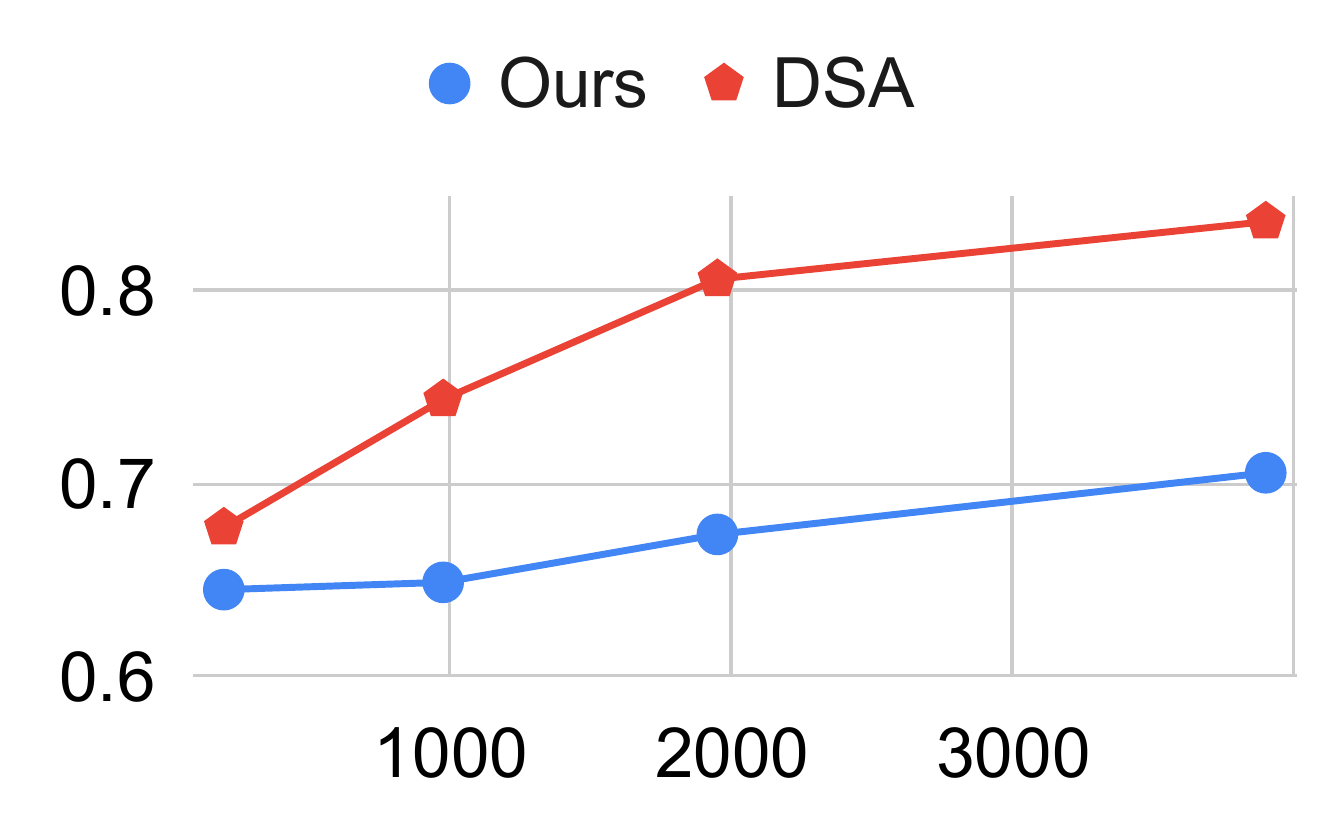}\\[-1.3ex]
         & \fontsize{6}{5}\selectfont{\;\;\;\;\;$M$: Expert Steps} & & \fontsize{6}{5}\selectfont{\;\;$\Delta t$: Expert Steps between Init.~and Target}
    \end{tabular}
    \endgroup
        \vspace{-9pt}
    \caption{CIFAR-100 (1 image / class). \textbf{Left:} Smaller $M$ and $N$ match shorter-range behavior, which performs worse than longer-range matching. 
    \textbf{Right:} Over $1000$ training steps on distilled data, we track the closest distance in parameter space (normalized MSE in \refeq{weight_matching}) to a target set of parameters, obtained with $\Delta_t$ training steps on real data. Matching long-range behavior, our method better approximate real data training for longer ranges (large $\Delta_t$).
    }
    \lblfig{mn-alpha}
    \vspace{-5pt}
\end{figure}

In  \reffig{mn-alpha} (right), we evaluate our method and a short-range matching work (\texttt{DSA}) on their abilities to approximate real training behavior over short and long ranges. Starting from a set of initial parameters, we set the target parameters to be the result of $\Delta_t$ training steps on real data (\ie the long-range behavior that distilled data should mimic). A small (or large) $\Delta_t$ means evaluating matching over a short (or long) range. For both methods, we test how close they can train the network (using distilled data) from the same initial parameters to the target parameters. \texttt{DSA} is only optimized to match short-range behavior, and thus errors may accumulate during longer training. Indeed, as $\Delta_t$ grows larger, \texttt{DSA} fails to mimic the real data behavior over longer ranges. In comparison, our method is optimized for long-range matching and thus performs much better.

\begin{table}[]
\centering
\resizebox{1\linewidth}{!}{
\begin{tabular}{@{\hskip 2pt}c@{\hskip 4pt}|@{\hskip 5pt}c@{\hskip 5pt}c@{\hskip 5pt}c@{\hskip 5pt}c@{\hskip 5pt}c@{\hskip 5pt}c@{\hskip 0pt}}
\toprule
 & ImageNette & ImageWoof & ImageFruit & ImageMeow & ImageSquawk & ImageYellow \\ \midrule 

1 Img/Cls & 47.7 $\pm$ 0.9 & 28.6 $\pm$ 0.8 & 26.6 $\pm$ 0.8 & 30.7 $\pm $1.6 & 39.4 $\pm$ 1.5 & 45.2 $\pm$ 0.8\\
10  Img/Cls & 63.0 $\pm$ 1.3 & 35.8 $\pm$ 1.8 & 40.3 $\pm$ 1.3 & 40.4 $\pm$ 2.2 & 52.3 $\pm$ 1.0 & 60.0 $\pm$ 1.5\\
                            \midrule
Full Dataset & 87.4 $\pm$ 1.0 & 67.0 $\pm$ 1.3 & 63.9 $\pm$ 2.0 & 66.7 $\pm$ 1.1 & 87.5 $\pm$ 0.3 & 84.4 $\pm$ 0.6\\\bottomrule
\end{tabular}
}
\vspace{-8pt}
\caption{Applying our method to 128$\times$128 resolution ImageNet subsets. On this higher resolution, across various subsets, our method continues to produce high-quality distilled images.}
\lbltbl{imagenet}
     \vspace{-10pt}
\end{table}

\subsection{Tiny ImageNet (64$\times$64)}
\lblsec{tiny}

Introduced to the dataset distillation task by the concurrent work, Distribution Matching (\texttt{DM})~\cite{dm}, we also show the effectiveness of our algorithm on the 200 class, 64$\times$64 Tiny ImageNet~\cite{tiny} dataset (a downscaled subset of ImageNet \cite{deng2009imagenet}). %
To account for the higher image resolution, we move up to a depth-4 ConvNet, similar to \texttt{DM}~\cite{dm}.

Most dataset distillation methods (other than \texttt{DM}) are unable to handle this larger resolution due to extensive memory or time requirement, as the \texttt{DM} authors also observed \citep{dm}. In \reftbl{sota}, our method consistently outperforms the only viable such baseline, \texttt{DM}. Notably, on the 10 images per class task, our method improves the concurrent work \texttt{DM} from  12.9\% and 23.2\%. A subset of our results is shown in \reffig{tinyimagenet}. The supplementary material contains the rest of the images. %

At 200 classes and 64$\times$64 resolution, Tiny ImageNet certainly poses a much harder task than previous datasets. Despite this, many of our distilled images are still recognizable, with a clear color, texture, or shape pattern.
\begin{figure}[t]
    \centering
    \includegraphics[width=\linewidth]{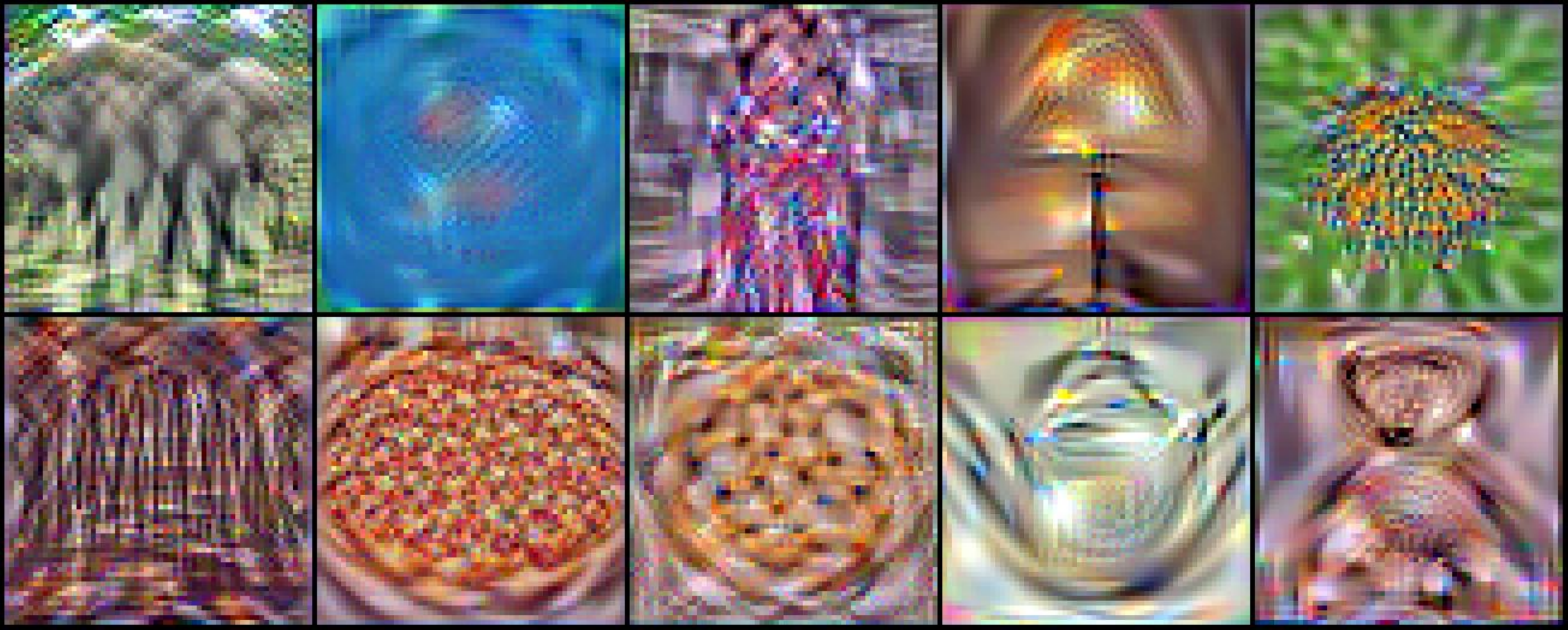}
         \vspace{-7pt}
    \caption{Selected samples distilled from Tiny ImageNet, one image per class. Despite the higher resolution, our method still produces high-fidelity images. (Can you guess which classes these images represent? Check your answers in the footnote!\protect\footnotemark)}
    \lblfig{tinyimagenet}%
    \vspace{-7pt}
\end{figure}

\footnotetext{\smaller{Answers for \reffig{tinyimagenet}: \textbf{First Row:} African Elephant, Jellyfish, Kimono, Lampshade, Monarch} \textbf{Second Row:} Organ, Pizza, Pretzel, Teapot, Teddy}
\subsection{ImageNet Subsets (128$\times$128)}
\lblsec{imagenet}
\newcommand\imagenetwidth{0.45}
\begin{figure*}[ht]
\vspace{-8pt}
\centering
\begingroup
\setlength{\tabcolsep}{2pt}
\begin{tabular}{cccc}
\rotatebox[origin=c]{90}{\small{ImageNette}}    &\includegraphics[align=c,width=\imagenetwidth\linewidth]{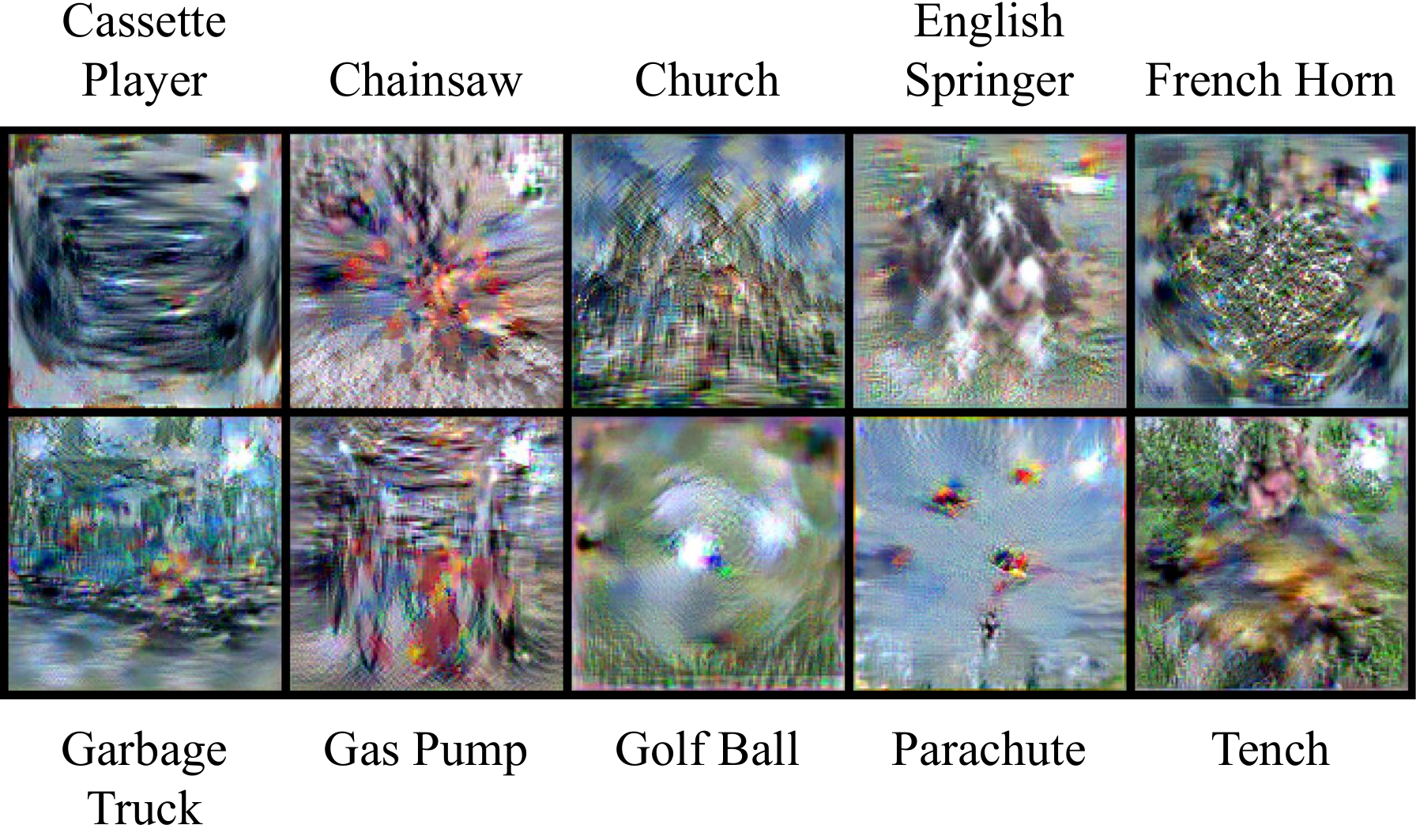}
    &\includegraphics[align=c,width=\imagenetwidth\linewidth]{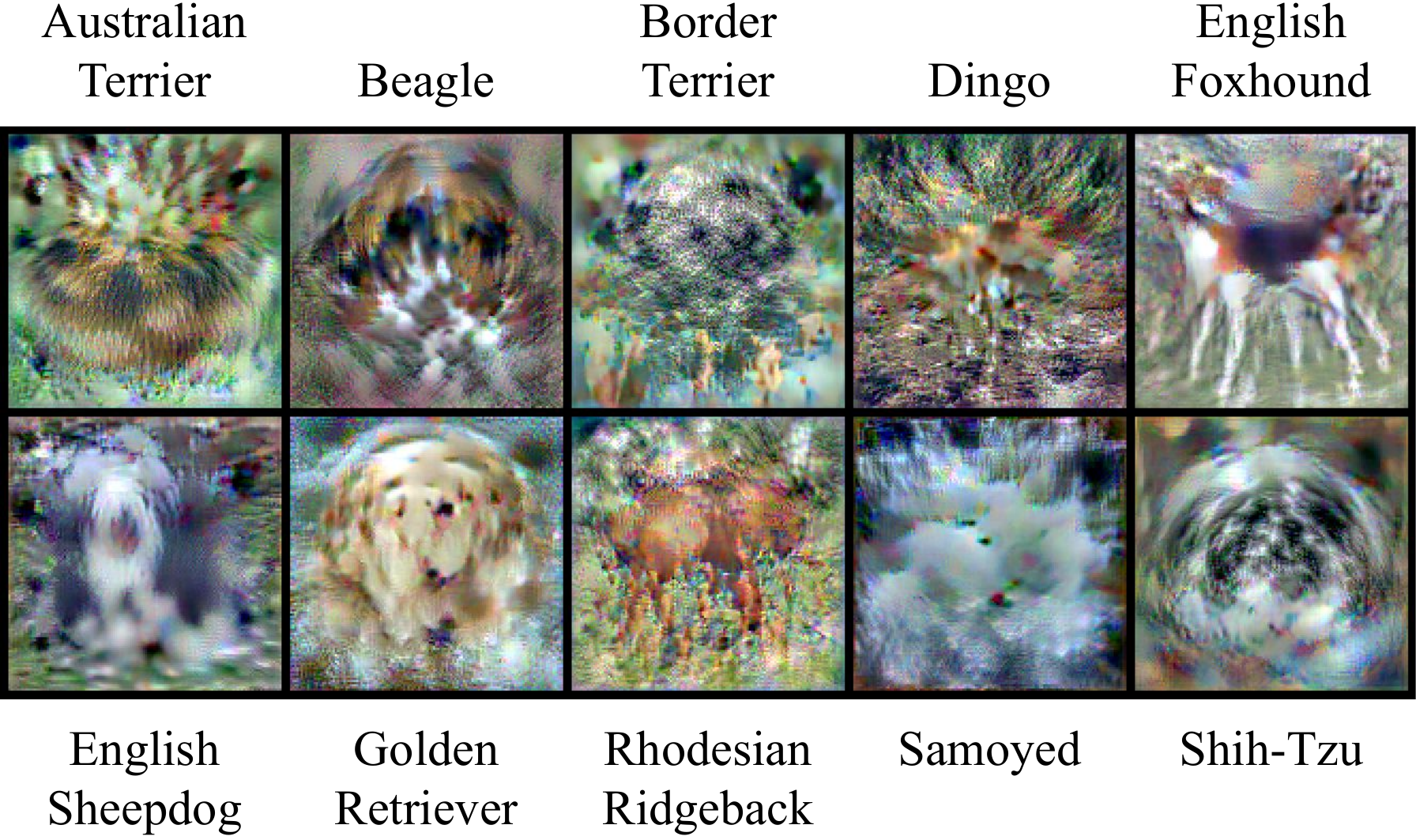} & \rotatebox[origin=c]{270}{\small{ImageWoof}} \\[-0.8ex]
\rotatebox[origin=c]{90}{\small{ImageSquawk}}    &\includegraphics[align=c,width=\imagenetwidth\linewidth]{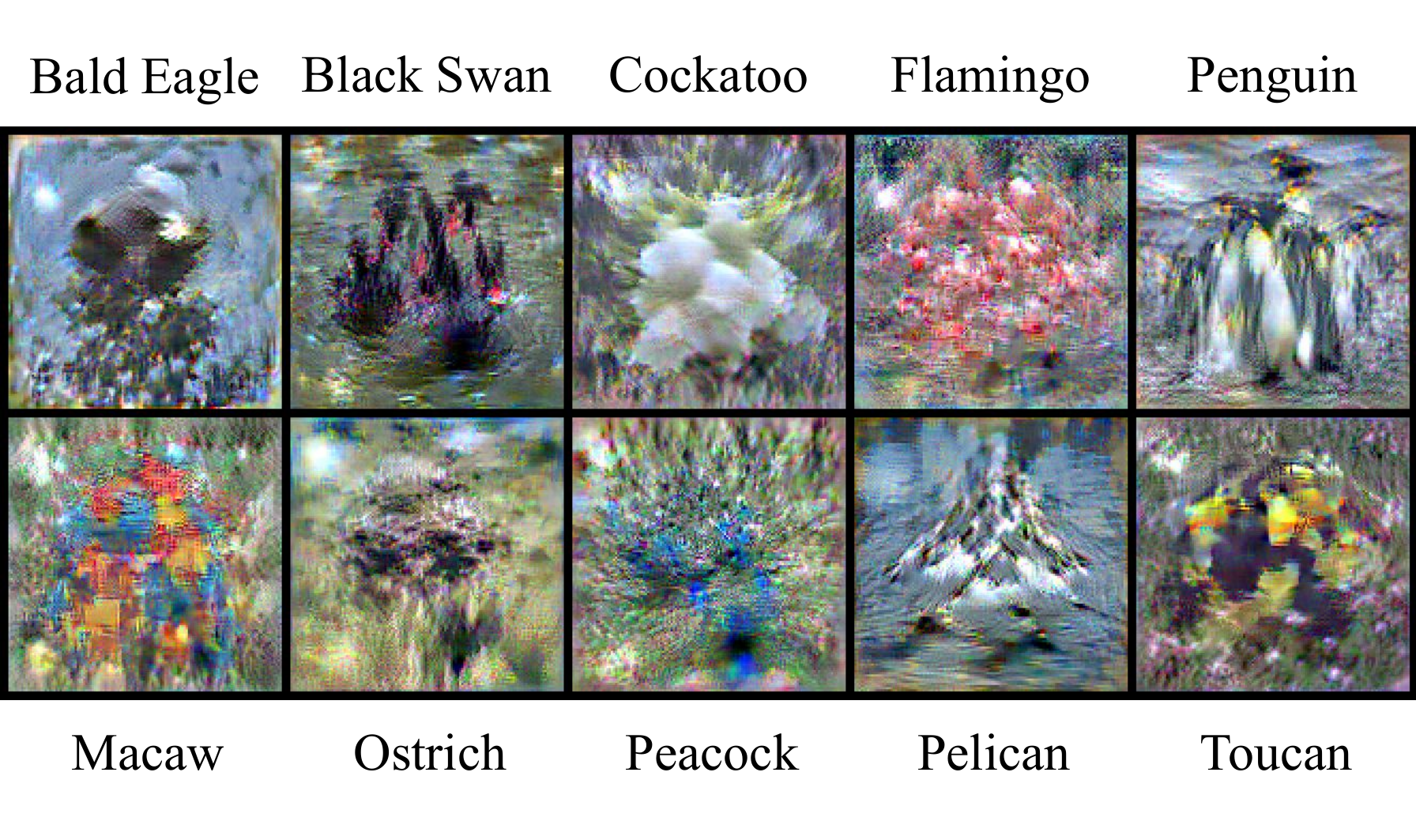}
    &\includegraphics[align=c,width=\imagenetwidth\linewidth]{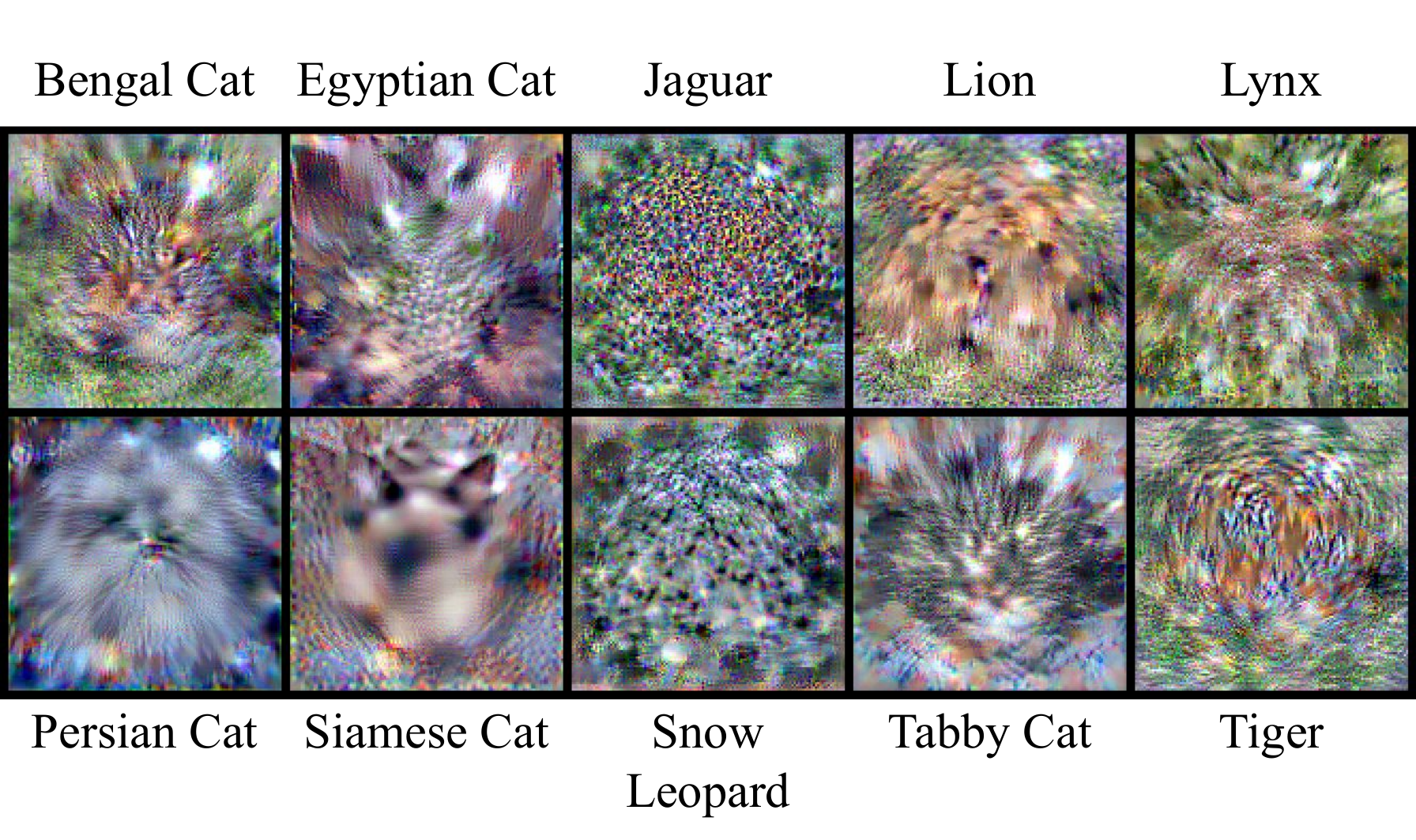} & \rotatebox[origin=c]{270}{\small{ImageMeow}} \\[-0.8ex]
\rotatebox[origin=c]{90}{\small{ImageFruit}}    &\includegraphics[align=c,width=\imagenetwidth\linewidth]{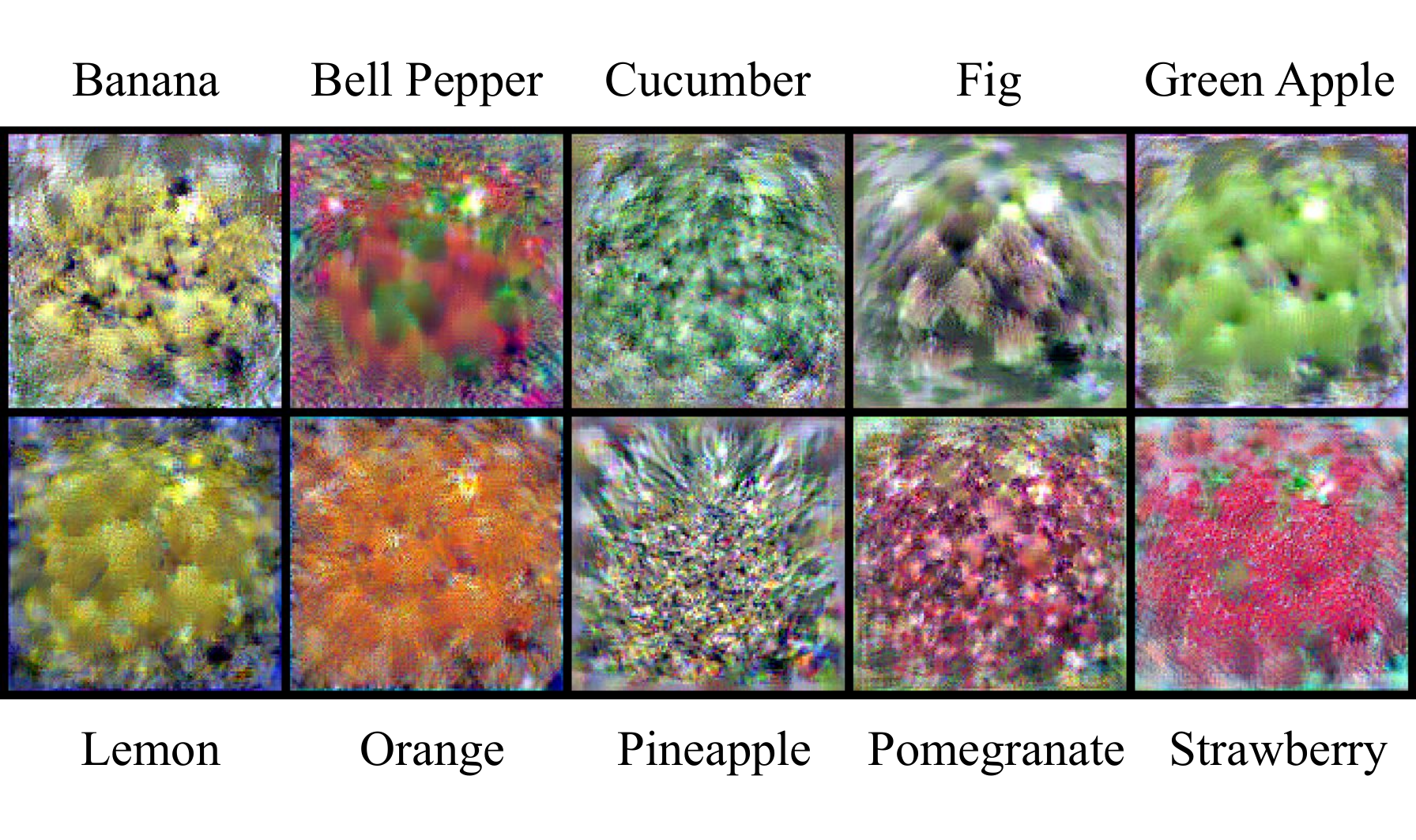}
    &\includegraphics[align=c,width=\imagenetwidth\linewidth]{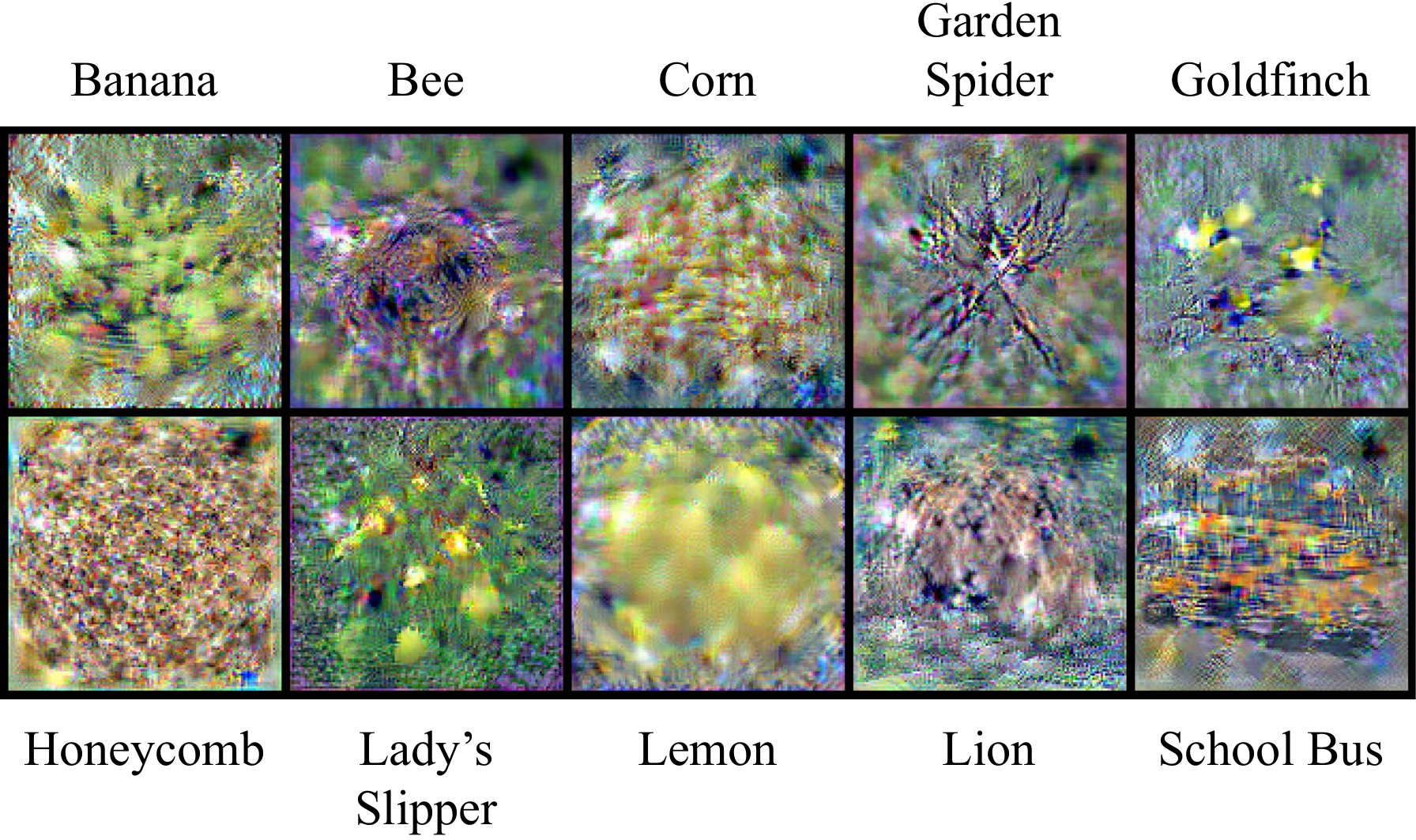} & \rotatebox[origin=c]{270}{\small{ImageYellow}}
\end{tabular}
\endgroup
\vspace{-0.3cm}
    \captionof{figure}{Our method is the first capable of distilling higher-resolution (128$\times$128) images, allowing us to explore the ImageNet \cite{deng2009imagenet} dataset.}
    \lblfig{imagenet}
    \vspace{-6pt}
\end{figure*}

Next, we push the boundaries of dataset distillation even further by running our method on yet higher resolution images in the form of 128$\times$128 subsets of ImageNet~\cite{deng2009imagenet}. Again, due to the higher resolution, we increase the depth of our architecture and use a depth-5 ConvNet for the 128$\times$128 ImageNet subsets.

ImageNette (assorted objects) and ImageWoof (dog breeds) are existing subsets~\cite{imagenette} designed to be easy and hard to learn respectively. We also introduce ImageFruit (fruits), ImageMeow (cats), ImageSquawk (birds), and ImageYellow (yellow-ish things) to further illustrate our algorithm. %

Similar to Tiny ImageNet, most dataset distillation baselines do not scale up to our ImageNet subset settings.
As the code of \texttt{DM}~\cite{dm} is not publicly available now, we choose to only compare to the networks trained on the full dataset. We wish to show that our method transfers well to large images and still produces meaningful results at a higher resolution. Validation set accuracies are presented in \reftbl{imagenet}.

While all of the generated images are devoid of high-frequency noise, the tasks still differ in the type of distilled image they induce. For tasks where all the classes have a similar structure but unique textures like ImageSquawk (\reffig{imagenet}), the distilled images may not have much structure but instead store discriminating information in the textures.

Conversely, for tasks where all classes have similar color or textures like ImageYellow (\reffig{imagenet}), the distilled images seem to diverge from their common trait and accentuate the structure or secondary color that makes them unique. Specifically, note the differences between the distilled ``Banana'' images for the ImageFruit and ImageYellow (bottom row, \reffig{imagenet}). Although the expert trajectory-generating networks saw the same ``Banana'' training images, the distilled images differ between the two tasks. The distilled ``Banana'' for the ImageYellow task is clearly much ``greener'' than the equivalent image for the ImageFruit task. This implies that the expert networks identify different features by which to identify classes based on the other classes in the task.

\section{Discussion and Limitations}
\lblsec{discussion}
In this work, we introduced a dataset distillation algorithm by means of directly optimizing the synthetic data to induce similar network training dynamics as the real data.  The main difference between ours and prior approaches is that we are neither limited to the short-range single-step matching nor subject to instability and compute intensity of optimizing over the full training process. Our method balances these two regimes and shows improvement over both. 

Unlike prior methods, ours is the first to scale to $128\times 128$ ImageNet images, which not only allows us to gain interesting insights of the dataset (\eg in \refsec{imagenet}) but also may serve as an important step towards practical applications of dataset distillation on real-world datasets. 

\myparagraph{Limitations.} Our use of pre-computed trajectories allows for significant memory saving, at the cost of additional disk storage and computational cost for expert model training. The computational overhead of training and storing expert trajectories is
quite high. For example, CIFAR experts took $\sim$3 seconds per epoch (8 GPU hours total for all 200 CIFAR experts) while each ImageNet (subset) expert took $\sim$11 seconds per epoch (15 GPU hours total for all 100 ImageNet experts). Storage-wise, each CIFAR expert took up $\sim$60MB of storage while each ImageNet expert took up $\sim$120MB.

\myparagraph{Acknowledgements.}
We would like to thank Alexander Li, Assaf Shocher,  Gokul Swamy, Kangle Deng, Ruihan Gao, Nupur Kumari, Muyang Li, Garuav Parmar, Chonghyuk Song, Sheng-Yu Wang, and Bingliang Zhang as well as Simon Lucey's Vision Group at the University of Adelaide for their valuable feedback.
This work is supported, in part, by the NSF Graduate Research Fellowship under Grant No. DGE1745016 and grants from J.P. Morgan Chase, IBM, and SAP.

{\small
\bibliographystyle{ieee_fullname}
\bibliography{main}
}
\clearpage

\appendix
\section{Appendix}\subsection{Additional Visualizations}
\begin{figure}
    \centering
    \includegraphics[width=\linewidth]{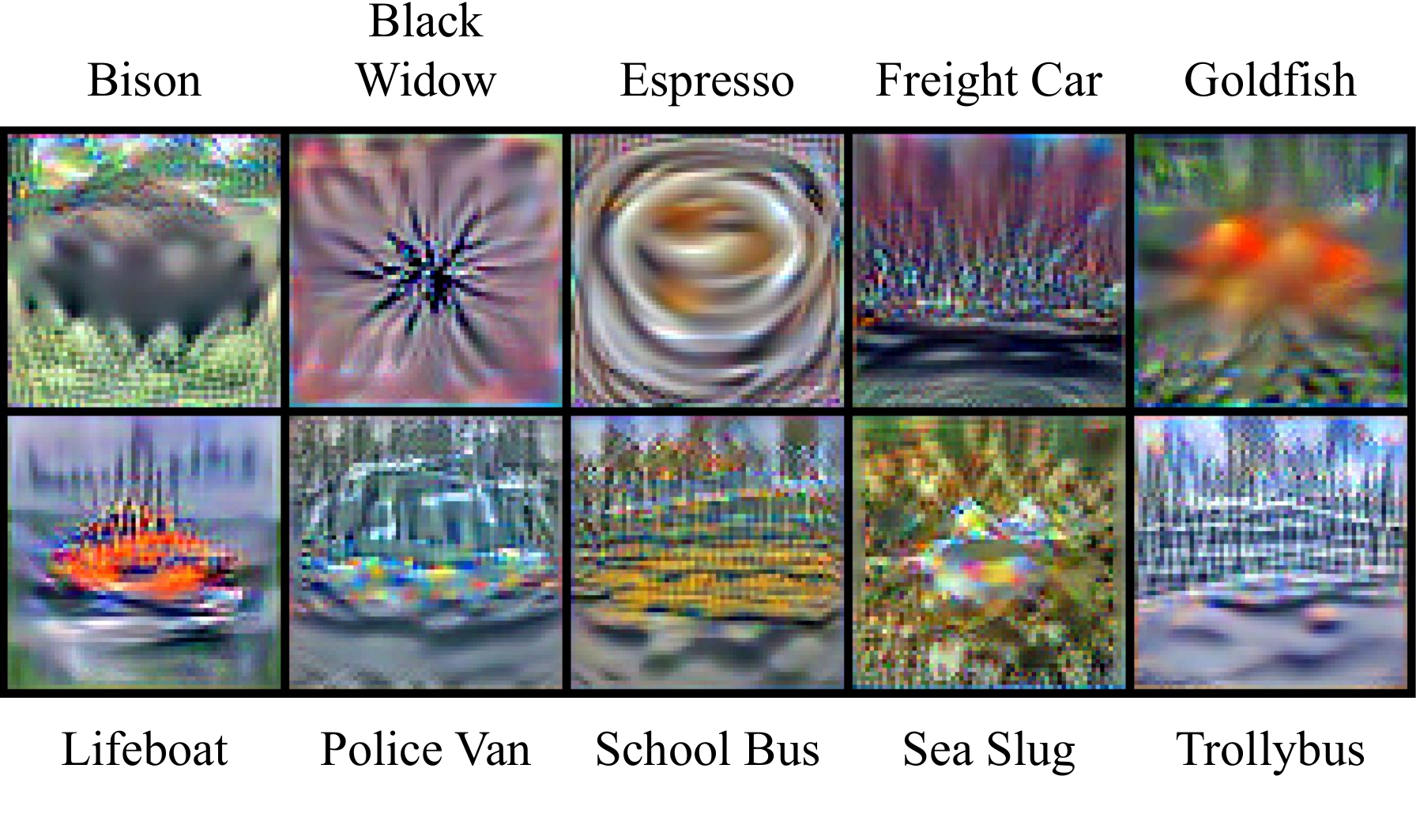}
    \vspace{-20pt}
    \caption{Most-correct distilled images for Tiny ImageNet \smaller{($\geq 30\%$)}}
    \label{fig:tinygood}
    \vspace{-8pt}
\end{figure}

\begin{figure}
    \centering
    \includegraphics[width=\linewidth]{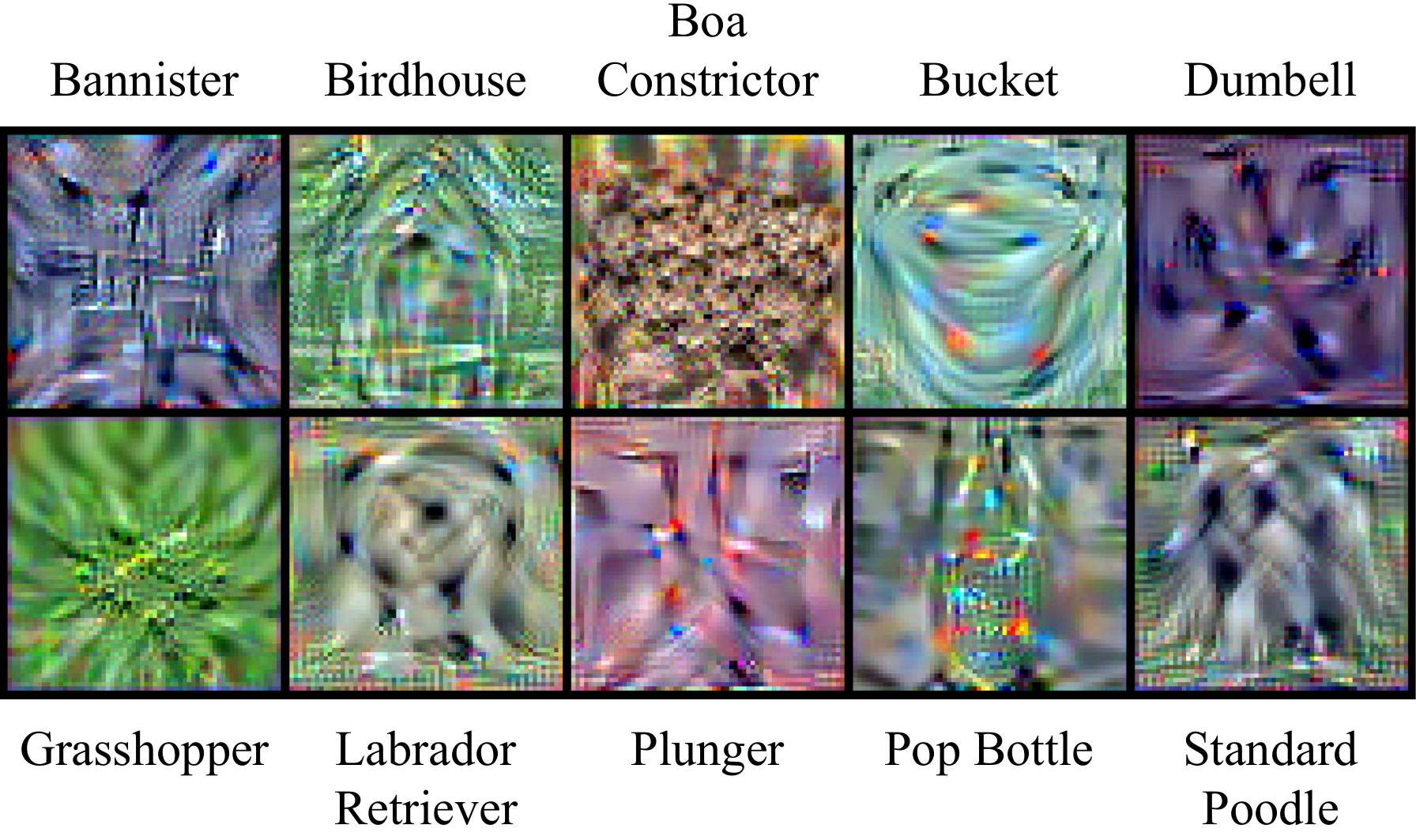}
    \vspace{-8pt}
    \caption{Least-correct distilled images for Tiny ImageNet \smaller{($\leq 4\%$)}}
    \label{fig:tinybad}
    \vspace{-8pt}
\end{figure}

\begin{figure*}
    \centering
    \includegraphics[width=\linewidth]{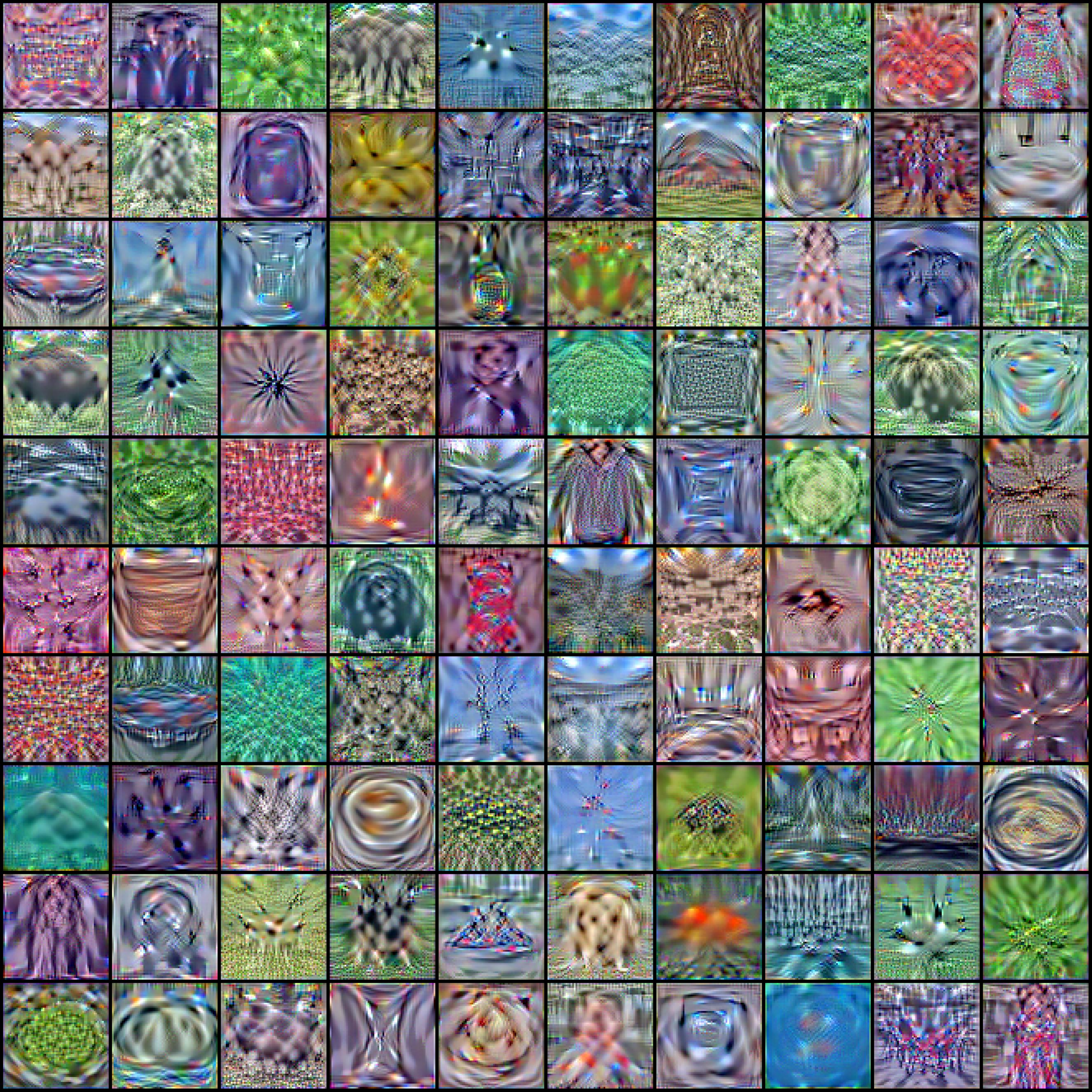}
    \caption{Tiny ImageNet, 1 Image Per Class (Classes 1-100)}
    \label{fig:tiny1}
\end{figure*}

\begin{figure*}
    \centering
    \includegraphics[width=\linewidth]{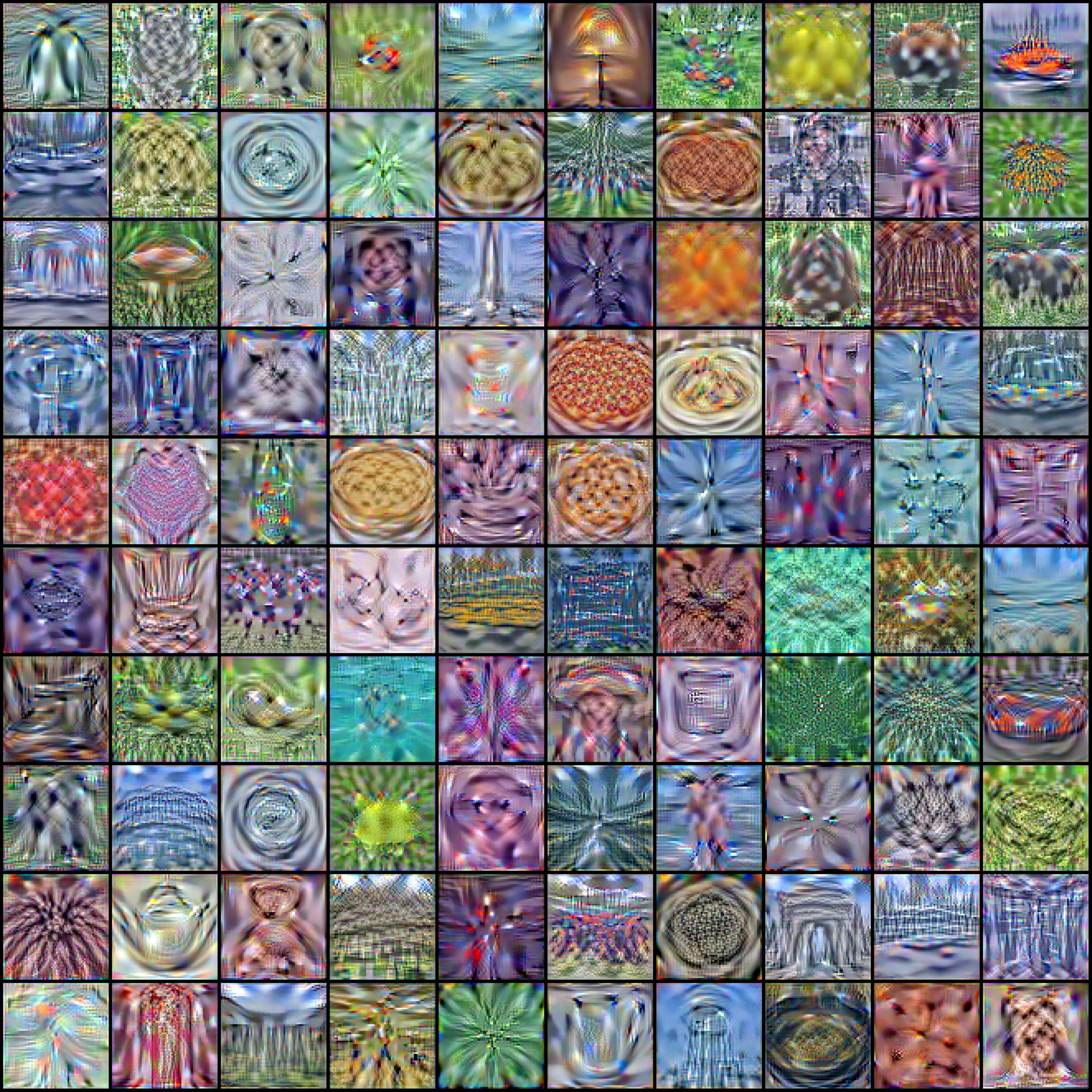}
    \caption{Tiny ImageNet, 1 Image Per Class (Classes 101-200)}
    \label{fig:tiny2}
\end{figure*}

\begin{figure*}
    \centering
    \includegraphics[width=\linewidth]{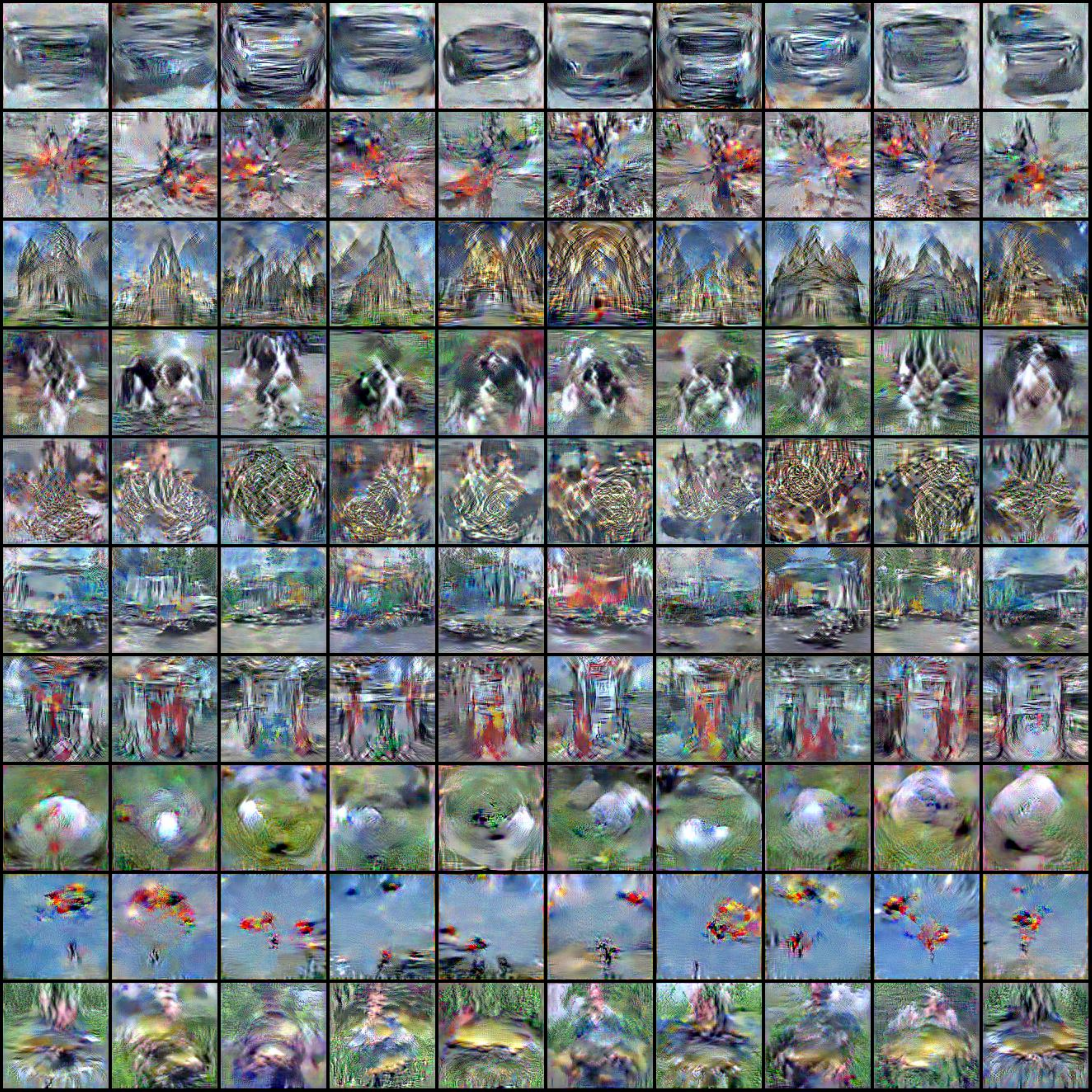}
    \caption{ImageNette, 10 Images Per Class}
    \label{fig:nette_10}
\end{figure*}

\begin{figure*}
    \centering
    \includegraphics[width=\linewidth]{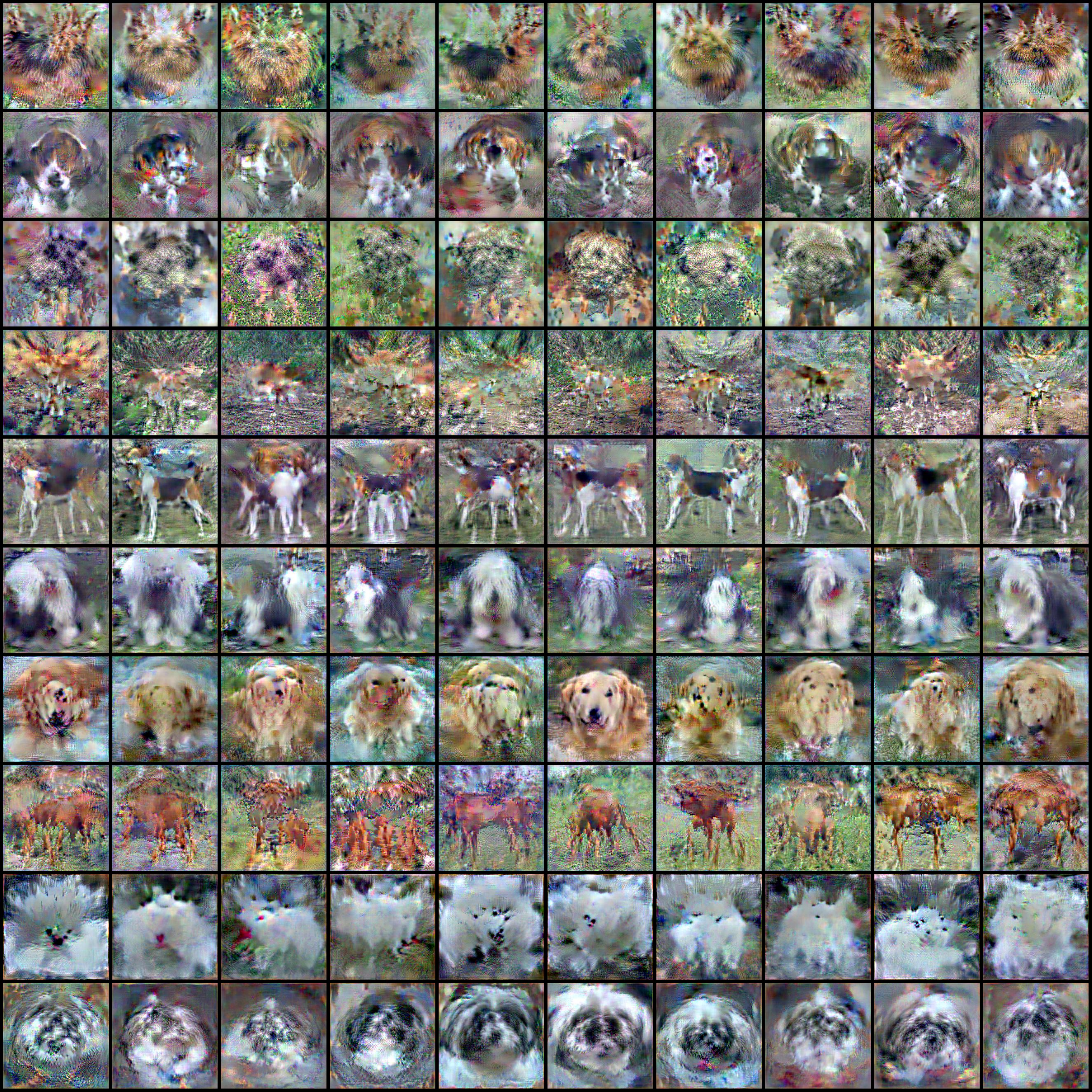}
    \caption{ImageWoof, 10 Images Per Class}
    \label{fig:woof_10}
\end{figure*}

\begin{figure*}
    \centering
    \includegraphics[width=\linewidth]{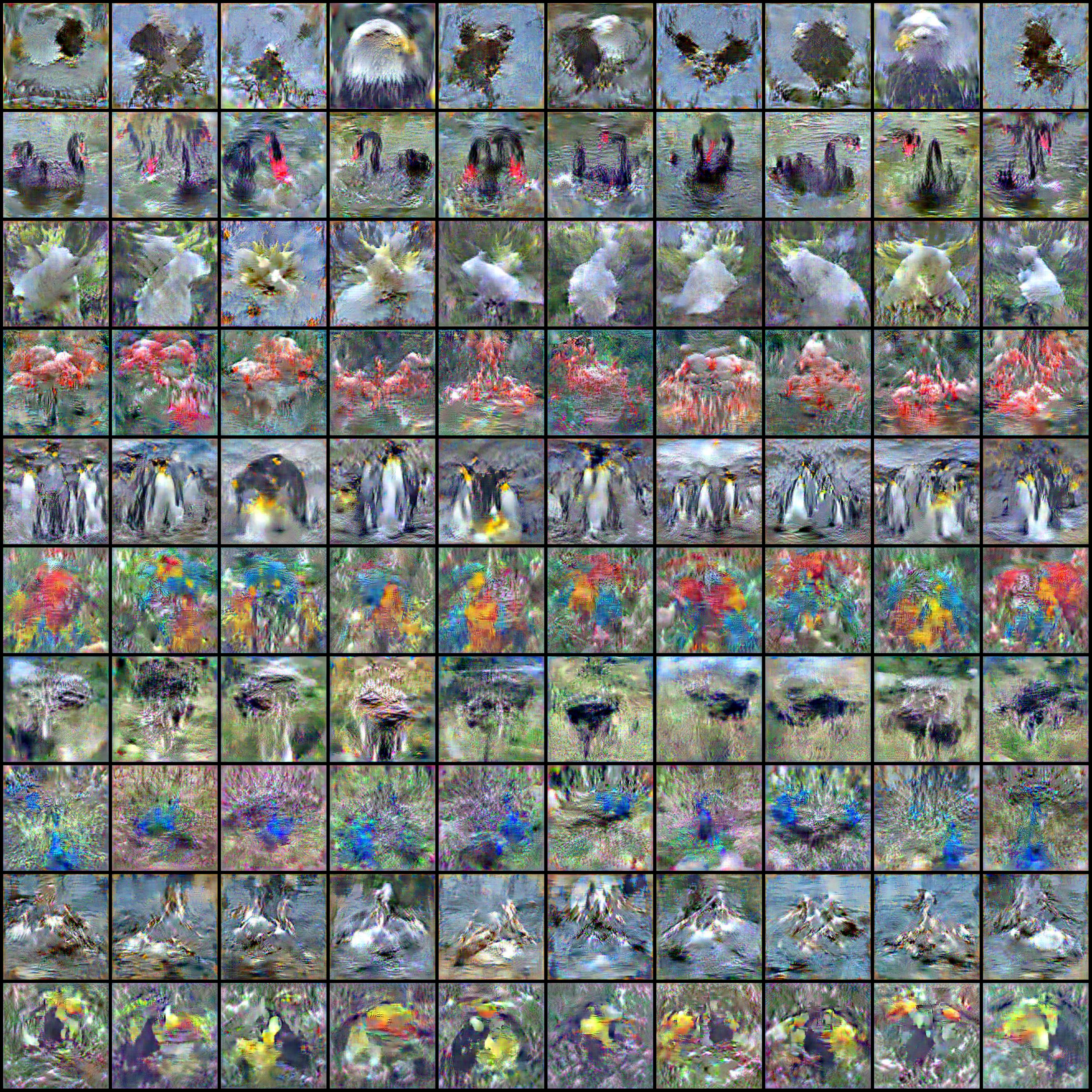}
    \caption{ImageSquawk, 10 Images Per Class}
    \label{fig:squawk_10}
\end{figure*}

\begin{figure*}
    \centering
    \includegraphics[width=\linewidth]{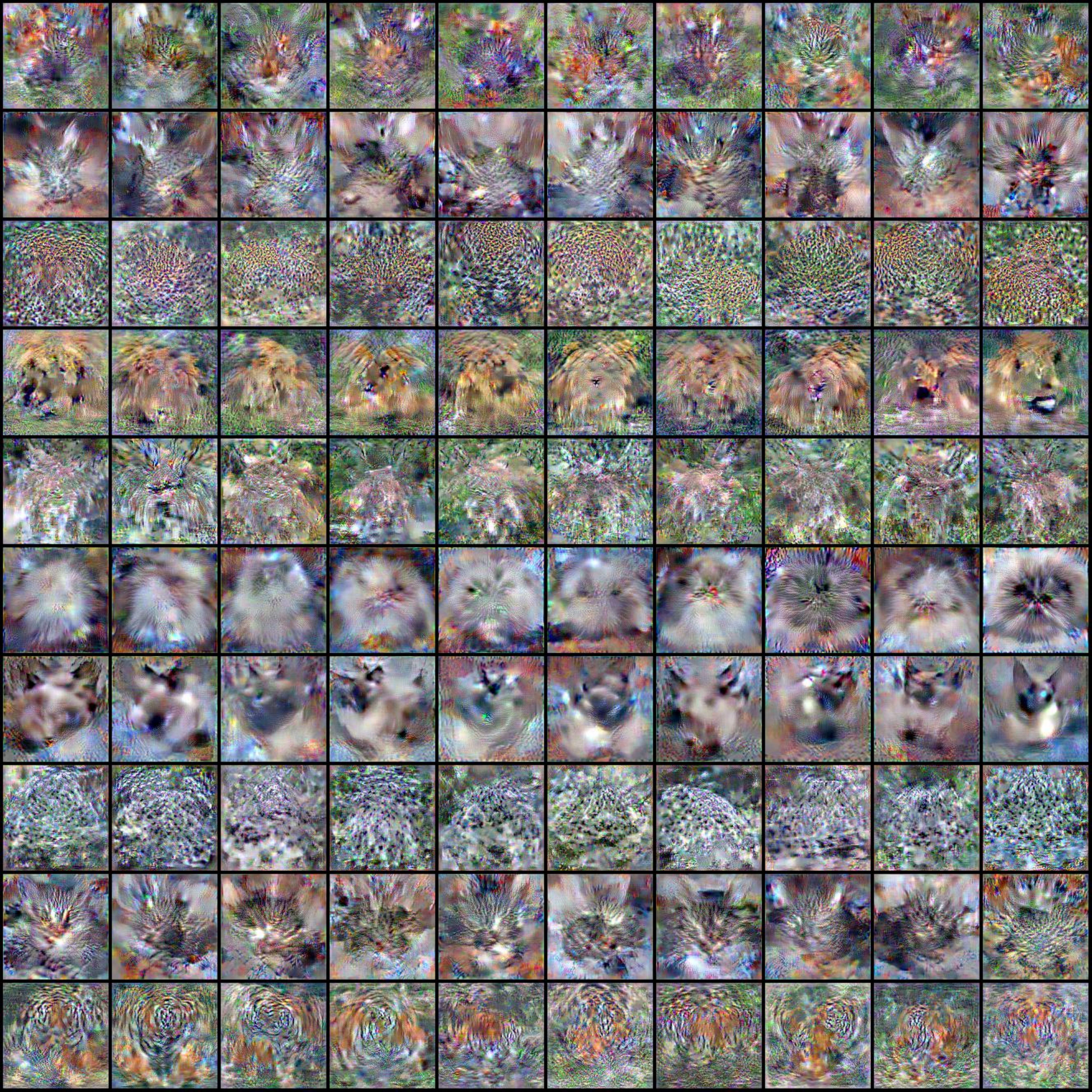}
    \caption{ImageMeow, 10 Images Per Class}
    \label{fig:meow_10}
\end{figure*}

\begin{figure*}
    \centering
    \includegraphics[width=\linewidth]{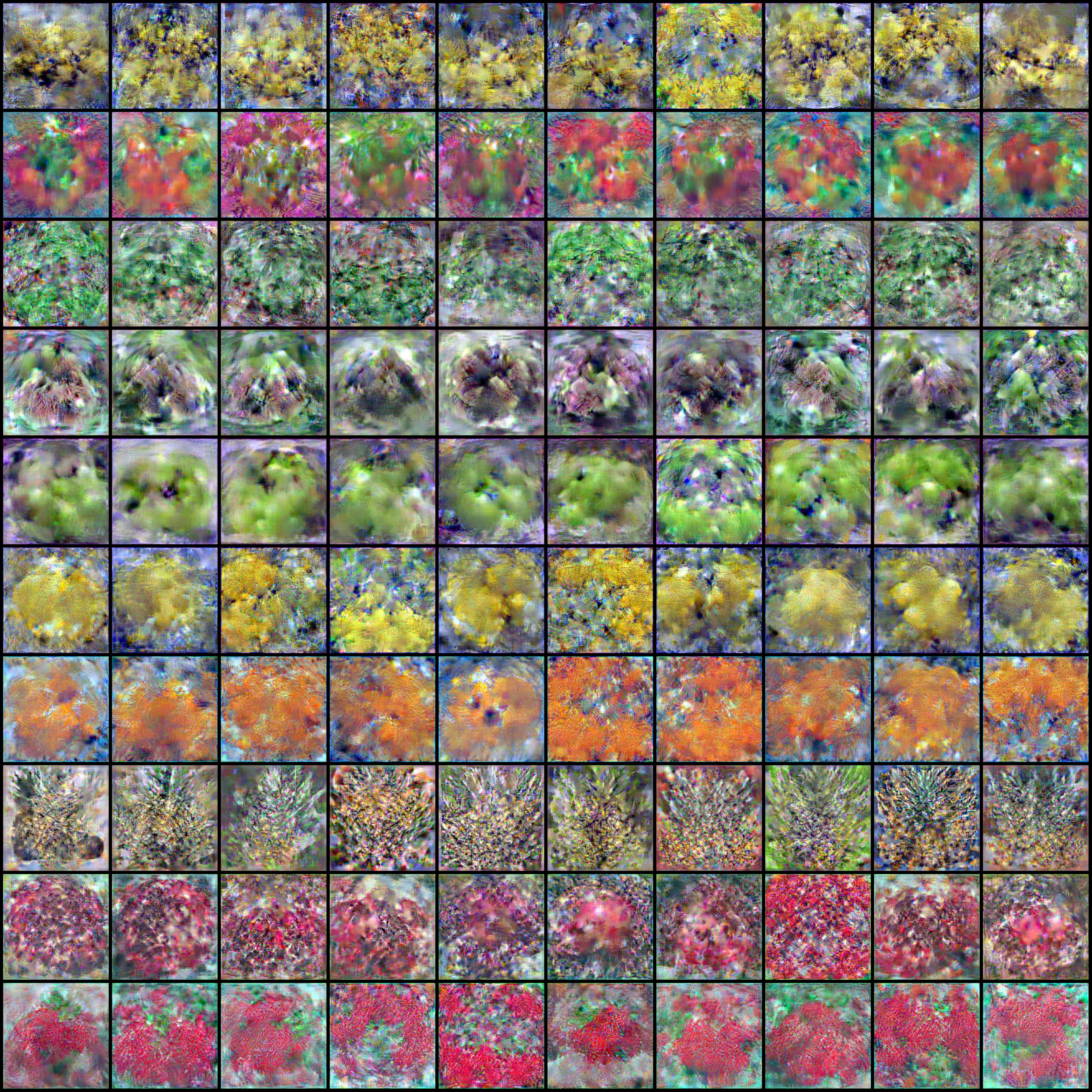}
    \caption{ImageFruit, 10 Images Per Class}
    \label{fig:fruit_10}
\end{figure*}

\begin{figure*}
    \centering
    \includegraphics[width=\linewidth]{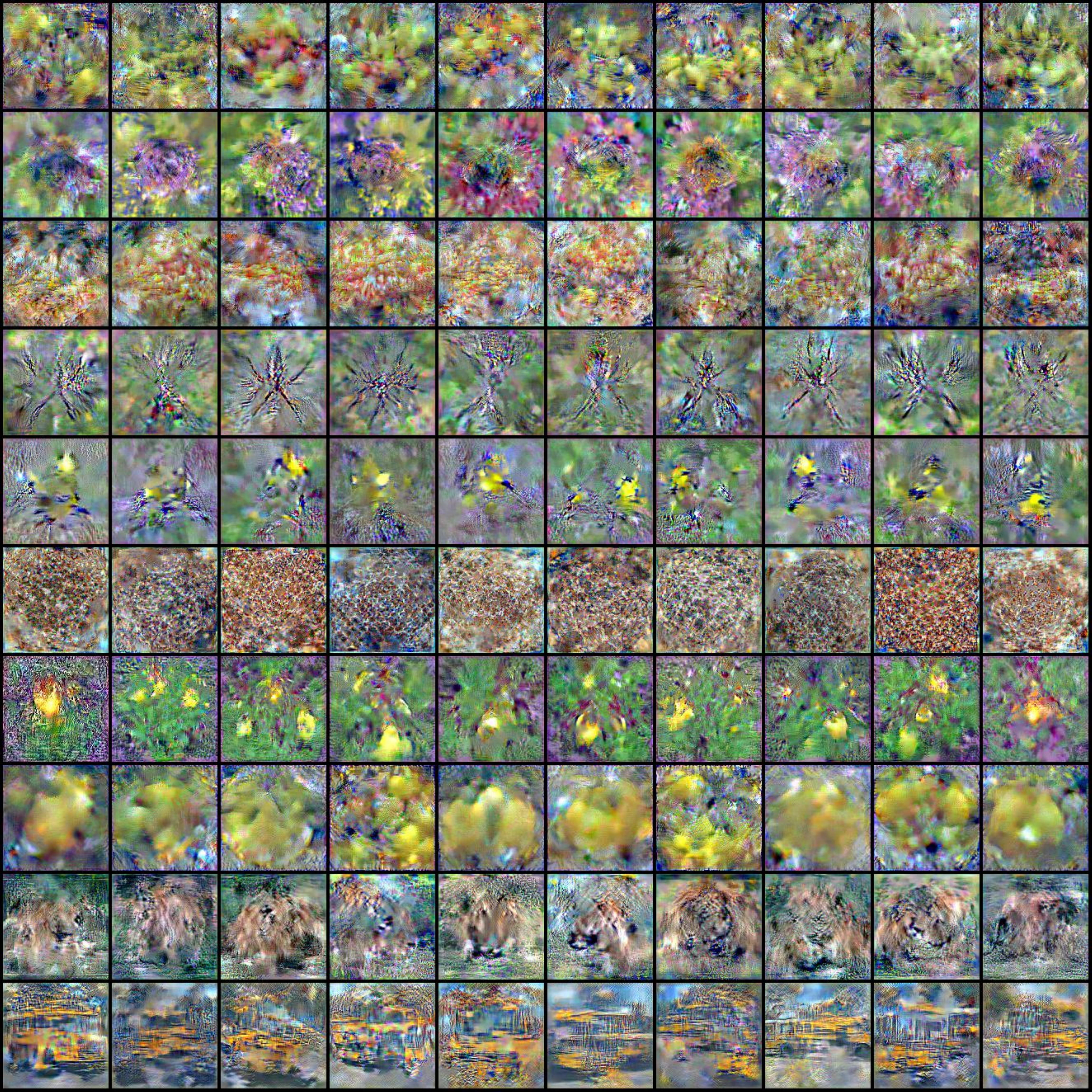}
    \caption{ImageYellow, 10 Images Per Class}
    \label{fig:yellow_10}
\end{figure*}

We first include some additional visualizations here. CIFAR-100 (1 image per class) can be seen in \reffig{cifar-100}. 
All of Tiny ImageNet (1 image per class) is broken up into Figures \ref{fig:tiny1} and \ref{fig:tiny2}. We specifically show the 10 best and worst-performing distilled classes in Figures \ref{fig:tinygood} and \ref{fig:tinybad} respectively.
We include 10 image per class visualizations of all our 128$\times$128 ImageNet subsets in Figures \ref{fig:nette_10}-\ref{fig:yellow_10}.

\subsection{Additional Quantitative Results}

\myparagraph{Analysis of learned learning rates \boldmath{$\alpha$}.}
In \reffig{lr}, we explore the effect of our learnable synthetic step size $\alpha$. The left plot confirms that we learn different values of $\alpha$ for different combinations of $M$ and $N$. The logic here is that different numbers of synthetic steps $N$ require a different step size $\alpha$ to cover the same distance as $M$ real steps. The right plot illustrates the practical benefits of our adaptive learning rate; instead of yet another hyper-parameter to tune, our adaptive learning rate works from a wide range of initializations.

\myparagraph{Effects of ZCA Whitening}
Note that \texttt{DC}, \texttt{DSA}, and \texttt{DM} do not use ZCA normalization, while \texttt{KIP} started using ZCA as it was a ``crucial ingredient for [their] strong results.'' We report our results w/o ZCA in \reffig{zca} (Left).
We find that ZCA normalization is \textit{not} critical to our performance. 
However, the expert models trained without ZCA normalization take significantly longer to converge. Thus, when distilling using these models as experts, we must use a larger value of $T^+$ (and therefore save more model snapshots). When we use a larger value of $T^+$ for non-ZCA distillations, we get results comparable to or even better than those of the ZCA distillations.  
In short, ZCA helps expert convergence but does not notably improve distillation performance. 

\begin{figure}[!h]
\renewcommand\arraystretch{0.9}
\centering
\scriptsize
\setlength{\tabcolsep}{2pt}
\resizebox{0.48\linewidth}{!}{
\begin{tabular}{cc|c|ccc|cccccc|c}
\toprule
        Dataset & Img/Cls  & Yes-ZCA & No-ZCA \\ \midrule  %

\multirow{3}{*}{CIFAR-10}        & 1      & 46.3 & 45.2 \\
                                & 10     & 65.3 & 62.8\\  
                                & 50       & 71.5 & 71.6\\ \midrule
                                
\multirow{3}{*}{CIFAR-100}     & 1      &  24.3 & 22.7\\ 
                              & 10       & 39.4 & 40.1\\  
                              & 50       & 47.7 & 47.2\\   \bottomrule

\end{tabular}
}
\setlength{\tabcolsep}{2pt}
\resizebox{0.48\linewidth}{!}{
\begin{tabular}{cc}
&CIFAR-10, 10 img/cls\\
\rotatebox[origin=c]{90}{Validation Acc. \%}& \hspace{0cm}\includegraphics[trim={0.8cm 0.4cm 0.8cm 0.4cm},clip,align=c,width=0.4\linewidth]{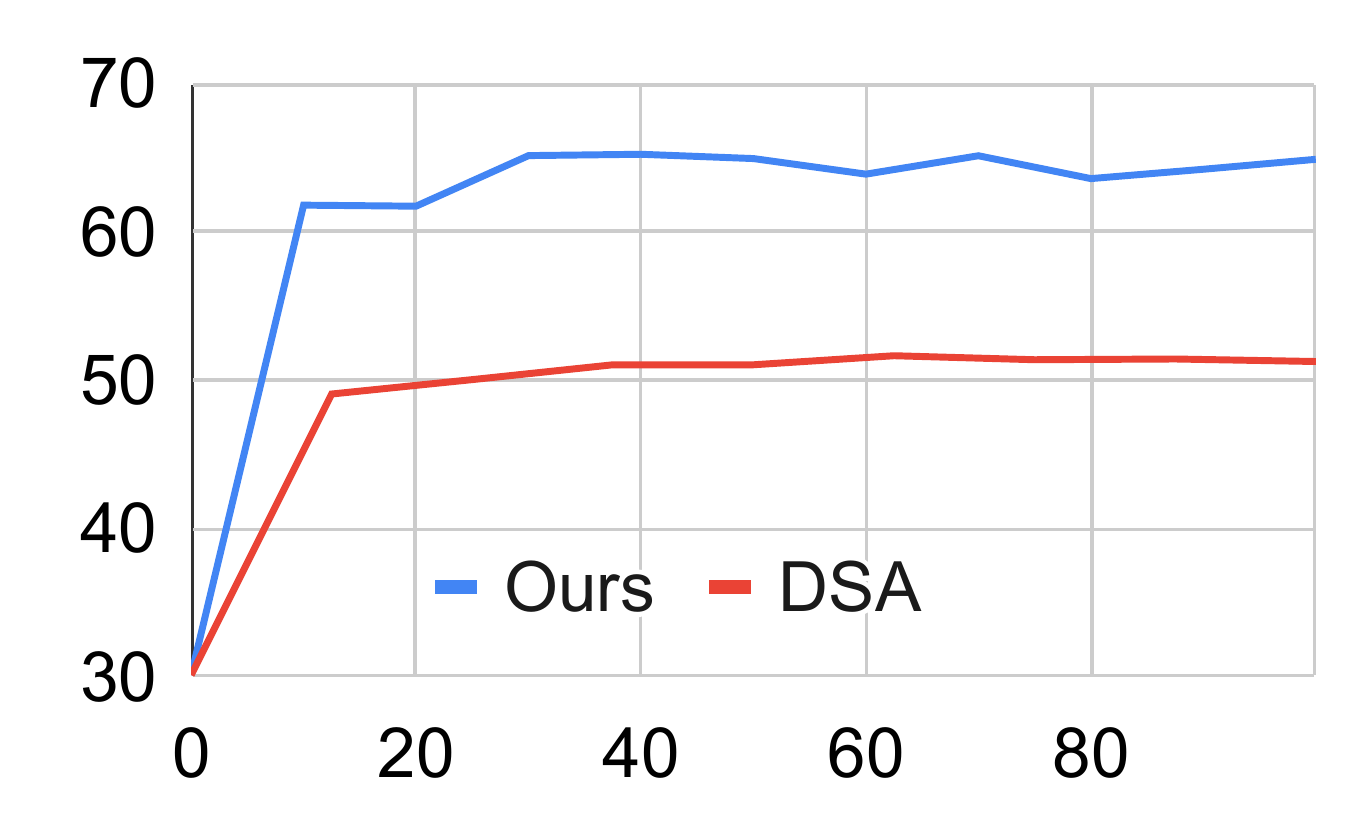}\\
&Distilling Time (mins)
\end{tabular}
}
\vspace{-8pt}
\caption{\textbf{Left}: ZCA Ablation. \textbf{Right}: Distillation Time.}
\vspace{-8pt}
\lblfig{zca}
\end{figure}

\subsubsection{Additional Ablation Studies}

\begin{figure}
    \centering
    \includegraphics[width=\linewidth]{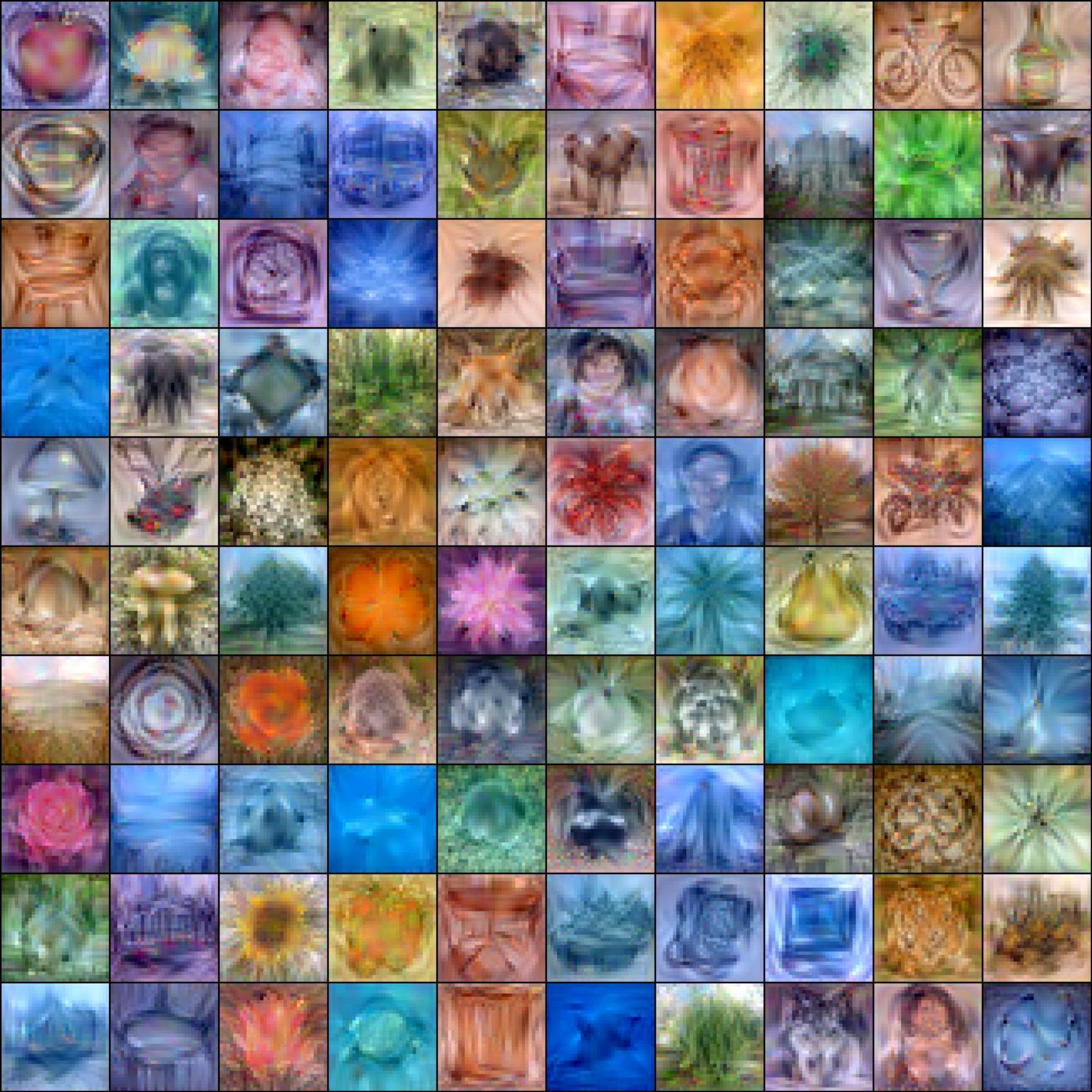}
    \caption{CIFAR-100, Initialized as Random Noise}
    \label{fig:noise}
\end{figure}

\begin{figure}
    \centering
    \includegraphics[width=\linewidth]{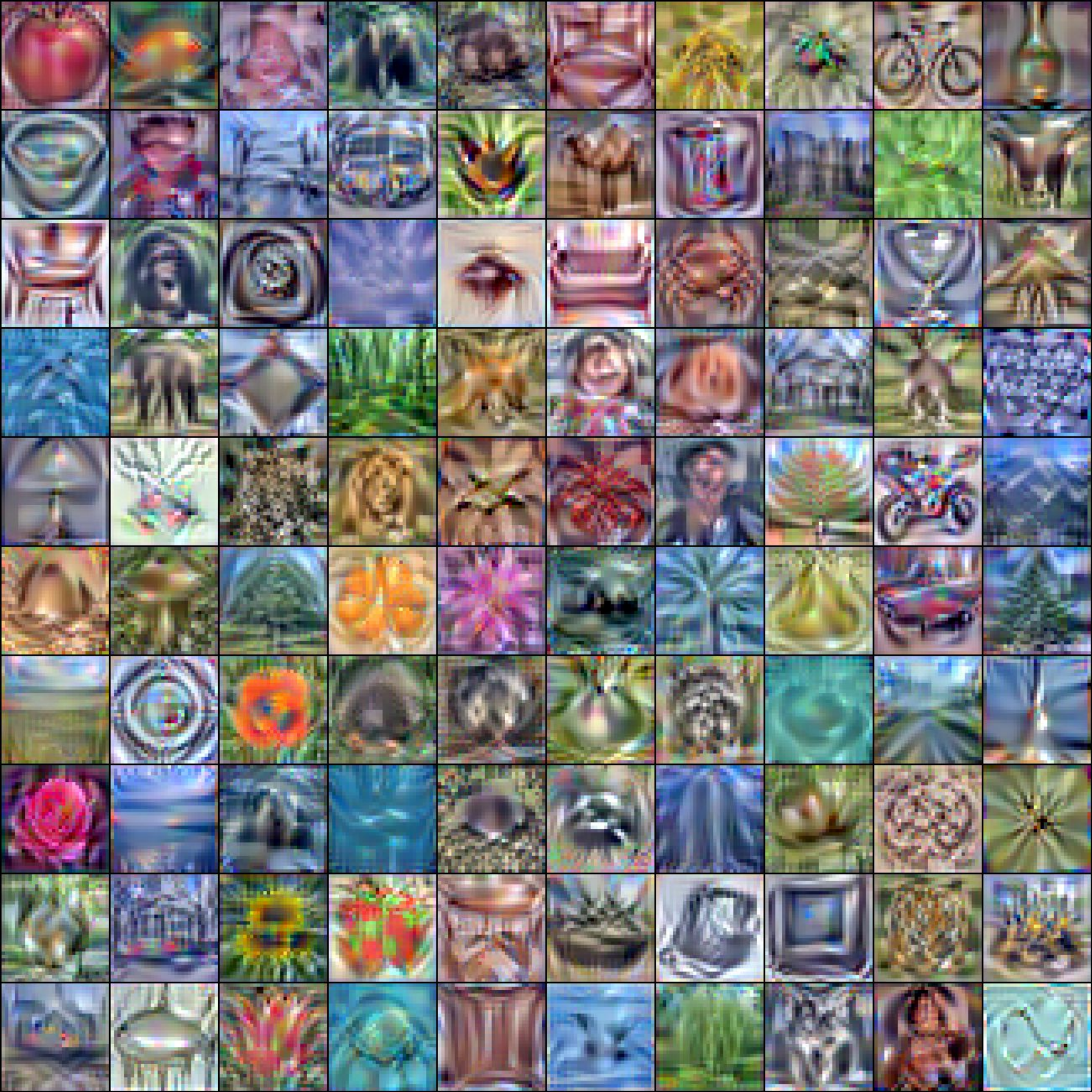}
    \caption{CIFAR-100, No ZCA Whitening}
    \label{fig:nozca}
\end{figure}

\begin{figure}
    \centering
    \includegraphics[width=\linewidth]{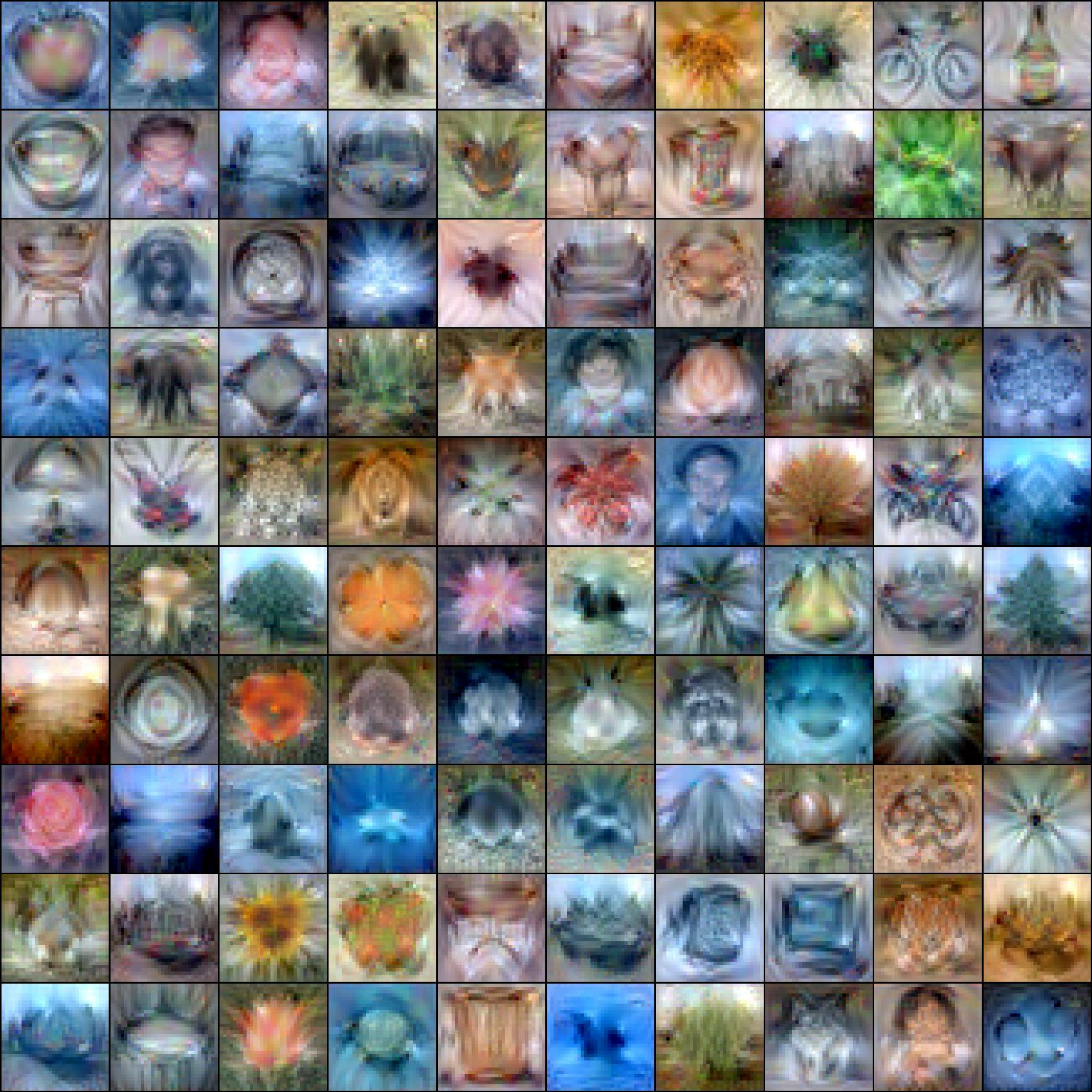}
    \caption{CIFAR-100, No Differentiable Augmentation}
    \label{fig:noaug}
\end{figure}

\begin{figure}
    \centering
    \includegraphics[width=\linewidth]{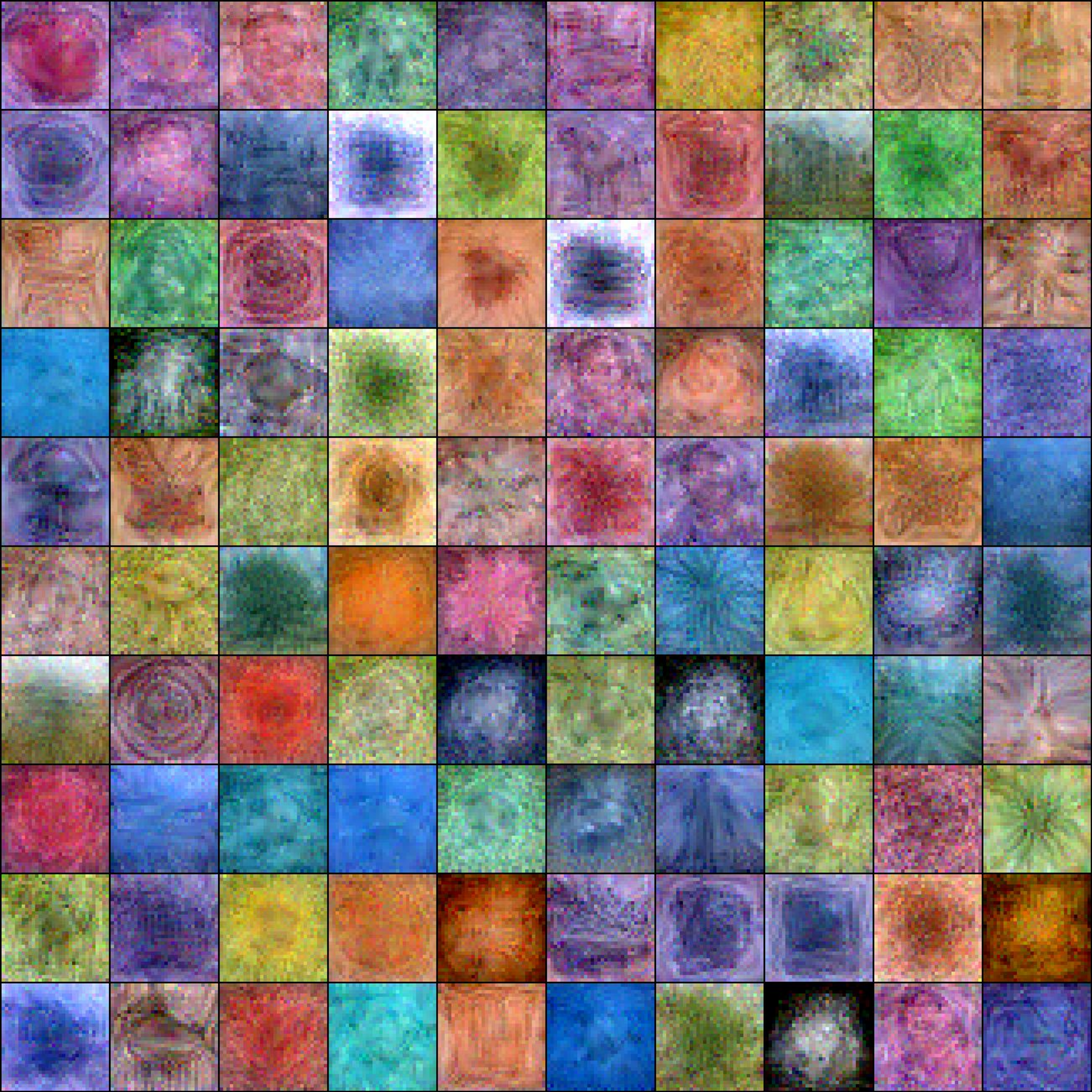}
    \caption{CIFAR-100, Only 1 Expert Trajectory}
    \label{fig:1exp}
\end{figure}

\myparagraph{Initialization, normalization, and augmentation.}
In the main paper, we show ablations over several hyper-parameters. Here, we study the role of initialization, data normalization, and data augmentation for CIFAR-100 (1 image per class) in \reftbl{ablation}. For initialization in particular, recall that we typically initialize our synthetic images with real samples. Here, we evaluate initializing with Gaussian noise instead. Visualizations of these distilled sets can be seen in Figures \ref{fig:noise}-\ref{fig:noaug}. We also include a visualization of a set distilled with only one expert trajectory in \reffig{1exp}.

\begin{table}[!h]
    \centering
        \smaller
    \resizebox{\linewidth}{!}{
    \begin{tabular}{c|cccc}
       \toprule Setting  &  Ours & Gauss. Init. &  w/o ZCA & w/o Diff. Aug.\\
       \midrule Acc. & 24.3 $\pm$ 0.3 &  22.9 $\pm$ 0.4 & 19.1 $\pm$ 0.6 & 20.7 $\pm$ 0.7\\
       \bottomrule
    \end{tabular}
    }
    \caption{As we ablate our categorical hyper-parameters, we still achieve state-of-the-art performance (compared to \texttt{DSA}: 13.9\%). This is evaluated on CIFAR-100 with 1 image per class. Each design choice in our final method improves the performance of distilled images.  Here we use the default set of hyper-parameters for these ablations.}%
    \lbltbl{ablation}
\end{table}

\begin{figure}
\vspace{-4pt}
    \begingroup
    \setlength{\tabcolsep}{1pt}
    \centering
    \begin{tabular}{cccc}
        & \smaller{CIFAR-100 (1 Image / Class)} & & \smaller{CIFAR-100 (1 Image / Class)}\\
        \rotatebox[origin=c]{90}{\smaller{Validation Acc. \%}} &  \includegraphics[align=c,width=0.43\linewidth]{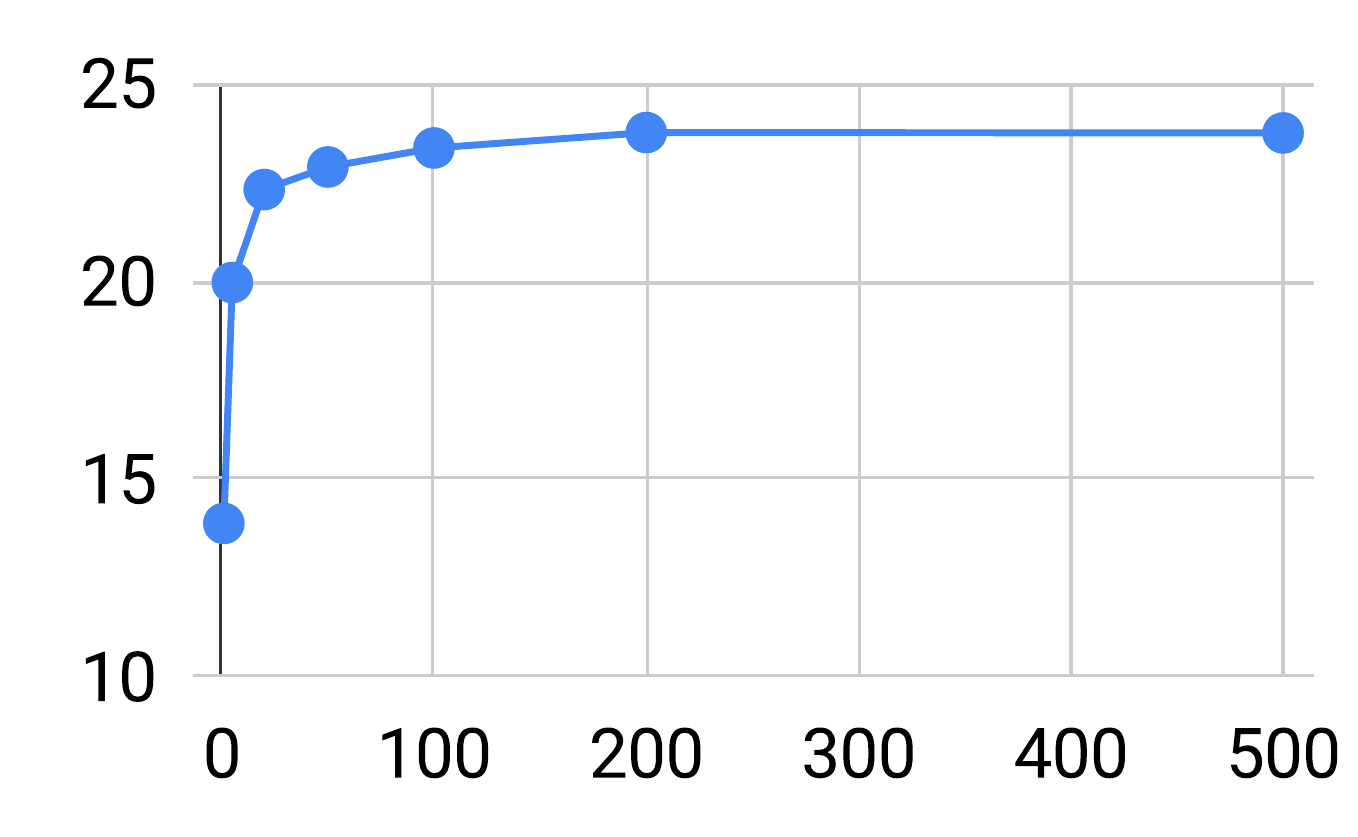}& \hfill\;\;\;\;\rotatebox[origin=c]{90}{\smaller{Validation Acc. \%}} &  \includegraphics[align=c,width=0.43\linewidth]{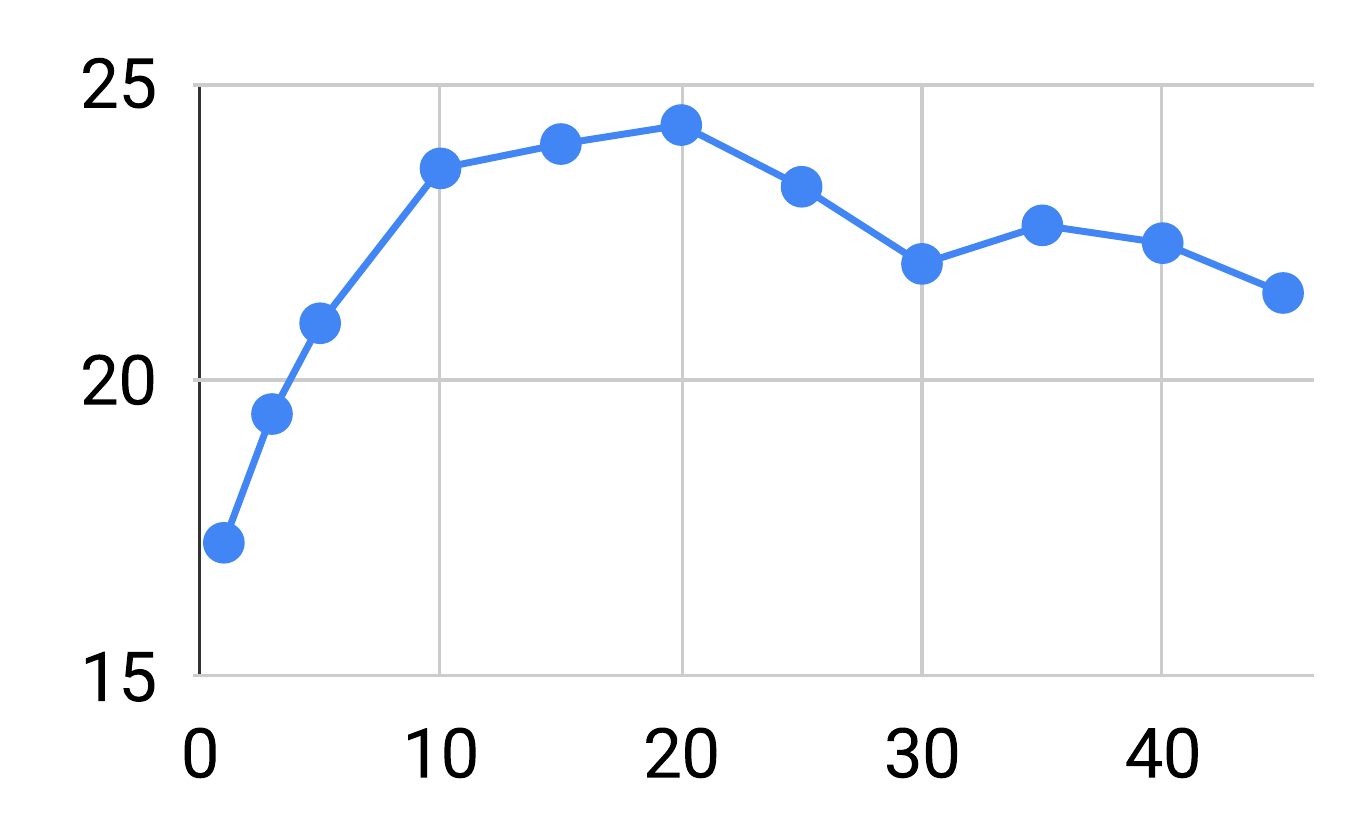}\\
         & \smaller{\;\;\;\;\;\# Expert Trajectories} & & \smaller{\;\;$T^+$: Max Start Epoch}
    \end{tabular}
    \endgroup
        \vspace{-7pt}
    \caption{\textbf{Left:} We see logarithmic performance improvement with respect to the number of expert trajectories used, quickly saturating near 200. \textbf{Right:} The upper bound on the expert epoch at which the synthetic data starts working cannot be too high or low to ensure quality learning signal.}
    \lblfig{experts}
    \vspace{-10pt}
\end{figure}

\begin{figure}
    \begingroup
    \setlength{\tabcolsep}{1pt}
    \centering
    \begin{tabular}{@{\hskip 0pt}cc@{\hskip 0pt}cc@{\hskip 0pt}}
    & \smaller{CIFAR-100 (1 Image / Class)} & & \smaller{CIFAR-100 (1 Image / Class)}\\
         \rotatebox[origin=c]{90}{\fontsize{6}{5}\selectfont{$\alpha$: Learned Step Size}} &  \includegraphics[align=c,width=0.455\linewidth]{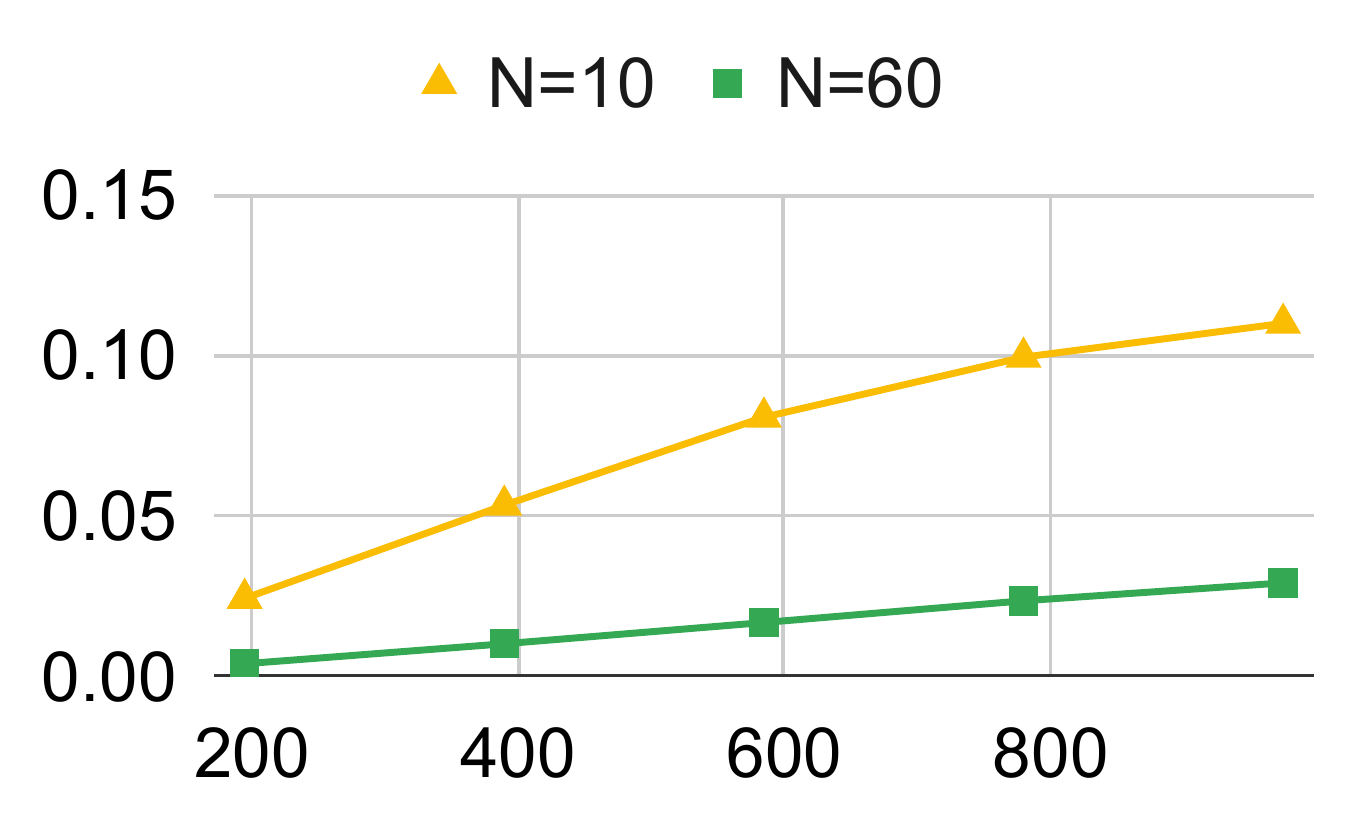}& \hfill\;\;\rotatebox[origin=c]{90}{\fontsize{6}{5}\selectfont{Validation Acc. \%}} &  \includegraphics[align=c,width=0.458\linewidth]{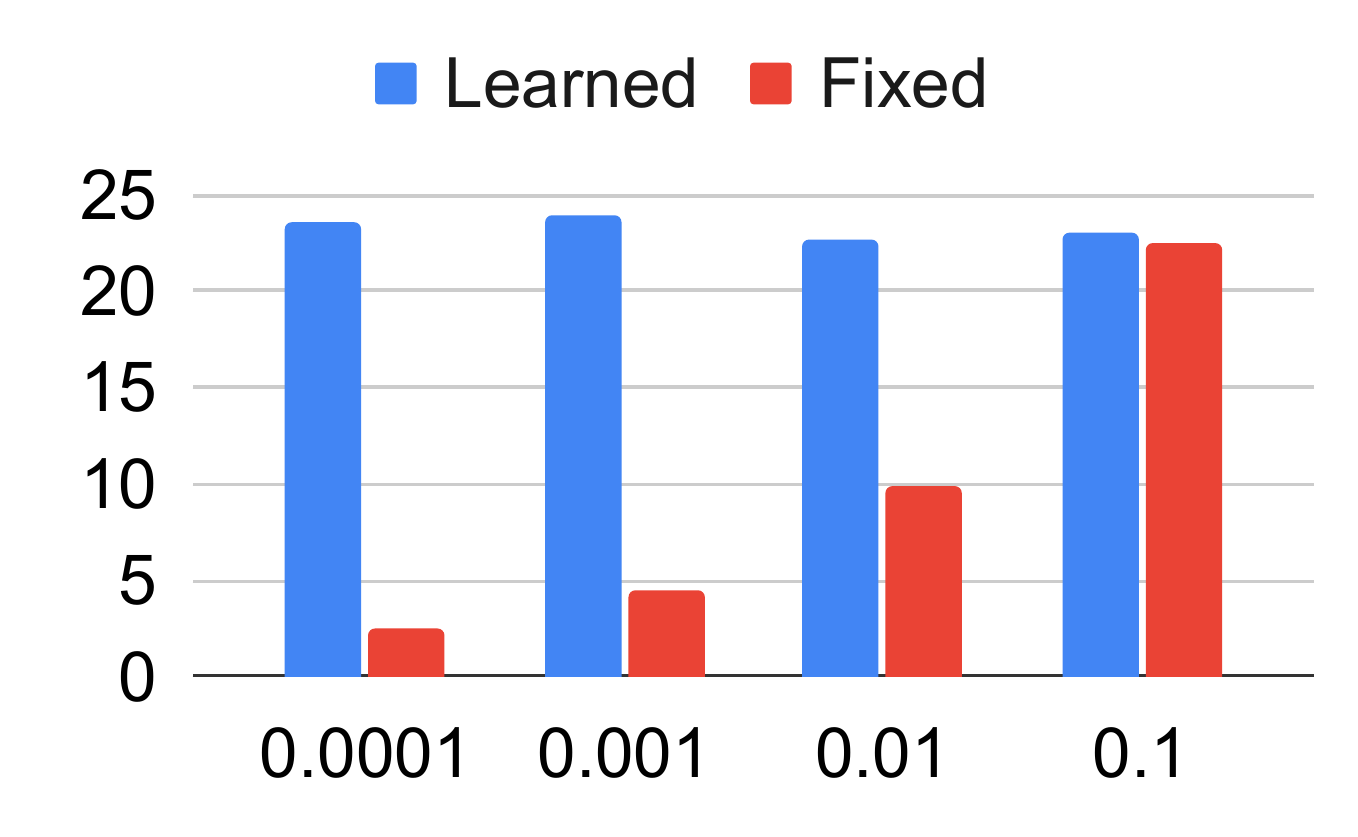}\\[-1.3ex]
         & \fontsize{6}{5}\selectfont{\;\;\;\;\;$M$: Expert Steps} & & \fontsize{6}{5}\selectfont{\;\;$\alpha_0$: Initial Synthetic Step Size}
    \end{tabular}
    \endgroup
        \vspace{-9pt}
    \caption{\textbf{Left:} Our learned synthetic step size $\alpha$ seems to scale inversely with the number of synthetic steps $N$ to cover the same distance in parameter space as the expert steps $M$.
    \textbf{Right:} Having a learnable step size $\alpha$ saves us from having to search for an appropriate fixed $\alpha_0$.
    }
    \lblfig{lr}
    \vspace{-10pt}
\end{figure}

\myparagraph{Performance w.r.t. the number of expert trajectories.}
Since they effectively make up our method's ``training set,'' it is reasonable to assume that having more expert trajectories would lead to better performance. We see that this is indeed the case for the CIFAR-100, 1 image per class setting in \reffig{experts} (left). However, what's most interesting is the sharp, logarithmic increase in validation accuracy w.r.t. the number of experts. We note the most amount of improvement when increasing from  1 to 20 experts but see almost complete saturation by the time we reach 200. Given how high-dimensional the parameter space of a neural network is, it is remarkable that we can achieve such high performance with so few expert trajectories. \vspace{10pt}

\myparagraph{Performance w.r.t. expert time-step range.}
When we initialize our student networks, we do so at a randomly selected time-step from an expert trajectory. We find that it is important to put an upper bound on this starting time-step (\reffig{experts}, right). If the upper bound is too high, the synthetic data receives gradients from points where the experts movements are small and uninformative. If it is too low, the synthetic data is never exposed to mid and later points in the trajectories, missing out on a significant portion of the training dynamics.

\begin{figure}
    \centering
    \includegraphics[width=\linewidth]{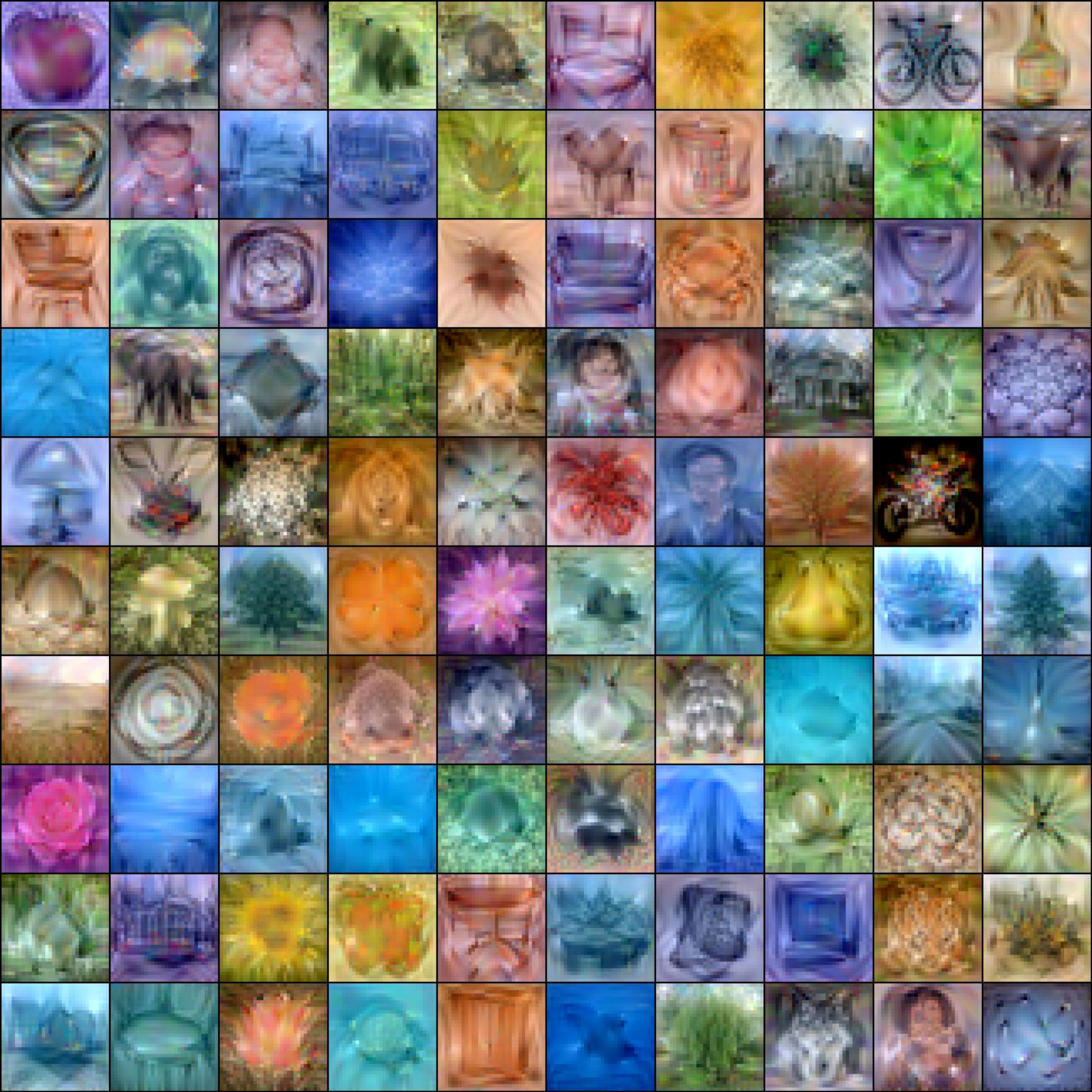}
    \caption{CIFAR-100, 1 Image Per Class}
    \lblfig{cifar-100}
    \vspace{-9pt}
\end{figure}

\subsection{Experiment Details}
\myparagraph{Hyper-Parameters.}
In \reftbl{hparams}, we enumerate the hyper-parameters used for our results reported in the main text. Limited compute forced us to batch our synthetic data for some of the larger sets. The ``ConvNet'' architectures are as explained in the main text.
\begin{table}
\renewcommand\arraystretch{0.9}
\centering
\scriptsize
\setlength{\tabcolsep}{2pt}
\resizebox{0.98\linewidth}{!}{
\begin{tabular}{ccc|ccccc}
\toprule
Dataset         & Model & Img/Cls & \makecell{Synthetic \\ Steps \\ ($N$)} & \makecell{Expert \\ Epochs \\($M^\dagger$)}  & \makecell{Max Start \\ Epoch \\ ($T^+$)}  & \makecell{Synthetic \\ Batch Size \\ ($|b|$)} & ZCA \\ \midrule

\multirow{3}{*}{CIFAR-10}  & \multirow{3}{*}{ConvNetD3}      & 1   & 50   & 2  & 2 & - & Y\\
                               & & 10  & 30    & 2  & 20 & - & Y\\
                               & & 50  & 30      & 2 & 40  &  -  & N\\ \midrule
                                
\multirow{3}{*}{CIFAR-100} & \multirow{3}{*}{ConvNetD3}     & 1   & 20   & 3  &  20  & - & Y \\ 
                              && 10  & 20      & 2  & 20  & - & N\\  
                              && 50  & 80     & 2  & 40  & -   & Y     \\  \midrule

\multirow{3}{*}{Tiny ImageNet} & \multirow{3}{*}{ConvNetD4} & 1   & 10    &  2  &  10  &  - & -\\ 
                             & & 10  & 20      &  2  & 40  &  200 & -  \\  
                             & & 50  & 20     & 2  & 40  & 300  & -\\  \midrule
                              
\multirow{2}{*}{ImageNet (All)} & \multirow{2}{*}{ConvNetD5} & 1   & 20    &  2  &  10  &  - & -\\ 
                             & & 10  & 20      &  2  & 10  &  20  & - \\  \bottomrule

\end{tabular}
}
\caption{Hyper-parameters used for our best-performing distillation experiments. A synthetic batch size of ``-'' indicates that we used the full support set at each synthetic step. Note: instead of number of expert \textit{updates} ($M$), here we list number of expert \textit{epochs} ($M^\dagger$) for simplicity across datasets.}
\lbltbl{hparams}
     \vspace{-6pt}
\end{table}

\myparagraph{Compute resources.} We had a relatively limited compute budget for our experiments, using any GPU we could access. As such, our experiments were run on a mixture of RTX2080ti, RTX3090, and RTX6000 GPUs. The largest amount of VRAM we used for a single experiment was 144GB over 6xRTX6000 GPUs.

\myparagraph{Training Time.} Distillation time varied based on dataset and type and number of GPUs used. Regardless of dataset or compute resources, time per distillation iteration scaled linearly with the number of synthetic steps $N$. For CIFAR-100, 1 image per class with $N=20$, we had an average time of 0.6 seconds per distillation step when using a single RTX3090. We ran our experiments for 10000 distillation steps but saw the most improvement within the first 1000.

Our distillation time, in general, is comparable to \texttt{DC/DSA}, as they also utilize a bi-level optimization. In the 10 img/class setting (for example), \texttt{DC/DSA} trains on the synthetic data for 50 epochs on the between each update. We include a sample distillation curve in \reffig{zca} (Right). Both experiments were run on RTX3090. Note that  \texttt{KIP} requires over 1,000 GPU hours.

Regarding the distillation time for learning different sets on CIFAR10/100 and TinyImageNet, we report them in \reftbl{time}. Note that most improvement occurs within the first 1k iterations, but we continue training for 10k.
\begin{table}[!h]
\renewcommand\arraystretch{0.9}
\centering
\scriptsize
\setlength{\tabcolsep}{2pt}
\resizebox{0.75\linewidth}{!}{
\begin{tabular}{cc|c|ccc}
\toprule
        Dataset & Img/Cls  & \makecell{1 Iter.\\(sec)} & \makecell{1k Iter.\\(min)} & \makecell{5k Iter.\\(min)} & \makecell{10k Iter.\\(min)}\\ \midrule  %

\multirow{3}{*}{CIFAR-10}        & 1      & 0.5 & 8 & 42 & 83\\
                                & 10     & 0.6 & 10 & 50 & 100\\  
                                & 50       & 0.8 & 13 & 67 & 133\\ \midrule
                                
\multirow{3}{*}{CIFAR-100}     & 1      &  0.6 & 10 &50 & 100\\ 
                              & 10       & 0.8 & 13 & 67 & 133\\  
                              & 50       & 1.9 & 32 & 158 & 317\\  \midrule

\multirow{3}{*}{Tiny ImageNet} & 1       &  1.1 & 18 & 92 & 183\\ 
                              & 10      &  2.3 & 38 & 192 & 383\\  
                              & 50      & 2.6 & 43 & 217 & 433\\  \bottomrule

\end{tabular}
}
\vspace{-8pt}
\caption{Distillation time for each dataset and support size.}
\lbltbl{time}
\end{table}

\myparagraph{KIP to NN}
In the \texttt{KIP} paper, results are presented for images distilled using the neural tangent kernel method and then evaluated by training a modified width-1024 ConvNetD3. Aside from the increased width of the finite model, the ConvNet architecture used in the \texttt{KIP} paper also has an additional 1-layer convolutional stem.

Using the training notebook provided with the \texttt{KIP} paper, we perform an exhaustive search over a reasonable set of hyper-parameters for the KIP to width-128 NN problem: \texttt{checkpoint} $\in$ \{112, 335, 1000\}, \texttt{weight\_decay} $\in $ \{0, 0.0001, 0.001, 0.01\}, \texttt{aug} $\in$ \{\texttt{True}, \texttt{False}\}, \texttt{zca} $\in$ \{\texttt{True}, \texttt{False}\}, \texttt{label\_learn} $\in$ \{\texttt{True}, \texttt{False}\}, and \texttt{norm} $\in$ \{\texttt{none}, \texttt{instance}\}. The architecture originally used for KIP to NN in the \texttt{KIP} paper contained no normalization layers. However, we found that with the smaller width, this model could not even converge on the synthetic training data for CIFAR-100, so we added instance normalization layers as found in the ConvNets we and \texttt{DC}, \texttt{DSA}, and \texttt{DM} use.

In \reftbl{kip_hyperparams}, we include the optimal hyper-parameters from this search that were used to obtain the KIP to NN (128\nobreakdash-width) values reported in the main text.

\begin{table}
\centering\vspace{-8pt}
\resizebox{0.95\linewidth}{!}{
\begin{tabular}{cc|cccccc}
\toprule
 & Img/Cls  & \makecell{Learn\\Labels} & Aug. & ZCA & Norm & \makecell{Weight\\Decay} & Ckpt.\\ \midrule 
\multirow{3}{*}{CIFAR-10}  & 1  & N & Y & Y & N & 0.001 & 1000\\
                          & 10  & N &  N & Y & N & 0.001 & 112\\
                          & 50  & N & N & Y & I & 0.01 & 112\\ \midrule
\multirow{2}{*}{CIFAR-100} & 1   & N & N & Y & I & 0.001 & 1000\\
                          & 10  & N & N & N & I & 0.001 & 1000\\
\bottomrule
\end{tabular}
}
\caption{Optimal hyper-parameters for our reported width-128 KIP to NN results. These were obtained via grid search using the notebook provided by the KIP authors.}
\lbltbl{kip_hyperparams}
\vspace{-8pt}
\end{table}


\clearpage


\vfill\break

\end{document}